# Parameter Learning of Logic Programs for Symbolic-statistical Modeling

**Taisuke Sato** SATO@MI.CS.TITECH.AC.JP
**Yoshitaka Kameya** KAME@MI.CS.TITECH.AC.JP
*Dept. of Computer Science, Graduate School of Information Science and Engineering, Tokyo Institute of Technology*
*2-12-1 Ookayama Meguro-ku Tokyo Japan 152-8552*

## Abstract

We propose a logical/mathematical framework for statistical parameter learning of parameterized logic programs, i.e. definite clause programs containing probabilistic facts with a parameterized distribution. It extends the traditional least Herbrand model semantics in logic programming to *distribution semantics*, possible world semantics with a probability distribution which is unconditionally applicable to arbitrary logic programs including ones for HMMs, PCFGs and Bayesian networks.

We also propose a new EM algorithm, the *graphical EM algorithm*, that runs for a class of parameterized logic programs representing sequential decision processes where each decision is exclusive and independent. It runs on a new data structure called *support graph*s describing the logical relationship between observations and their explanations, and learns parameters by computing inside and outside probability generalized for logic programs.

The complexity analysis shows that when combined with OLDT search for all explanations for observations, the graphical EM algorithm, despite its generality, has the same time complexity as existing EM algorithms, i.e. the Baum-Welch algorithm for HMMs, the Inside-Outside algorithm for PCFGs, and the one for singly connected Bayesian networks that have been developed independently in each research field. Learning experiments with PCFGs using two corpora of moderate size indicate that the graphical EM algorithm can significantly outperform the Inside-Outside algorithm.

## 1. Introduction

Parameter learning is common in various fields from neural networks to reinforcement learning to statistics. It is used to tune up systems for their best performance, be they classifiers or statistical models. Unlike these numerical systems described by mathematical formulas however, symbolic systems, typically programs, do not seem amenable to any kind of parameter learning. Actually there has been little literature on parameter learning of programs.

This paper is an attempt to incorporate parameter learning into computer programs. The reason is twofold. Theoretically we wish to add the ability of learning to computer programs, which the authors believe is a necessary step toward building intelligent systems. Practically it broadens the class of probability distributions, beyond traditionally used numerical ones, which are available for modeling complex phenomena such as gene inheritance, consumer behavior, natural language processing and so on.

The type of learning we consider here is statistical parameter learning applied to logic programs.[1] We assume that facts (unit clauses) in a program are probabilistically true and have a parameterized distribution.[2] Other clauses, non-unit definite clauses, are always true as they encode laws such as "if one has a pair of blood type genes `a` and `b`, one's blood type is `AB`". We call logic programs of this type a *parameterized logic program* and use for statistical modeling in which ground atoms[3] provable from the program represent our observations such as "one's blood type is `AB`" and the parameters of the program are inferred by performing ML (maximum likelihood) estimation on the observed atoms.

The probabilistic first-order framework sketched above is termed *statistical abduction* (Sato & Kameya, 2000) as it is an amalgamation of statistical inference and abduction where probabilistic facts play the role of *abducible*s, i.e. primitive hypotheses.[4] Statistical abduction is powerful in that it not only subsumes diverse symbolic-statistical frameworks such as HMMs (hidden Markov models, Rabiner, 1989), PCFGs (probabilistic context free grammars, Wetherell, 1980; Manning & Schütze, 1999) and (discrete) Bayesian networks (Pearl, 1988; Castillo, Gutierrez, & Hadi, 1997) but gives us freedom of using arbitrarily complex logic programs for modeling.[5]

The semantic basis for statistical abduction is *distribution semantics* introduced by Sato (1995). It defines a parameterized distribution, actually a probability measure, over the set of possible truth assignments to ground atoms and enables us to derive a new EM algorithm[6] for ML estimation called the *graphical EM algorithm* (Kameya & Sato, 2000).

Parameter learning in statistical abduction is done in two phases, search and EM learning. Given a parameterized logic program and observations, the first phase searches for all explanations for the observations. Redundancy in the first phase is eliminated by tabulating partial explanations using OLDT search (Tamaki & Sato, 1986; Warren, 1992; Sagonas, T., & Warren, 1994; Ramakrishnan, Rao, Sagonas, Swift, & Warren, 1995; Shen, Yuan, You, & Zhou, 2001). It returns a *support graph* which is a compact representation of the discovered explanations. In the second phase, we run the graphical EM algorithm on the support graph

1. In this paper, logic programs mean definite clause programs. A *definite clause program* is a set of definite clauses. A *definite clause* is a clause of the form $A \leftarrow L_1, \ldots, L_n$ $(0 \leq n)$ where $A, L_1, \ldots, L_n$ are atoms. $A$ is called the head, $L_1, \ldots, L_n$ the body. All variables are universally quantified. It reads if $L_1$ and $\cdots$ and $L_n$ hold, then $A$ holds. In case of $n = 0$, the clause is called a *unit clause*. A *general clause* is one whose body may contain negated atoms. A program including general clauses is sometimes called a general program (Lloyd, 1984; Doets, 1994).
2. Throughout this paper, for familiarity and readability, we will somewhat loosely use "distribution" as a synonym for "probability measure".
3. In logic programming, the adjective "ground" means no variables contained.
4. Abduction means inference to the best explanation for a set of observations. Logically, it is formalized as a search for an *explanation* $E$ such that $E, KB \vdash G$ where $G$ is an atom representing our observation, $KB$ a knowledge base and $E$ a conjunction of atoms chosen from *abducible*s, i.e. a class of formulas allowed as primitive hypotheses (Kakas, Kowalski, & Toni, 1992; Flach & Kakas, 2000). $E$ must be consistent with $KB$.
5. Existing symbolic-statistical modeling frameworks have restrictions and limitations of various types compared with arbitrary logic programs (see Section 7 for details). For example, Bayesian networks do not allow recursion. HMMs and PCFGs, stochastic grammars, allow recursion but lack variables and data structures. Recursive logic programs are allowed in Ngo and Haddawy's (1997) framework but they assume domains are finite and function symbols seem prohibited.
6. "EM algorithm" stands for a class of iterative algorithms for ML estimation with incomplete data (McLachlan & Krishnan, 1997).

and learn the parameters of the distribution associated with the program. Redundancy in the second phase is removed by the introduction of inside and outside probability for logic programs computed from the support graph.

The graphical EM algorithm has accomplished, when combined with OLDT search for all explanations, the same time complexity as the specialized ones, e.g. the Baum-Welch algorithm for HMMs (Rabiner, 1989) and the Inside-Outside algorithm for PCFGs (Baker, 1979), despite its generality. What is surprising is that, when we conducted learning experiments with PCFGs using real corpora, it outperformed the Inside-Outside algorithm by orders of magnitudes in terms of time for one iteration to update parameters. These experimental results enhance the prospect for symbolic-statistical modeling by parameterized logic programs of even more complex systems than stochastic grammars whose modeling has been difficult simply because of the lack of an appropriate modeling tool and their sheer complexities. The contributions of this paper therefore are

- distribution semantics for parameterized logic programs which unifies existing symbolic-statistical frameworks,
- the graphical EM algorithm (combined with tabulated search), a general yet efficient EM algorithm that runs on support graphs and
- the prospect suggested by the learning experiments for modeling and learning complex symbolic-statistical phenomena.

The rest of this paper is organized as follows. After preliminaries in Section 2, a probability space for parameterized logic programs is constructed in Section 3 as a mathematical basis for the subsequent sections. We then propose a new EM algorithm, the graphical EM algorithm, for parameterized logic programs in Section 4. Complexity analysis of the graphical EM algorithm is presented in Section 5 for HMMs, PCFGs, pseudo PCSGs and sc-BNs.[7] Section 6 contains experimental results of parameter learning with PCFGs by the graphical EM algorithm using real corpora that demonstrate the efficiency of the graphical EM algorithm. We state related work in Section 7, followed by conclusion in Section 8. The reader is assumed to be familiar with the basics of logic programming (Lloyd, 1984; Doets, 1994), probability theory (Chow & Teicher, 1997), Bayesian networks (Pearl, 1988; Castillo et al., 1997) and stochastic grammars (Rabiner, 1989; Manning & Schütze, 1999).

## 2. Preliminaries

Since our subject intersects logic programming and EM learning which are quite different in nature, we separate preliminaries.

### 2.1 Logic Programming and OLDT

In logic programming, a program *DB* is a set of definite clauses[8] and the execution is search for an *SLD refutation* of a given goal $\leftarrow G$. The top-down interpreter recursively selects the

7. Pseudo PCSGs (probabilistic context sensitive grammars) are a context-sensitive extension of PCFGs proposed by Charniak and Carroll (1994). sc-BN is a shorthand for a singly connected Bayesian network (Pearl, 1988).
8. We do not deal with general logic programs in this paper.

next goal and unfolds it (Tamaki & Sato, 1984) into subgoals using a nondeterministically chosen clause. The computed result by the SLD refutation, i.e. a *solution*, is an answer substitution (variable binding) $\theta$ such that $DB \vdash G\theta$.[9] Usually there is more than one refutation for $\leftarrow G$, and the search space for all refutations is described by an *SLD tree* which may be infinite depending on the program and the goal (Lloyd, 1984; Doets, 1994).

More often than not, applications require all solutions. In natural language processing for instance, a parser must be able to find all possible parse trees for a given sentence as every one of them is syntactically correct. Similarly in statistical abduction, we need to examine all explanations to determine the most likely one. All solutions are obtained by searching the entire SLD tree, and there is a choice of the search strategy. In Prolog, the standard logic programming language, backtracking is used to search for all solutions in conjunction with a fixed search order for goals (textually from left-to-right) and clauses (textually top-to-bottom) due to the ease and simplicity of implementation.

The problem with backtracking is that it forgets everything until up to the previous choice point, and hence it is quite likely to prove the same goal again and again, resulting in exponential search time. One answer to avoid this problem is to store computed results and reuse them whenever necessary. OLDT is such an instance of memoizing scheme (Tamaki & Sato, 1986; Warren, 1992; Sagonas et al., 1994; Ramakrishnan et al., 1995; Shen et al., 2001). Reuse of proved subgoals in OLDT search often drastically reduces search time for all solutions, especially when refutations of the top goal include many common sub-refutations. Take as an example a logic program coding an HMM. For a given string $s$, there exist exponentially many transition paths that output $s$. OLDT search applied to the program however only takes time linear in the length of $s$ to find all of them unlike exponential time by Prolog's backtracking search.

What does OLDT have to do with statistical abduction? From the viewpoint of statistical abduction, reuse of proved subgoals, or equivalently, structure sharing of sub-refutations for the top-goal $G$ brings about structure sharing of explanations for $G$, in addition to the reduction of search time mentioned above, thereby producing a highly compact representation of all explanations for $G$.

### 2.2 EM Learning

Parameterized distributions such as the multinomial distribution and the normal distribution provide convenient modeling devices in statistics. Suppose a random sample $x_1, \ldots, x_T$ of size $T$ on a random variable $X$ drawn from a distribution $P(X = x \mid \theta)$ parameterized by unknown $\theta$, is observed. The value of $\theta$ is determined by ML estimation as the MLE (maximum likelihood estimate) of $\theta$, i.e. as the maximizer of the likelihood $\prod_{1 \le i \le T} P(x_i \mid \theta)$.

Things get much more difficult when data are *incomplete*. Think of a probabilistic relationship between non-observable cause $X$ and observable effect $Y$ such as one between diseases and symptoms in medicine and assume that $Y$ does not uniquely determine the cause $X$. Then $Y$ is incomplete in the sense that $Y$ does not carry enough information to completely determine $X$. Let $P(X = x, Y = y \mid \theta)$ be a parameterized joint distribution over $X$ and $Y$. Our task is to perform ML estimation on $\theta$ under the condition that $X$ is

9. By a solution we ambiguously mean both the answer substitution $\theta$ itself and the proved atom $G\theta$, as one gives the other.

non-observable while $Y$ is observable. Let $y_1, \ldots, y_T$ be a random sample of size $T$ drawn from the marginal distribution $P(Y = y \mid \theta) = \sum_x P(X = x, Y = y \mid \theta)$. The MLE of $\theta$ is obtained by maximizing the likelihood $\prod_{1 \le i \le T} P(y_i \mid \theta)$ as a function of $\theta$.

While mathematical formulation looks alike in both cases, the latter, ML estimation with incomplete data, is far more complicated and direct maximization is practically impossible in many cases. People therefore looked to indirect approaches to tackle the problem of ML estimation with incomplete data to which the EM algorithm has been a standard solution (Dempster, Laird, & Rubin, 1977; McLachlan & Krishnan, 1997). It is an iterative algorithm applicable to a wide class of parameterized distributions including the multinomial distribution and the normal distribution such that the MLE computation is replaced by the iteration of two easier, more tractable steps. At $n$-th iteration, it first calculates the value of $Q$ function introduced below using current parameter value $\theta^{(n)}$ (E-step)[10] :

$$Q(\theta \mid \theta^{(n)}) \stackrel{\text{def}}{=} \sum_x P(x \mid y, \theta^{(n)}) \ln P(x, y \mid \theta). \tag{1}$$

Next, it maximizes $Q(\theta \mid \theta^{(n)})$ as a function of $\theta$ and updates $\theta^{(n)}$ (M-step):

$$\theta^{(n+1)} = \text{argmax}_{\theta} Q(\theta \mid \theta^{(n)}). \tag{2}$$

Since the old value $\theta^{(n)}$ and the updated value $\theta^{(n+1)}$ do not necessarily coincide, the E-steps and M-steps are iterated until convergence, during which the (log) likelihood is assured to increase monotonically (McLachlan & Krishnan, 1997).

Although the EM algorithm merely performs local maximization, it is used in a variety of settings due to its simplicity and relatively good performance. One must notice however that the EM algorithm is just a class name, taking different form depending on distributions and applications. The development of a concrete EM algorithm such as the Baum-Welch algorithm for HMMs (Rabiner, 1989) and the Inside-Outside algorithm for PCFGs (Baker, 1979) requires individual effort for each case.

10. $Q$ function is related to ML estimation as follows. We assume here only one data, $y$, is observed. From Jensen's inequality (Chow & Teicher, 1997) and the concavity of ln function, it follows that

$$\sum_x P(x \mid y, \theta^{(n)}) \ln P(x \mid y, \theta) - \sum_x P(x \mid y, \theta^{(n)}) \ln P(x \mid y, \theta^{(n)}) \le 0$$

and hence that

$$\begin{aligned} & Q(\theta \mid \theta^{(n)}) - Q(\theta^{(n)} \mid \theta^{(n)}) \\ = \; & \sum_x P(x \mid y, \theta^{(n)}) \ln P(x \mid y, \theta) - \sum_x P(x \mid y, \theta^{(n)}) \ln P(x \mid y, \theta^{(n)}) + \ln P(y \mid \theta) - \ln P(y \mid \theta^{(n)}) \\ \le \; & \ln P(y \mid \theta) - \ln P(y \mid \theta^{(n)}). \end{aligned}$$

Consequently, we have

$$Q(\theta \mid \theta^{(n)}) \ge Q(\theta^{(n)} \mid \theta^{(n)}) \;\Rightarrow\; \ln p(y \mid \theta) \ge \ln p(y \mid \theta^{(n)}) \;\Rightarrow\; p(y \mid \theta) \ge p(y \mid \theta^{(n)}).$$

## 3. Distribution Semantics

In this section, we introduce *parameterized logic programs* and define their declarative semantics. The basic idea is as follows. We start with a set $F$ of probabilistic facts (atoms) and a set $R$ of non-unit definite clauses. Sampling from $F$ determines a set $F'$ of true atoms, and the least Herbrand model of $F' \cup R$ determines the truth value of *every* atom in $DB = F \cup R$. Hence every atom can be considered as a random variable, taking on 1 (true) or 0 (false). In what follows, we formalize this process and construct the underlying probability space for the denotation of $DB$.

### 3.1 Basic Distribution $P_F$

Let $DB = F \cup R$ be a definite clause program in a first-order language $\mathcal{L}$ with countably many variables, function symbols and predicate symbols where $F$ is a set of unit clauses (*facts*) and $R$ a set of non-unit clauses (*rules*). In the sequel, unless otherwise stated, we consider for simplicity $DB$ as the set of all ground instances of the clauses in $DB$, and assume that $F$ and $R$ consist of countably infinite ground clauses (the finite case is similarly treated). We then construct a probability space for $DB$ in two steps. First we introduce a probability space over the *Herbrand interpretations*[11] of $F$ i.e. the truth assignments to ground atoms in $F$. Next we extend it to a probability space over the Herbrand interpretations of *all* ground atoms in $\mathcal{L}$ by using the least model semantics (Lloyd, 1984; Doets, 1994).

Let $A_1, A_2, \ldots$ be a fixed enumeration of atoms in $F$. We regard an infinite vector $\omega = \langle x_1, x_2, \ldots \rangle$ of 0s and 1s as a Herbrand interpretation of $F$ in such a way that for $i = 1, 2, \ldots$ $A_i$ is true (resp. false) if and only if $x_i = 1$ (resp. $x_i = 0$). Under this isomorphism, the set of all possible Herbrand interpretations of $F$ coincides with the Cartesian product:

$$\Omega_F \stackrel{\text{def}}{=} \prod_{i=1}^{\infty} \{0,1\}_i.$$

We construct a probability measure $P_F$ over the sample space $\Omega_F$[12] from a collection of finite joint distributions $P_F^{(n)}(A_1 = x_1, \ldots, A_n = x_n)$ $(n = 1, 2, \ldots, x_i \in \{0,1\}, 1 \leq i \leq n)$ such that

$$\left\{ \begin{array}{l} 0 \leq P_F^{(n)}(A_1 = x_1, \ldots, A_n = x_n) \leq 1 \\ \sum_{x_1,\ldots,x_n} P_F^{(n)}(A_1 = x_1, \ldots, A_n = x_n) = 1 \\ \sum_{x_{n+1}} P_F^{(n+1)}(A_1 = x_1, \ldots, A_{n+1} = x_{n+1}) = P_F^{(n)}(A_1 = x_1, \ldots, A_n = x_n). \end{array} \right. \tag{3}$$

The last equation is called the *compatibility condition*. It can be proved (Chow & Teicher, 1997) from the compatibility condition that there exists a probability space $(\Omega_F, \mathcal{F}, P_F)$ where $P_F$ is a probability measure on $\mathcal{F}$, the minimal $\sigma$ algebra containing open sets of $\Omega_F$, such that for any $n$,

$$P_F(A_1 = x_1, \ldots, A_n = x_n) = P_F^{(n)}(A_1 = x_1, \ldots, A_n = x_n).$$

11. A *Herbrand interpretation* interprets a function symbol uniquely as a function on ground terms and assigns truth values to ground atoms. Since the interpretation of function symbols is common to all Herbrand interpretations, given $\mathcal{L}$, they have a one-to-one correspondence with truth assignments to ground atoms in $\mathcal{L}$. So we do not distinguish them.
12. We regard $\Omega_F$ as a topological space with the product topology such that each $\{0,1\}$ is equipped with the discrete topology.

We call $P_F$ a *basic distribution.*[13]

The choice of $P_F^{(n)}$ is free as long as the compatibility condition is met. If we want all interpretations to be equiprobable, we should set $P_F^{(n)}(A_1 = x_1, \ldots, A_n = x_n) = 1/2^n$ for every $\langle x_1, \ldots, x_n \rangle$. The resulting $P_F$ is a uniform distribution over $\Omega_F$ just like the one over the unit interval $[0, 1]$. If, on the other hand, we stipulate no interpretation except $\omega_0 = \langle c_1, c_2, \ldots \rangle$ should be possible, we put, for each $n$,

$$P_F^{(n)}(A_1 = x_1, \ldots, A_n = x_n) = \begin{cases} 1 & \text{if} \quad \forall i \; x_i = c_i \; (1 \le i \le n) \\ 0 & \text{o.w.} \end{cases}$$

Then $P_F$ places all probability mass on $\omega_0$ and gives probability 0 to the rest.

Define a *parameterized logic program* as a definite clause program[14] $DB = F \cup R$ where $F$ is a set of unit clauses, $R$ is a set of non-unit clauses such that no clause head in $R$ is unifiable with a unit clause in $F$ and a parameterized basic distribution $P_F$ is associated with $F$. A parameterized $P_F$ is obtained from a collection of parameterized joint distributions satisfying the compatibility condition. Generally, the more complex $P_F^{(n)}$'s are, the more flexible $P_F$ is, but at the cost of tractability. The choice of parameterized finite distributions made by Sato (1995) was simple:

$$\begin{aligned} & P_F^{(2n)}(ON_1 = x_1, OFF_2 = x_2, \ldots, OFF_{2n} = x_{2n} \mid \theta_1, \ldots, \theta_n) \\ = & \prod_{i=1}^{n} P_{bs}(ON_{2i-1} = x_{2i-1}, OFF_{2i} = x_{2i} \mid \theta_i) \end{aligned}$$

where

$$\begin{aligned} & P_{bs}(ON_{2i-1} = x_{2i-1}, OFF_{2i} = x_{2i} \mid \theta_i) \\ = & \begin{cases} 0 & \text{if } x_{2i-1} = x_{2i} \\ \theta_i & \text{if } x_{2i-1} = 1, \; x_{2i} = 0 \\ 1 - \theta_i & \text{if } x_{2i-1} = 0, \; x_{2i} = 1. \end{cases} \end{aligned} \tag{4}$$

$P_{bs}(ON_{2i-1} = x_{2i-1}, OFF_{2i} = x_{2i} \mid \theta_i)$ $(1 \le i \le n)$ represents a probabilistic binary switch, i.e. a Bernoulli trial, using two exclusive atoms $ON_{2i-1}$ and $OFF_{2i}$ in such a way that either one of them is true on each trial but never both. $\theta_i$ is a parameter specifying the probability that the switch $i$ is on. The resulting $P_F$ is a probability measure over the infinite product of independent binary outcomes. It might look too simple but expressive enough for Bayesian networks, Markov chains and HMMs (Sato, 1995; Sato & Kameya, 1997).

### 3.2 Extending $P_F$ to $P_{DB}$

In this subsection, we extend $P_F$ to a probability measure $P_{DB}$ over the *possible worlds* for $\mathcal{L}$, i.e. the set of all possible truth assignments to ground atoms in $\mathcal{L}$ through the least

13. This naming of $P_F$, despite its being a probability measure, partly reflects the observation that it behaves like an infinite joint distribution $P_F(A_1 = x_1, A_2 = x_2, \ldots)$ for an infinite random vector $\langle A_1, A_2, \ldots \rangle$ of which $P_F^{(n)}(A_1 = x_1, \ldots, A_n = x_n)$ $(n = 1, 2, \ldots)$ are marginal distributions. Another reason is intuitiveness. These considerations apply to $P_{DB}$ defined in the next subsection as well.
14. Here clauses are not necessarily ground.

Herbrand model (Lloyd, 1984; Doets, 1994). Before proceeding however, we need a couple of notations. For an atom $A$, define $A^x$ by

$$\begin{cases} A^x = A & \text{if} \quad x = 1 \\ A^x = \neg A & \text{if} \quad x = 0. \end{cases}$$

Next take a Herbrand interpretation $\nu \in \Omega_F$ of $F$. It makes some atoms in $F$ true and others false. Let $F_\nu$ be the set of atoms made true by $\nu$. Then imagine a definite clause program $DB' = R \cup F_\nu$ and its least Herbrand model $M_{DB'}$ (Lloyd, 1984; Doets, 1994). $M_{DB'}$ is characterized as the least fixed point of a mapping $T_{DB'}(\cdot)$ below

$$T_{DB'}(I) \stackrel{\text{def}}{=} \left\{ A \;\middle|\; \begin{array}{l} \text{there is some } A \leftarrow B_1, \ldots, B_k \in DB' \ (0 \leq k) \\ \text{such that } \{B_1, \ldots, B_k\} \subseteq I \end{array} \right\}$$

where $I$ is a set of ground atoms.[15] Or equivalently, it is inductively defined by

$$\begin{aligned} I_0 &= \emptyset \\ I_{n+1} &= T_{DB'}(I_n) \\ M_{DB'} &= \bigcup_n I_n. \end{aligned}$$

Taking into account the fact that $M_{DB'}$ is a function of $\nu \in \Omega_F$, we henceforth employ a functional notation $M_{DB}(\nu)$ to denote $M_{DB'}$.

Turning back, let $A_1, A_2, \ldots$ be again an enumeration, but of *all* ground atoms in $\mathcal{L}$.[16] Form $\Omega_{DB}$, similarly to $\Omega_F$, as the Cartesian product of denumerably many $\{0,1\}$'s and identify it with the set of all possible Herbrand interpretations of the ground atoms $A_1, A_2, \ldots$ in $\mathcal{L}$, i.e. the *possible worlds* for $\mathcal{L}$. Then extend $P_F$ to a probability measure $P_{DB}$ over $\Omega_{DB}$ as follows. Introduce a series of finite joint distributions $P_{DB}^{(n)}(A_1 = x_1, \ldots, A_n = x_n)$ for $n = 1, 2, \ldots$ by

$$\begin{aligned} [A_1^{x_1} \wedge \cdots \wedge A_n^{x_n}]_F &\stackrel{\text{def}}{=} \{\nu \in \Omega_F \mid M_{DB}(\nu) \models A_1^{x_1} \wedge \cdots \wedge A_n^{x_n}\} \\ P_{DB}^{(n)}(A_1 = x_1, \ldots, A_n = x_n) &\stackrel{\text{def}}{=} P_F([A_1^{x_1} \wedge \cdots \wedge A_n^{x_n}]_F). \end{aligned}$$

Note that the set $[A_1^{x_1} \wedge \cdots \wedge A_n^{x_n}]_F$ is $P_F$-measurable and by definition, $P_{DB}^{(n)}$'s satisfy the compatibility condition

$$\sum_{x_{n+1}} P_{DB}^{(n+1)}(A_1 = x_1, \ldots, A_{n+1} = x_{n+1}) = P_{DB}^{(n)}(A_1 = x_1, \ldots, A_n = x_n).$$

Hence there exists a probability measure $P_{DB}$ over $\Omega_{DB}$ which is an extension of $P_F$ such that

$$P_{DB}(A_1 = x_1, \ldots, A_n = x_n) = P_F(A_1 = x_1, \ldots, A_n = x_n)$$

15. $I$ defines, mutually, a Herbrand interpretation such that a ground atom $A$ is true if and only if $A \in I$. A *Herbrand model* of a program is a Herbrand interpretation that makes every ground instance of every clause in the program true.
16. Note that this enumeration enumerates ground atoms in $F$ as well.

for any finite atoms $A_1, \ldots, A_n$ in $F$ and for every binary vector $\langle x_1, \ldots, x_n \rangle$ ($x_i \in \{0, 1\}$, $1 \leq i \leq n$). Define the *denotation* of the program $DB = F \cup R$ w.r.t. $P_F$ to be $P_{DB}$. The denotational semantics of parameterized logic programs defined above is called *distribution semantics*. As remarked before, we regard $P_{DB}$ as a kind of infinite joint distribution $P_{DB}(A_1 = x_1, A_2 = x_2, \ldots)$. Mathematical properties of $P_{DB}$ are listed in Appendix A where our semantics is proved to be an extension of the standard least model semantics in logic programming to possible world semantics with a probability measure.

### 3.3 Programs as Distributions

Distribution semantics views parameterized logic programs as expressing distributions. Traditionally distributions have been expressed by using mathematical formulas but the use of programs as (discrete) distributions gives us far more freedom and flexibility than mathematical formulas in the construction of distributions because they have *recursion* and arbitrary composition. In particular a program can contain infinitely many random variables as probabilistic atoms through recursion, and hence can describe stochastic processes that potentially involve infinitely many random variables such as Markov chains and derivations in PCFGs (Manning & Schütze, 1999).[17]

Programs also enable us to *procedurally* express complicated constraints on distributions such as "the sum of occurrences of alphabets `a` or `b` in an output string of an HMM must be a multiple of three". This feature, procedural expression of arbitrarily complex (discrete) distributions, seems quite helpful in symbolic-statistical modeling.

Finally, providing mathematically sound semantics for parameterized logic programs is one thing, and implementing distribution semantics in a tractable way is another. In the next section, we investigate conditions on parameterized logic programs which make probability computation tractable, thereby making them usable as a means for large scale symbolic-statistical modeling.

## 4. Graphical EM Algorithm

According to the preceding section, a parameterized logic program $DB = F \cup R$ in a first-order language $\mathcal{L}$ with a parameterized basic distribution $P_F(\cdot \mid \theta)$ over the Herbrand interpretations of ground atoms in $F$ specifies a parameterized distribution $P_{DB}(\cdot \mid \theta)$ over the Herbrand interpretations for $\mathcal{L}$. In this section, we develop, step by step, an efficient EM algorithm for the parameter learning of parameterized logic programs by interpreting $P_{DB}$ as a distribution over the observable and non-observable events. The new EM algorithm is termed the *graphical EM algorithm*. It is applicable to arbitrary logic programs satisfying certain conditions described later provided the basic distribution is a direct product of multi-ary random switches, which is a slight complication of the binary ones introduced in Section 3.1.

From this section on, we assume that $DB$ consists of usual definite clauses containing (universally quantified) variables. Definitions and changes relating to this assumption are

17. An infinite derivation can occur in PCFGs. Take a simple PCFG $\{p : S \rightarrow a, q : S \rightarrow SS\}$ where $S$ is a start symbol, $a$ a terminal symbol, $p + q = 1$ and $p, q > 0$. In this PCFG, $S$ is rewritten either to $a$ with probability $p$ or to $SS$ with probability $q$. The probability of the occurrence of an infinite derivation is calculated as $\max\{0, 1 - (p/q)\}$ which is non-zero when $q > p$ (Chi & Geman, 1998).

listed below. For a predicate $p$, we introduce $\mathrm{iff}(p)$, the *iff definition of* $p$ by

$$\mathrm{iff}(p) \stackrel{\text{def}}{=} \forall x\ \left(p(x) \leftrightarrow \exists y_1(x = t_1 \wedge W_1) \vee \cdots \vee \exists y_n(x = t_n \wedge W_n)\right).$$

Here $x$ is a vector of new variables of length equal to the arity of $p$, $p(t_i) \leftarrow W_i$ $(1 \leq i \leq n, 0 \leq n)$, an enumeration of clauses about $p$ in $DB$, and $y_i$, a vector of variables occurring in $p(t_i) \leftarrow W_i$. Then define $comp(R)$ as follows.

$$\begin{aligned}
head(R) &\stackrel{\text{def}}{=} \{B \mid B \text{ is a ground instance of a clause head appearing in } R\} \\
\mathrm{iff}(R) &\stackrel{\text{def}}{=} \{\mathrm{iff}(p) \mid p \text{ appears in a clause head in } R\} \\
E_q &\stackrel{\text{def}}{=} \{f(x) = f(y) \rightarrow x = y \mid f \text{ is a function symbol}\} \\
&\qquad \cup \{f(x) \neq g(y) \mid f \text{ and } g \text{ are different function symbols}\} \\
&\qquad \cup \{t \neq x \mid t \text{ is a term properly containing } x\} \\
comp(R) &\stackrel{\text{def}}{=} \mathrm{iff}(R) \cup E_q
\end{aligned}$$

$E_q$, Clark's equational theory (Clark, 1978), deductively simulates unification. Likewise $comp(R)$ is a first-order theory which deductively simulates SLD refutation with the help of $E_q$ by replacing a clause head atom with the clause body (Lloyd, 1984; Doets, 1994).

We here introduce some definitions which will be frequently used. Let $B$ be an atom. An *explanation for* $B$ w.r.t. $DB = F \cup R$ is a conjunction $S$ such that $S, R \vdash B$, and as a set comprised of its conjuncts, $S \subset F$ holds and no proper subset of $S$ satisfies this. The set of all explanations for $B$ is called the *support set for* $B$ and designated by $\psi_{DB}(B)$.[18]

### 4.1 Motivating Example

First of all, we review distribution semantics by a concrete example. Consider the following program $DB_b = F_b \cup R_b$ in Figure 1 modeling how one's blood type is determined by blood type genes probabilistically inherited from the parents.[19]

The first four clauses in $R_b$ state a blood type is determined by a genotype, i.e. a pair of blood type genes `a`, `b` and `o`. For instance, `btype('A'):- (gtype(a,a) ; gtype(a,o) ; gtype(o,a))` says that one's blood type is `A` if his (her) genotype is ⟨`a`,`a`⟩, ⟨`a`,`o`⟩ or ⟨`o`,`a`⟩. These are propositional rules.

Succeeding clauses state general rules in terms of logical variables. The fifth clause says that regardless of the values of `X` and `Y`, event `gtype(X,Y)` (one's having genotype ⟨`X`,`Y`⟩) is caused by two events, `gene(father,X)` (inheriting gene `X` from the father) and `gene(mother,Y)` (inheriting gene `Y` from the mother). `gene(P,G):- msw(gene,P,G)` is a clause connecting rules in $R_b$ with probabilistic facts in $F_b$. It tells us that the gene `G` is inherited from a parent `P` if a choice represented by `msw(gene,P,G)`[20] is made. The

18. This definition of a support set differs from the one used by Sato (1995) and Kameya and Sato (2000).

19. When we implicitly emphasize the procedural reading of logic programs, Prolog conventions are employed (Sterling & Shapiro, 1986). Thus, `;` stands for "or", `,` "and" `:-` "implied by" respectively. Strings beginning with a capital letter are (universally quantified) variables, but quoted ones such as `'A'` are constants. The underscore `_` is an anonymous variable.

20. `msw` is an abbreviation of "multi-ary random switch" and `msw(·,·,·)` expresses a probabilistic choice from finite alternatives. In the framework of statistical abduction, `msw` atoms are abducibles from which explanations are constructed as a conjunction.

$$R_b = \left\{ \begin{array}{lll} \texttt{btype('A')} & \texttt{:-} & \texttt{(gtype(a,a) ; gtype(a,o) ; gtype(o,a)).} \\ \texttt{btype('B')} & \texttt{:-} & \texttt{(gtype(b,b) ; gtype(b,o) ; gtype(o,b)).} \\ \texttt{btype('O')} & \texttt{:-} & \texttt{gtype(o,o).} \\ \texttt{btype('AB')} & \texttt{:-} & \texttt{(gtype(a,b) ; gtype(b,a)).} \\ \texttt{gtype(X,Y)} & \texttt{:-} & \texttt{gene(father,X), gene(mother,Y).} \\ \texttt{gene(P,G)} & \texttt{:-} & \texttt{msw(gene,P,G).} \end{array} \right.$$

$$\begin{array}{rcl} F_b & = & \{\texttt{msw(gene,father,a)},\texttt{msw(gene,father,b)},\texttt{msw(gene,father,o)}, \\ & & \texttt{msw(gene,mother,a)},\texttt{msw(gene,mother,b)},\texttt{msw(gene,mother,o)}\} \end{array}$$

Figure 1: ABO blood type program $DB_b$

genetic knowledge that the choice of G is by chance and made from $\{\texttt{a},\texttt{b},\texttt{o}\}$ is expressed by specifying a joint distribution $F_b$ as follows.

$$P_{F_b}(\texttt{msw(gene,}t\texttt{,a)} = x, \texttt{msw(gene,}t\texttt{,b)} = y, \texttt{msw(gene,}t\texttt{,o)} = z \mid \theta_a, \theta_b, \theta_o) \stackrel{\text{def}}{=} \theta_a^x \theta_b^y \theta_o^z$$

where $x, y, z \in \{0,1\}$, $x + y + z = 1$, $\theta_a, \theta_b, \theta_o \in [0,1]$, $\theta_a + \theta_b + \theta_o = 1$ and $t$ is either `father` or `mother`. Thus $\theta_a$ is the probability of inheriting gene `a` from a parent. Statistical independence of the choice of gene, once from `father` and once from `mother`, is expressed by putting

$$\begin{array}{l} P_{F_b}(\, \texttt{msw(gene,father,a)} = x, \texttt{msw(gene,father,b)} = y, \texttt{msw(gene,father,o)} = z, \\ \qquad \texttt{msw(gene,mother,a)} = x', \texttt{msw(gene,mother,b)} = y', \texttt{msw(gene,mother,o)} = z' \\ \qquad \mid \theta_a, \theta_b, \theta_o) \\ = \quad P_{F_b}(x, y, z \mid \theta_a, \theta_b, \theta_o) P_{F_b}(x', y', z' \mid \theta_a, \theta_b, \theta_o). \end{array}$$

In this setting, atoms representing our observation are $obs(DB_b) = \{$`btype('A')`, `btype('B')`, `btype('O')`, `btype('AB')`$\}$. We observe one of them, say `btype('A')`, and infer a possible explanation $S$, i.e. a minimal conjunction of abducibles $\texttt{msw(gene,}\cdot\texttt{,}\cdot\texttt{)}$ such that

$$S, R_b \vdash \texttt{btype('A')}.$$

$S$ is obtained by applying a special SLD refutation procedure to the goal $\leftarrow$ `btype('A')` which preserves `msw` atoms resolved upon in the refutation. Three explanations are found.

$$\begin{array}{rcl} S_1 & = & \texttt{msw(gene,father,a)} \wedge \texttt{msw(gene,mother,a)} \\ S_2 & = & \texttt{msw(gene,father,a)} \wedge \texttt{msw(gene,mother,o)} \\ S_3 & = & \texttt{msw(gene,father,o)} \wedge \texttt{msw(gene,mother,a)} \end{array}$$

So $\psi_{DB_b}(\texttt{btype(a)})$, the support set for `btype(a)`, is $\{S_1, S_2, S_3\}$. The probability of each explanation is respectively computed as $P_{F_b}(S_1) = \theta_a^2$ and $P_{F_b}(S_2) = P_{F_b}(S_3) = \theta_a\theta_o$. From Proposition A.2 in Appendix A, it follows that $P_{DB_b}(\texttt{btype('A')}) = P_{DB_b}(S_1 \vee S_2 \vee S_3) = P_{F_b}(S_1 \vee S_2 \vee S_3)$ and that

$$\begin{array}{rcl} P_{DB_b}(\texttt{btype('A')} \mid \theta_a, \theta_b, \theta_o) & = & P_{F_b}(S_1) + P_{F_b}(S_2) + P_{F_b}(S_3) \\ & = & \theta_a^2 + 2\theta_a\theta_o. \end{array}$$

Here we used the fact that $S_1$, $S_2$ and $S_3$ are mutually exclusive as the choice of gene is exclusive. Parameters, i.e. $\theta_a$, $\theta_b$ and $\theta_o$ are determined by ML estimation performed on a random sample such as $\{$`btype('A'),btype('O'),btype('AB')`$\}$ of `btype` as follows.

$$\begin{aligned}\langle\theta_a,\theta_b,\theta_o\rangle &= \mathrm{argmax}_{\langle\theta_a,\theta_b,\theta_o\rangle}\ P_{DB_b}(\texttt{btype('A')})P_{DB_b}(\texttt{btype('O')})P_{DB_b}(\texttt{btype('AB')}) \\ &= \mathrm{argmax}_{\langle\theta_a,\theta_b,\theta_o\rangle}\ (\theta_a^2+2\theta_a\theta_o)\theta_o^2\theta_a\theta_b\end{aligned}$$

This program contains neither function symbol nor recursion though our semantics allows for them. Later we see an example containing both, a program for an HMM (Rabiner & Juang, 1993).

## 4.2 Four Simplifying Conditions

$DB_b$ in Figure 1 is simple and probability computation is easy. This is not generally the case. Since our primary interest is learning, especially efficient parameter learning of parameterized logic programs, we hereafter concentrate on identifying what property of a program makes probability computation easy like $DB_b$, thereby makes efficient parameter learning possible.

To answer this question precisely, let us formulate the whole modeling process. Suppose there exist symbolic-statistical phenomena such as gene inheritance for which we hope to construct a probabilistic computational model. We first specify a *target predicate* $p$ whose ground atom $p(s)$ represents our observation of the phenomena. Then to explain the empirical distribution of $p$, we write down a parameterized logic program $DB = F \cup R$ having a basic distribution $P_F$ with parameter $\theta$ that can reproduce all observable patterns of $p(s)$. Finally, observing a random sample $p(s_1),\ldots,p(s_T)$ of ground atoms of $p$, we adjust $\theta$ by ML estimation, i.e. by maximizing the likelihood $L(\theta) = \prod_{t=1}^{T} P_{DB}(p(s_t) \mid \theta)$ so that $P_{DB}(p(\cdot) \mid \theta)$ approximates as closely to the empirically observed distribution of $p$ as possible.

At first sight, this formulation looks right, but in reality it is not. Suppose two events $p(s)$ and $p(s')$ $(s \neq s')$ are observed. We put $L(\theta) = P_{DB}(p(s) \mid \theta)P_{DB}(p(s') \mid \theta)$. But this *cannot be* a likelihood at all simply because in distribution semantics, $p(s)$ and $p(s')$ are two different random variables, not two realizations of the same random variable.

A quick remedy is to note that in the case of blood type program $DB_b$ where $obs(DB_b) =$ $\{$`btype('A'),btype('B'),btype('O'),btype('AB')`$\}$ are observable atoms, only one of them is true for each observation, and if some atom is true, others must be false. In other words, these atoms collectively behave as a single random variable having the distribution $P_{DB_b}$ whose values are $obs(DB_b)$.

Keeping this in mind, we introduce the following condition. Let $obs(DB)$ $(\subset head(R))$ be a set of ground atoms which represent observable events. We call them *observable atoms*.

**Uniqueness condition:**

$$P_{DB}(G \wedge G') = 0 \text{ for any } G \neq G' \in obs(DB), \quad \text{and} \quad \sum_{G \in obs(DB)} P_{DB}(G) = 1.$$

The uniqueness condition enables us to introduce a new random variable $Y_o$ representing our observation. Fix an enumeration $G_1, G_2, \ldots$ of observable atoms in $obs(DB)$ and define $Y_o$ by[21]

$$Y_o(\omega) = k \quad \text{iff} \quad \omega \models G_k \ \text{ for } \omega \in \Omega_{DB} \ (k \geq 1). \tag{5}$$

Let $G_{k_1}, G_{k_2}, \ldots, G_{k_T} \in obs(DB)$ be a random sample of size $T$. Then $L(\theta) = \prod_{t=1}^{T} P_{DB}(G_{k_t} \mid \theta) = \prod_{t=1}^{T} P_{DB}(Y_o = k_t \mid \theta)$ qualifies for the likelihood function w.r.t. $Y_o$.

The second condition concerns the reduction of probability computation to addition. Take again the blood type exmaple. The computation of $P_{DB_b}$(`btype('A')`) is decomposed into a summation because explanations in the support set are mutualy exclusive. So we introduce

**Exclusiveness condition:**

For every $G \in obs(DB)$ and the support set $\psi_{DB}(G)$, $P_{DB}(S \wedge S') = 0$ for any $S \neq S' \in \psi_{DB}(G)$.

Using the exclusiveness condition (and Proposition A.2 in Appendix A), we have

$$P_{DB}(G) = \sum_{S \in \psi_{DB}(G)} P_F(S).$$

From a modeling point of view, it means that while a single event, or a single observation, $G$, may have several (or even infinite) explanations $\psi_{DB}(G)$, only one of $\psi_{DB}(G)$ is allowed to be true for each observation.

Now introduce $\Psi_{DB}$, i.e. the set of all explanations relevant to $obs(DB)$ by

$$\Psi_{DB} \stackrel{\text{def}}{=} \bigcup_{G \in obs(DB)} \psi_{DB}(G)$$

and fix an enumeration $S_1, S_2, \ldots$ of explanations in $\Psi_{DB}$. It follows from Proposition A.2, the uniqueness condition and the exclusiveness condition that

$$\begin{aligned} P_{DB}(S_i \wedge S_j) &= 0 \ \text{ for } i \neq j \qquad \text{and} \\ \sum_{S \in \Psi_{DB}} P_{DB}(S) &= \sum_{G \in obs(DB)} \sum_{S \in \psi_{DB}(G)} P_{DB}(S) \\ &= \sum_{G \in obs(DB)} P_{DB}(G) \\ &= 1. \end{aligned}$$

So we are able to introduce under the uniqueness condition and the exclusiveness condition yet another random variable $X_e$, representing an explanation for $G$, defined by

$$X_e(\omega) = k \quad \text{iff} \quad \omega \models S_k \ \text{ for } \omega \in \Omega_{DB}. \tag{6}$$

The third condition concerns termination.

21. $\sum_{G \in obs(DB)} P_{DB}(G) = 1$ only guarantees that the measure of $\{\omega \mid \omega \models G_k \text{ for some } k\,(\geq 1)\}$ is one, so there can be some $\omega$ satisfying no $G_k$'s. In such case, we put $Y_o(\omega) = 0$. But values on a set of measure zero do not affect any part of the discussion that follows. This also applies to the definition of $X_e$ in (6).

**Finite support condition:**

For every $G \in obs(DB)$ $\psi_{DB}(G)$ is finite.

$P_{DB}(G)$ is then computed from the support set $\psi_{DB}(G) = \{S_1, \ldots, S_m\}$ $(0 \le m)$, with the help of the exclusiveness condition, as a finite summation $\sum_{i=1}^{m} P_F(S_i)$. This condition prevents an infinite summation that is hardly computable.

The fourth condition simplifies the probability computation to multiplication. Recall that an explanation $S$ for $G \in obs(DB)$ is a conjunction $a_1 \wedge \cdots \wedge a_m$ of some abducibles $\{a_1, \ldots, a_m\} \subset F$ $(1 \le m)$. In order to reduce the computation of $P_F(S) = P_F(a_1 \wedge \cdots \wedge a_m)$ to the multiplication $P_F(a_1) \cdots P_F(a_m)$, we assume

**Distribution condition:**

$F$ is a set $F_{\texttt{msw}}$ of ground atoms with a parameterized distribution $P_{\texttt{msw}}$ specified below.

Here atom $\texttt{msw}(i,n,v)$ is intended to simulate a multi-ary random switch whose name is $i$ and whose outcome is $v$ on trial $n$. It is a generalization of primitive probabilistic events such as coin tossing and dice rolling.

1. $F_{\texttt{msw}}$ consists of probabilistic atoms $\texttt{msw}(i,n,v)$. The arguments $i$, $n$ and $v$ are ground terms called *switch name*, *trial-id* and a *value* (of the switch $i$), respectively. We assume that a finite set $V_i$ of ground terms called the *value set of* $i$ is associated with each $i$, and $v \in V_i$ holds.

2. Write $V_i$ as $\{v_1, v_2, \ldots, v_m\}$ $(m = |V_i|)$. Then, one of the ground atoms $\{\texttt{msw}(i,n,v_1), \texttt{msw}(i,n,v_2), \ldots, \texttt{msw}(i,n,v_m)\}$ becomes exclusively true (takes on value 1) on each trial. With each $i$, a *parameter* $\theta_{i,v} \in [0,1]$ such that $\sum_{v \in V_i} \theta_{i,v} = 1$ is associated. $\theta_{i,v}$ is the probability of $\texttt{msw}(i,\cdot,v)$ being true $(v \in V_i)$.

3. For each ground terms $i$, $i'$, $n$, $n'$, $v \in V_i$ and $v' \in V_{i'}$, random variable $\texttt{msw}(i,n,v)$ is independent of $\texttt{msw}(i',n',v')$ if $n \neq n'$ or $i \neq i'$.

In other words, we introduce a family of parameterized finite distributions $P_{(i,n)}$ such that

$$P_{(i,n)}(\texttt{msw}(i,n,v_1) = x_1, \ldots, \texttt{msw}(i,n,v_m) = x_m \mid \theta_{i,v_1}, \ldots, \theta_{i,v_m}) \stackrel{\text{def}}{=} \begin{cases} \theta_{i,v_1}^{x_1} \cdots \theta_{i,v_m}^{x_m} & \text{if } \sum_{k=1}^{m} x_k = 1 \\ 0 & \text{o.w.} \end{cases} \tag{7}$$

where $m = |V_i|$, $x_k \in \{0,1\}$ $(1 \le k \le m)$, and define $P_{\texttt{msw}}$ as their infinite product

$$P_{\texttt{msw}} \stackrel{\text{def}}{=} \prod_{i,n} P_{(i,n)}.$$

Under this condition, we can compute $P_{\texttt{msw}}(S)$, the probability of an explanation $S$, as the product of parameters. Suppose $\texttt{msw}(i_j,n,v)$ and $\texttt{msw}(i_{j'},n',v')$ are different conjuncts in an explanation $S = \texttt{msw}(i_1,n_1,v_1) \wedge \cdots \wedge \texttt{msw}(i_k,n_k,v_k)$. If either $j \neq j'$ or $n \neq n'$ holds, they are independent by construction. Else if $j = j'$ and $n = n'$ but $v \neq v'$, they are not independent but $P_{\texttt{msw}}(S) = 0$ by construction. As a result, whichever condition may hold, $P_{\texttt{msw}}(S)$ is computed from the parameters.

### 4.3 Modeling Principle

Up to this point, we have introduced four conditions, the uniqueness condition, the exclusiveness condition, the finite support condition and the distribution condition, to simplify probability computation. The last one is easy to satisfy. We just adopt $F_{\mathtt{msw}}$ together with $P_{\mathtt{msw}}$. So, from here on, we always assume that $F_{\mathtt{msw}}$ has a parameterized distribution $P_{\mathtt{msw}}$ introduced in the previous subsection. Unfortunately the rest are not satisfied automatically. According to our modeling experiences however, it is only mildly difficult to satisfy the uniqueness condition and the exclusiveness condition as long as we obey the following modeling principle.

> **Modeling principle:** $DB = F_{\mathtt{msw}} \cup R$ describes a sequential decision process (modulo auxiliary computations) that uniquely produces an observable atom $G \in obs(DB)$ where each decision is expressed by some `msw` atom.[22]

Translated into programming level, it says that we must take care when writing a program so that *for any sample $F'$ from $P_{\mathtt{msw}}$, there must uniquely exist goal $\leftarrow G$ ($G \in obs(DB)$) which has a successful refutation from $DB' = F' \cup R$.* We can confirm the principle by the blood type program $DB_b = F_b \cup R_b$. It describes a process of gene inheritance, and for an arbitrary sample $F'_b$ from $P_{\mathtt{msw}}$, say $F'_b = \{$`msw(gene,father,a)`,`msw(gene,mother,o)`$\}$, there exists a unique goal, $\leftarrow$`btype('A')` in this case, that has a successful SLD refutation from $F'_b \cup R_b$.

The idea behind this principle is that a decision process always produces some result (an observable atom), and different decision processes must differ at some `msw` thereby entailing mutually exclusive observable atoms. So the uniqueness condition and the exclusiveness condition will be automatically satisfied.

Satisfying the finite support condition is more difficult as it is virtually equivalent to writing a program $DB$ for which all solution search for $\leftarrow G$ ($G \in obs(DB)$) always terminates. Apparently we have no general solution to this problem, but as far as specific models such as HMMs, PCFGs and Bayesian networks are concerned, it can be met. All programs for these models satisfy the finite support condition (and other conditions as well).

### 4.4 Four Conditions Revisited

In this subsection, we discuss how to relax the four simplifying conditions introduced in Subsection 4.2 for the purpose of flexible modeling. We first examine the uniqueness condition considering its crucial role in the adaptation of the EM algorithm to our semantics.

The uniqueness condition guarantees that there exists a (many-to-one) mapping from explanations to observations so that the EM algorithm is applicable (Dempster et al., 1977). It is possible, however, to relax the uniqueness condition while justifying the application of the EM algorithm. We assume the *MAR (missing at random) condition* introduced by Rubin (1976) which is a statistical condition on how a complete data (explanation) becomes an incomplete data (observation), and is customarily assumed implicitly or explicitly in statistics (see Appendix B). By assuming the MAR condition, we can apply our EM

22. Decisions made in the process are a finite subset of $F_{\mathtt{msw}}$.

algorithm to non-exclusive observations $O$ such that $\sum_O P(O) \geq 1$ where the uniqueness condition is seemingly destroyed.

Let us see the MAR condition in action with a simple example. Imagine we walk along a road in front of a lawn. We occasionally observe their state such as "the road is dry but the lawn is wet". Assume that the lawn is watered by a sprinkler running probabilistically. The program $DB_{\tt rl} = R_{\tt rl} \cup F_{\tt rl}$ in Figure 2 describes a sequential process which outputs an observation `observed(road(`$x$`),lawn(`$y$`))` ("the road is $x$ and the lawn is $y$") where $x, y \in \{\tt wet, dry\}$.

```
R_rl  =  { observed(road(X),lawn(Y)):-
              msw(rain,once,A),
              ( A = yes, X = wet, Y = wet
              ; A = no,  msw(sprinkler,once,B),
                   ( B = on, X = dry, Y = wet
                   ; B = off, X = dry, Y = dry ) ).  }
F_rl  =  { msw(rain,once,yes), msw(rain,once,no),
              msw(sprinkler,once,on), msw(sprinkler,once,off) }
```

Figure 2: $DB_{\tt rl}$

The basic distribution over $F_{\tt rl}$ is specified like $P_{F_b}(\cdot)$ in Subsection 4.1, so we omit it. `msw(rain,once,A)` in the program determines whether it rains ($\tt A = yes$) or not ($\tt A = no$), whereas `msw(sprinkler,once,B)` determines whether the sprinkler works fine ($\tt B = on$) or not ($\tt B = off$). Since for each sampled values of $\tt A$ $= a$ ($a \in \{\tt yes, no\}$) and $\tt B$ $= b$ ($b \in \{\tt on, off\}$), there uniquely exists an observation `observed(road(`$x$`),lawn(`$y$`))` ($x, y \in \{\tt wet, dry\}$), there is a many-to-one mapping $\chi : \chi(a, b) = \langle x, y\rangle$. In other words, we can apply the EM algorithm to the observations `observed(road(`$x$`),lawn(`$y$`))` ($x, y \in \{\tt wet, dry\}$). What would happen if we observe exclusively either a state of the road or that of the lawn? Logically, this means we observe $\exists y$ `observed(road(`$x$`),lawn(`$y$`))` or $\exists x$ `observed(road(`$x$`),lawn(`$y$`))`. Apparently the uniqueness condition is not met, because $\exists y$ `observed(road(wet),lawn(`$y$`))` and $\exists x$ `observed(road(`$x$`),lawn(wet))` are compatible (they are true when it rains). Despite the non-exclusiveness of the observations, we can still apply the EM algorithm to them under the MAR condition, which in this case translates into that we observe either the lawn or the road *randomly regardless of their state.*

We now briefly check other conditions. Basically they can be relaxed at the cost of increased computation. Without the exclusiveness condition for instance, we would need an additional process of transforming the support set $\psi_{DB}(G)$ for a goal $G$ into a set of exclusive explanations. For instance, if $G$ has explanations $\{{\tt msw}(a,n,v), {\tt msw}(b,m,w)\}$, we have to transform it into $\{{\tt msw}(a,n,v), \neg{\tt msw}(a,n,v) \wedge {\tt msw}(b,m,w)\}$ and so on.[23] Clearly, this transformation is exponential in the number of `msw` atoms and efficiency concern leads to assuming the exclusiveness condition.

The finite support condition is in practice equivalent to the condition that the SLD tree for $\leftarrow G$ is finite. So relaxing this condition might induce infinite computation.

23. $\neg{\tt msw}(a,n,v)$ is further transformed to a disjunction of exclusive `msw` atoms like $\bigvee_{v' \neq v, v' \in V_a} {\tt msw}(a,n,v')$.

Relaxing the distribution condition and accepting probability distributions other than $P_{\mathtt{msw}}$ serve to expand the horizon of the applicability of parameterized logic programs. In particular the introduction of parameterized joint distributions $P(v_1,\ldots,v_k)$ like Boltzmann distributions over switches $\mathtt{msw}_1,\ldots,\mathtt{msw}_k$ where $v_1,\ldots,v_k$ are values of the switches, makes them *correlated.* Such distributions facilitate writing parameterized logic programs for complicated decision processes in which decisions are not independent but interdependent. Obviously, on the other hand, they increase learning time, and whether the added flexibility of distributions deserves the increased learning time or not is yet to be seen.

### 4.5 Naive Approach to EM Learning

In this subsection, we derive a concrete EM algorithm for parameterized logic programs $DB = F_{\mathtt{msw}} \cup R$ assuming that they satisfy the uniqueness condition, the exclusiveness condition and the finite support condition.

To start, we introduce $Y_o$, a random variable representing our observations according to (5) based on a fixed enumeration of observable atoms in $obs(DB)$. We also introduce another random variable $X_e$ representing their explanations according to (6) based on some fixed enumeration of explanations in $\Psi_{DB}$. Our understanding is that $X_e$ is non-observable while $Y_o$ is observable, and they have a joint distribution $P_{DB}(X_e = x, Y_o = y \mid \theta)$ where $\theta$ denotes relevant parameters. It is then immediate, following (1) and (2) in Section 2, to derive a concrete EM algorithm from the Q function defined by $Q(\theta \mid \theta') \stackrel{\mathrm{def}}{=} \sum_x P_{DB}(x \mid y, \theta') \ln P_{DB}(x, y \mid \theta)$ whose input is a random sample of observable atoms and whose output is the MLE of $\theta$.

In the following, for the sake of readability, we substitute an observable atom $G$ ($G \in obs(DB)$) for $Y_o = y$ and write $P_{DB}(G \mid \theta)$ instead of $P_{DB}(Y_o = y \mid \theta)$. Likewise we substitute an explanation $S$ ($S \in \Psi_{DB}$) for $X_e = x$ and write $P_{DB}(S, G \mid \theta)$ instead of $P_{DB}(X_e = x, Y_o = y \mid \theta)$. Then it follows from the uniqueness condition that

$$P_{DB}(S, G \mid \theta) \quad = \quad \begin{cases} 0 & \text{if} \quad S \notin \psi_{DB}(G) \\ P_{\mathtt{msw}}(S \mid \theta) & \text{if} \quad S \in \psi_{DB}(G). \end{cases}$$

We need yet another notation here. For an explanation $S$, define the *count of* $\mathtt{msw}(i,n,v)$ *in* $S$ by

$$\sigma_{i,v}(S) \stackrel{\mathrm{def}}{=} \left|\{\, n \mid \mathtt{msw}(i,n,v) \in S\}\right|.$$

We have done all preparations now. Suppose we make some observations $\mathcal{G} = G_1,\ldots,G_T$ where $G_t \in obs(DB)$ ($1 \leq t \leq T$). Put

$$\begin{aligned} I &\stackrel{\mathrm{def}}{=} \{i \mid \mathtt{msw}(i,n,v) \in S \in \psi_{DB}(G_t), 1 \leq t \leq T\} \\ \boldsymbol{\theta} &\stackrel{\mathrm{def}}{=} \{\theta_{i,v} \mid \mathtt{msw}(i,n,v) \in S \in \psi_{DB}(G_t), 1 \leq t \leq T\}. \end{aligned}$$

$I$ is a set of switch names that appear in some explanation for one of the $G_t$'s and $\boldsymbol{\theta}$ denotes parameters associated with these switches. $\boldsymbol{\theta}$ is finite due to the finite support condition.

Various probabilities and the Q function are computed by using Proposition A.2 in Appendix A together with our assumptions as follows.

$$P_{DB}(G_t \mid \boldsymbol{\theta}) = P_{DB}\left(\bigvee \psi_{DB}(G_t) \,\middle|\, \boldsymbol{\theta}\right) = \sum_{S \in \psi_{DB}(G_t)} P_{\mathtt{msw}}(S \mid \boldsymbol{\theta}) \tag{8}$$

$$P_{\mathtt{msw}}(S \mid \boldsymbol{\theta}) = \prod_{i \in I, v \in V_i} \theta_{i,v}^{\sigma_{i,v}(S)}$$

$$Q(\boldsymbol{\theta} \mid \boldsymbol{\theta}') \stackrel{\text{def}}{=} \sum_{t=1}^{T} \sum_{S \in \Psi_{DB}} P_{DB}(S \mid G_t, \boldsymbol{\theta}') \ln P_{DB}(S, G_t \mid \boldsymbol{\theta})$$

$$= \sum_{i \in I, v \in V_i} \eta(i, v, \boldsymbol{\theta}') \ln \theta_{i,v} \leq \sum_{i \in I, v \in V_i} \left( \eta(i, v, \boldsymbol{\theta}') \ln \frac{\eta(i, v, \boldsymbol{\theta}')}{\sum_{v' \in V_i} \eta(i, v', \boldsymbol{\theta}')} \right) \tag{9}$$

where

$$\eta(i, v, \boldsymbol{\theta}) \stackrel{\text{def}}{=} \sum_{t=1}^{T} \frac{1}{P_{DB}(G_t \mid \boldsymbol{\theta})} \sum_{S \in \psi_{DB}(G_t)} P_{\mathtt{msw}}(S \mid \boldsymbol{\theta}) \sigma_{i,v}(S)$$

Here we used Jensen's inequality to obtain (9). Note that $P_{DB}(G_t \mid \boldsymbol{\theta})^{-1} \sum_{S \in \psi_{DB}(G_t)} P_{\mathtt{msw}}(S \mid \boldsymbol{\theta}) \sigma_{i,v}(S)$ is the *expected count* of $\mathtt{msw}(i,\cdot,v)$ in an SLD refutation of $G_t$. Speaking of the likelihood function $L(\boldsymbol{\theta}) = \prod_{t=1}^{T} P_{DB}(G_t \mid \boldsymbol{\theta})$, it is already shown in Subsection 2.2 (footnote) that $Q(\boldsymbol{\theta} \mid \boldsymbol{\theta}') \geq Q(\boldsymbol{\theta}' \mid \boldsymbol{\theta}')$ implies $L(\boldsymbol{\theta}) \geq L(\boldsymbol{\theta}')$. Hence from (9), we reach the procedure *learn-naive(DB,$\mathcal{G}$)* below that finds the MLE of the parameters. The array variable $\eta[i, v]$ stores $\eta(i, v, \boldsymbol{\theta})$ under the current $\boldsymbol{\theta}$.

1: **procedure** *learn-naive(DB, $\mathcal{G}$)* **begin**
2: Initialize $\boldsymbol{\theta}$ with appropriate values and $\varepsilon$ with a small positive number ;
3: $\lambda^{(0)} := \sum_{t=1}^{T} \ln P_{DB}(G_t \mid \boldsymbol{\theta})$; % Compute the log-likelihood.
4: **repeat**
5: **foreach** $i \in I, v \in V_i$ **do**
6: $\eta[i, v] := \sum_{t=1}^{T} \frac{1}{P_{DB}(G_t \mid \boldsymbol{\theta})} \sum_{S \in \psi_{DB}(G_t)} P_{\mathtt{msw}}(S \mid \boldsymbol{\theta}) \sigma_{i,v}(S)$;
7: **foreach** $i \in I, v \in V_i$ **do**
8: $\theta_{i,v} := \frac{\eta[i, v]}{\sum_{v' \in V_i} \eta[i, v']}$; % Update the parameters.
9: $m := m + 1$;
10: $\lambda^{(m)} := \sum_{t=1}^{T} \ln P_{DB}(G_t \mid \boldsymbol{\theta})$ % Compute the log-likelihood again.
11: **until** $\lambda^{(m)} - \lambda^{(m-1)} < \varepsilon$ % Terminate if converged.
12: **end**

This EM algorithm is simple and correctly calculates the MLE of $\boldsymbol{\theta}$, but the calculation of $P_{DB}(G_t \mid \boldsymbol{\theta})$ and $\eta[i, v]$(Line 3, 6 and 10) may suffer a combinatorial explosion of explanations. That is, $|\psi_{DB}(G_t)|$ often grows exponentially in the complexity of the model. For instance, $|\psi_{DB}(G_t)|$ for an HMM with $N$ states is $O(N^L)$, exponential in the length $L$ of an input/output string. Nonetheless, suppressing the explosion to realize efficient computation in a polynomial order is possible, under suitable conditions, by avoiding multiple computations of the same subgoal as we see next.

### 4.6 Inside Probability and Outside Probability for Logic Programs

In this subsection, we generalize the notion of inside probability and outside probability (Baker, 1979; Lari & Young, 1990) to logic programs. Major computations in *learn-naive(DB,$\mathcal{G}$)* are those of two terms in Line 6, $P_{DB}(G_t \mid \boldsymbol{\theta})$ and $\sum_{S \in \psi_{DB}(G_t)} P_{\mathtt{msw}}(S \mid \boldsymbol{\theta})\sigma_{i,v}(S)$. Computational redundancy lurks in the naive computation of both terms. We show it by an example. Suppose there is a propositional program $DB_p = F_p \cup R_p$ where $F_p = \{\mathtt{a}, \mathtt{b}, \mathtt{c}, \mathtt{d}, \mathtt{m}\}$ and

$$R_p = \begin{cases} \mathtt{f} \leftarrow \mathtt{a} \wedge \mathtt{g} \\ \mathtt{f} \leftarrow \mathtt{b} \wedge \mathtt{g} \\ \mathtt{g} \leftarrow \mathtt{c} \\ \mathtt{g} \leftarrow \mathtt{d} \wedge \mathtt{h} \\ \mathtt{h} \leftarrow \mathtt{m}. \end{cases} \tag{10}$$

Here $\mathtt{f}$ is an observable atom. We assume that $\mathtt{a}$, $\mathtt{b}$, $\mathtt{c}$, $\mathtt{d}$ and $\mathtt{m}$ are independent and also that $\{\mathtt{a}, \mathtt{b}\}$ and $\{\mathtt{c}, \mathtt{d}\}$ are pair-wise exclusive. Then the support set for $\mathtt{f}$ is calculated as

$$\psi_{DB_p}(\mathtt{f}) = \{\mathtt{a} \wedge \mathtt{c},\ \mathtt{a} \wedge \mathtt{d} \wedge \mathtt{m},\ \mathtt{b} \wedge \mathtt{c},\ \mathtt{b} \wedge \mathtt{d} \wedge \mathtt{m}\}.$$

Hence, in light of (8), we may compute $P_{DB_p}(\mathtt{f})$ as

$$P_{DB_p}(\mathtt{f}) = P_{F_p}(\mathtt{a} \wedge \mathtt{c}) + P_{F_p}(\mathtt{a} \wedge \mathtt{d} \wedge \mathtt{m}) + P_{F_p}(\mathtt{b} \wedge \mathtt{c}) + P_{F_p}(\mathtt{b} \wedge \mathtt{d} \wedge \mathtt{m}). \tag{11}$$

This computation requires 6 multiplications (because $P_{F_p}(\mathtt{a} \wedge \mathtt{c}) = P_{F_p}(\mathtt{a})P_{F_p}(\mathtt{c})$ etc.) and 3 additions. On the other hand, it is possible to compute $P_{DB_p}(\mathtt{f})$ much more efficiently by factoring out common computations. Let $A$ be a ground atom. Define the *inside probability* $\beta(A)$ of $A$ as

$$\beta(A) \stackrel{\text{def}}{=} P_{DB}(A \mid \boldsymbol{\theta}).^{24} \tag{12}$$

Then by applying Theorem A.1 in Appendix A to

$$comp(R_p) \vdash \mathtt{f} \leftrightarrow (\mathtt{a} \wedge \mathtt{g}) \vee (\mathtt{b} \wedge \mathtt{g}),\ \mathtt{g} \leftrightarrow \mathtt{c} \vee (\mathtt{d} \wedge \mathtt{h}),\ \mathtt{h} \leftrightarrow \mathtt{m} \tag{13}$$

which unconditionally holds in our semantics, and by using the independent and the exclusiveness assumption made on $F_p$, the following equations about inside probability are derived.

$$\begin{cases} \beta(\mathtt{f}) &=& \beta(\mathtt{a})\beta(\mathtt{g}) + \beta(\mathtt{b})\beta(\mathtt{g}) \\ \beta(\mathtt{g}) &=& \beta(\mathtt{c}) + \beta(\mathtt{d})\beta(\mathtt{h}) \\ \beta(\mathtt{h}) &=& \beta(\mathtt{m}) \end{cases} \tag{14}$$

$P_{DB_p}(\mathtt{f})(= \beta(\mathtt{f}))$ is obtained by solving (14) about $\beta(\mathtt{f})$, for which only 3 multiplications and 2 additions are required.

It is quite straightforward to generalize (14) but before proceeding, look at a program $DB_q = \{\mathtt{m}\} \cup \{\mathtt{g{:}\text{-}m} \wedge \mathtt{m},\ \mathtt{g{:}\text{-}m}\}$ where $\mathtt{g}$ is an observable atom and $\mathtt{m}$ the only $\mathtt{msw}$ atom. We have $\mathtt{g} \leftrightarrow (\mathtt{m} \wedge \mathtt{m}) \vee \mathtt{m}$ in our semantics, but to compute $P(\mathtt{g}) = P(\mathtt{m})P(\mathtt{m}) + P(\mathtt{m})$ is clearly wrong as it ignores the fact that clause bodies for $\mathtt{g}$, i.e. $\mathtt{m} \wedge \mathtt{m}$ and $\mathtt{m}$ are not mutually exclusive, and atoms in the clause body $\mathtt{m} \wedge \mathtt{m}$ are not independent (here $P(\cdot) = P_{DB_q}(\cdot)$). Similarly, if we set $\mathtt{a} = \mathtt{b} = \mathtt{c} = \mathtt{d} = \mathtt{m}$, the equation (14) will be totally incorrect.

24. Note that if $A$ is a fact in $F$, $\beta(A) = P_{\mathtt{msw}}(A \mid \theta)$.

We therefore add, *temporarily in this subsection*, two assumptions on top of the exclusiveness condition and the finite support condition so that equations like (14) become mathematically correct. The first assumption is that "clause" bodies are mutually exclusive i.e. if there are two clauses $B \leftarrow W$ and $B \leftarrow W'$, $P_{DB}(W \wedge W' \mid \boldsymbol{\theta}) = 0$, and the second assumption is that body atoms are independent, i.e. if $A \leftarrow B_1 \wedge \cdots \wedge B_k$ is a rule, $P_{DB}(B_1 \wedge \cdots \wedge B_k \mid \boldsymbol{\theta}) = P_{DB}(B_1 \mid \boldsymbol{\theta}) \cdots P_{DB}\ (B_k \mid \boldsymbol{\theta})$ holds.

Please note that "clause" used in this subsection has a special meaning. It is intended to mean $G \leftarrow \tau$ where $G$ is a goal and $\tau$ is a *tabled explanation* for $G$ obtained by *OLDT search* both of which will be explained in the next subsection.[25] In other words, these additional conditions are not imposed on a source program but on the result of OLDT search. So clauses for auxiliary computations do not need to satisfy them.

Now suppose clauses about $A$ occur in $DB$ like

$$\begin{array}{l} A \leftarrow B_{1,1} \wedge \cdots \wedge B_{1,i_1} \\ \quad\cdots \\ A \leftarrow B_{L,1} \wedge \cdots \wedge B_{L,i_L} \end{array}$$

where $B_{h,j}$ $(1 \leq h \leq L, 1 \leq j \leq i_h)$ is an atom. Theorem A.1 in Appendix A and the above assumptions ensure

$$\beta(A) = \prod_{j=1}^{i_1} \beta(B_{1,j}) + \cdots + \prod_{j=1}^{i_L} \beta(B_{L,j}). \tag{15}$$

(15) suggests that $\beta(G_t)$ can be considered as a function of $\beta(A)$ if these equations about inside probabilities are hierarchically organized in such a way that $\beta(G_t)$ belongs to the top layer and any $\beta(A)$ appearing on the left hand side only refers to $\beta(B)$'s which belong to the lower layers. We refer to this condition as the *acyclic support condition*. Under the acyclic support condition, equations of the form (15) have a unique solution, and the computation of $P_{DB}(G \mid \boldsymbol{\theta})$ via inside probabilities allows us to take advantage of reusing intermediate results stored as $\beta(A)$, thereby contributing to faster computation of $P_{DB}(G_t \mid \boldsymbol{\theta})$.

Next we tackle a more intricate problem, the computation of $\sum_{S \in \psi_{DB}(G_t)} P_{\mathtt{msw}}(S \mid \boldsymbol{\theta}) \sigma_{i,v}(S)$. Since the sum equals $\sum_n \sum_{\mathtt{msw}(i,n,v) \in S \in \psi_{DB}(G_t)} P_{\mathtt{msw}}(S \mid \boldsymbol{\theta})$, we concentrate on the computation of

$$\mu(G_t, \mathtt{m}) \stackrel{\text{def}}{=} \sum_{\mathtt{m} \in S \in \psi_{DB}(G_t)} P_{\mathtt{msw}}(S \mid \boldsymbol{\theta})$$

where $\mathtt{m} = \mathtt{msw}(i,n,v)$. First we note that if an explanation $S$ contains $\mathtt{m}$ like $S = a_1 \wedge \cdots \wedge a_h \wedge \mathtt{m}$, then we have $\beta(S) = \beta(a_1) \cdots \beta(a_h) \beta(\mathtt{m})$. So $\mu(G_t, \mathtt{m})$ is expressed as

$$\mu(G_t, \mathtt{m}) = \alpha(G_t, \mathtt{m}) \beta(\mathtt{m}) \tag{16}$$

where $\alpha(G_t, \mathtt{m}) = \frac{\partial \mu(G_t, \mathtt{m})}{\partial \beta(\mathtt{m})}$ and $\alpha(G_t, \mathtt{m})$ does not depend on $\beta(\mathtt{m})$. Generalizing this observation to arbitrary ground atoms, we introduce the *outside probability* of ground atom $A$ w.r.t. $G_t$ by

$$\alpha(G_t, A) \stackrel{\text{def}}{=} \frac{\partial \beta(G_t)}{\partial \beta(A)}$$

25. The logical relationship (13) corresponds to (20) where `f`, `g` and `h` are table atoms.

assuming the same conditions as inside probability. In view of (16), the problem of computing $\mu(G_t,\mathtt{m})$ is now reduced to that of computing $\alpha(G_t,\mathtt{m})$, which is recursively computable as follows. Suppose $A$ occurs in the ground program $DB$ like

$$\begin{array}{l} B_1 \leftarrow A \wedge W_{1,1}, \cdots, B_1 \leftarrow A \wedge W_{1,i_1} \\ \quad\cdots \\ B_K \leftarrow A \wedge W_{K,1}, \cdots, B_K \leftarrow A \wedge W_{K,i_K}. \end{array}$$

As $\beta(G_t)$ is a function of $\beta(B_1),\ldots,\beta(B_K)$ by our assumption, the chain rule of derivatives leads to

$$\alpha(G_t, A) = \left(\frac{\partial\beta(G_t)}{\partial\beta(B_1)}\right)\left(\frac{\partial\beta(A \wedge W_{1,1})}{\partial\beta(A)}\right) + \cdots + \left(\frac{\partial\beta(G_t)}{\partial\beta(B_K)}\right)\left(\frac{\partial\beta(A \wedge W_{K,i_K})}{\partial\beta(A)}\right)$$

and hence to[26]

$$\alpha(G_t, G_t) = 1 \qquad (17)$$

$$\alpha(G_t, A) = \alpha(G_t, B_1)\sum_{j=1}^{i_1}\beta(W_{1,j}) + \cdots + \alpha(G_t, B_K)\sum_{j=1}^{i_K}\beta(W_{K,j}). \qquad (18)$$

Therefore if all inside probabilities have already been computed, outside probabilities are recursively computed from the top (17) using (18) downward along the program layers. In the case of $DB_p$ with f and m being chosen atoms, we compute

$$\left\{\begin{array}{lcl} \alpha(\mathtt{f},\mathtt{f}) & = & 1 \\ \alpha(\mathtt{f},\mathtt{g}) & = & \beta(\mathtt{a}) + \beta(\mathtt{b}) \\ \alpha(\mathtt{f},\mathtt{h}) & = & \alpha(\mathtt{f},\mathtt{g})\beta(\mathtt{d}) \\ \alpha(\mathtt{f},\mathtt{m}) & = & \alpha(\mathtt{f},\mathtt{h}). \end{array}\right. \qquad (19)$$

From (19), the desired sum $\mu(\mathtt{f},\mathtt{m})$ is calculated as

$$\mu(\mathtt{f},\mathtt{m}) = \alpha(\mathtt{f},\mathtt{m})\beta(\mathtt{m}) = (\beta(\mathtt{a}) + \beta(\mathtt{b}))\beta(\mathtt{d})\beta(\mathtt{m})$$

which requires only two multiplications and one addition compared to four multiplications and one addition in the naive computation.

Gains obtained by computing inside and outside probability may be small for this case, but as the problem size grows, they become enormous, and compensate enough for additional restrictions imposed on the result of OLDT search.

### 4.7 OLDT Search

To compute inside and outside probability recursively like (15) or (17) and (18), we need at programming level a tabulation mechanism for structure-sharing of partial explanations

26. Because of the independence assumption on body atoms, $W_{h,j}$ $(1 \leq h \leq K, 1 \leq j \leq i_h)$ and $A$ are independent. Therefore

$$\frac{\partial\beta(A \wedge W_{h,j})}{\partial\beta(A)} = \frac{\partial\beta(A)\beta(W_{h,j})}{\partial\beta(A)} = \beta(W_{h,j}).$$

between subgoals. We henceforth deal with programs *DB* in which a set *table*(*DB*) of *table predicates* are declared in advance. A ground atom containing a table predicate is called a *table atom*. The purpose of table atoms is to store their support sets and eliminate the need of recomputation, and by doing so, to construct hierarchically organized explanations made up of the table atoms and the `msw` atoms.

Let $DB = F_{\mathtt{msw}} \cup R$ be a parameterized logic program which satisfies the finite support condition and the uniqueness condition. Also let $G_1, G_2, \ldots, G_T$ be a random sample of observable atoms in $obs(DB)$. We make the following additional assumptions.

**Assumptions:**

For each $t$ $(1 \leq t \leq T)$, there exists a finite set $\{\tau_1^t, \ldots, \tau_{K_t}^t\}$ of table atoms associated with conjunctions $\widetilde{S}_{k,j}^t$ $(0 \leq k \leq K_t, 1 \leq j \leq m_k)$ such that

$$\begin{aligned} comp(R) \;\vdash\; & \left(G_t \leftrightarrow \widetilde{S}_{0,1}^t \vee \cdots \vee \widetilde{S}_{0,m_0}^t\right) \\ & \wedge \left(\tau_1^t \leftrightarrow \widetilde{S}_{1,1}^t \vee \cdots \vee \widetilde{S}_{1,m_1}^t\right) \wedge \cdots \wedge \left(\tau_{K_t}^t \leftrightarrow \widetilde{S}_{K_t,1}^t \vee \cdots \vee \widetilde{S}_{K_t,m_{K_t}}^t\right) \end{aligned} \tag{20}$$

where

- each $\widetilde{S}_{k,j}^t$ $(0 \leq k \leq K_t, 1 \leq j \leq m_k)$ is, as a set, a subset of $F_{\mathtt{msw}} \cup \{\tau_{k+1}, \ldots, \tau_{K_t}\}$ **(acyclic support condition)**. As a convention, we put $\tau_0 = G_t$ and call respectively $\tau_{DB}^t \stackrel{\text{def}}{=} \{\tau_0, \tau_1^t, \ldots, \tau_{K_t}^t\}$ the *set of table atoms for* $G_t$ and $\widetilde{S}_{k,j}^t$ $(k \geq 0)$ a *t-explanation for* $\tau_k^t$.[27] The set of all t-explanations for $\tau_k$ is denoted by $\widetilde{\psi}_{DB}(\tau_k^t)$ and we consider $\widetilde{\psi}_{DB}(\cdot)$ as a function of table atoms.

- t-explanations are mutually exclusive, i.e. for each $k$ $(0 \leq k \leq K_t)$, $P_{DB}(\widetilde{S}_{k,j}^t \wedge \widetilde{S}_{k,j'}^t) = 0$ $(1 \leq j \neq j' \leq m_k)$ **($t-$exclusiveness condition)**.

- $\widetilde{S}_{k,j}^t$ $(0 \leq k \leq K_t, 1 \leq j \leq m_k)$ is a conjunction of independent atoms **(independent condition)**.[28]

These assumptions are aimed at efficient probability computation. Namely, the acyclic support condition makes dynamic programming possible, the t-exclusiveness condition reduces $P_{DB}(A \vee B)$ to $P_{DB}(A) + P_{DB}(B)$ and the independent condition reduces $P_{DB}(A \wedge B)$ to $P_{DB}(A)P_{DB}(B)$. There is one more point concerning efficiency however. Note that the computation in dynamic programming proceeds following the *partial order* on $\tau_{DB}^t$[29] imposed by the acyclic support condition and access to the table atoms will be much simplified if they are linearly ordered. We therefore topologically sort $\tau_{DB}^t$ respecting the said partial order and call the linearized $\tau_{DB}^t$ satisfying the three assumptions (the acyclic support condition, the t-exclusiveness condition and the independent condition) a *hierarchical system of t-explanations for* $G_t$. We write it as $\tau_{DB}^t = \langle \tau_0^t, \tau_1^t, \ldots, \tau_{K_t}^t \rangle$ $(\tau_0 = G_t)$ assuming $\widetilde{\psi}_{DB}(\cdot)$ is implicitly given.[30] Once a hierarchical system of t-explanations for $G_t$ is successfully built

27. Prefix "t-" is an abbreviation of "tabled-".
28. The independence mentioned here only concerns positive propositions. For $B_1, B_2 \in head(DB)$, we say $B_1$ and $B_2$ are independent if $P_{DB}(B_1 \wedge B_2 \mid \theta) = P_{DB}(B_1 \mid \theta)P_{DB}(B_2 \mid \theta)$ for any $\theta$.
29. $\tau_i$ *precedes* $\tau_j$ if and only if the top-down execution of $\tau_i$ w.r.t. *DB* invokes $\tau_j$ directly or indirectly.
30. So now it holds that if $\tau_i$ precedes $\tau_j$ then $i < j$.

from the source program, equations on inside probability and outside probability such as (14) and (19) are automatically derived and solved in time proportional to the size of the equations. It plays a central role in our approach to efficient EM learning.

One way to obtain such t-explanations is to use OLDT search (Tamaki & Sato, 1986; Warren, 1992), a complete refutation method for logic programs. In OLDT search, when a goal $G$ is called for the first time, we set up an entry for $G$ in a *solution table* and store its answer substitutions $G\theta$ there. When a call to an instance $G'$ of $G$ occurs later, we stop solving $G'$ and instead try to retrieve an answer substitution $G\theta$ stored in the solution table by unifying $G'$ with $G\theta$. To record the remaining answer substitutions of $G$, we prepare a *lookup table* for $G'$ and hold a pointer to them.

For self-containedness, we look at details of OLDT search using a sample program $DB_h = F_h \cup R_h$ in Figure 4[31] which depicts an HMM[32] in Figure 3. This HMM has two states $\{\mathtt{s0}, \mathtt{s1}\}$. At a state transition, it probabilistically chooses the next destination from $\{\mathtt{s0}, \mathtt{s1}\}$

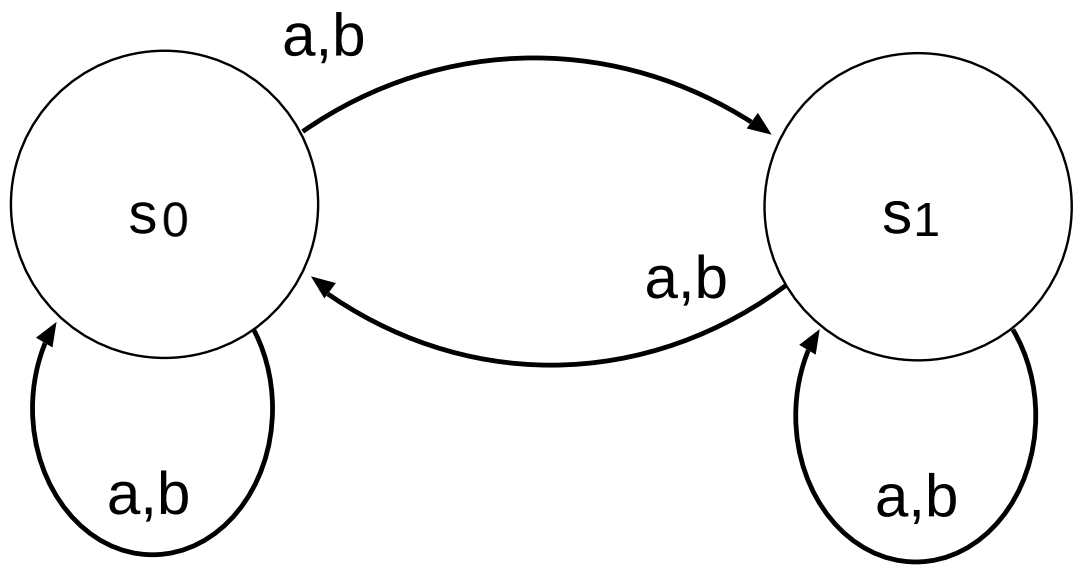

Figure 3: Two state HMM

$F_h$ =
```
f1: values(init, [s0,s1]).
f2: values(out(_),[a,b]).
f3: values(tr(_), [s0,s1]).
```

$R_h$ =
```
h1: hmm(Cs):-                  % To generate a string (chars) Cs,
      msw(init,once,Si),       % Set initial state to Si, and then
      hmm(1,Si,Cs).            % Enter the loop with clock = 1.
h2: hmm(T,S,[C|Cs]):- T=<3,    % Loop:
      msw(out(S),T,C),         % Output C in state S.
      msw(tr(S),T,NextS),      % Transit from S to NextS.
      T1 is T+1,               % Put the clock ahead.
      hmm(T1,NextS,Cs).        % Continue the loop (recursion).
h3: hmm(T,_,[]):- T>3.         % Finish the loop if clock > 3.
```

Figure 4: Two state HMM program $DB_h$

31. f1, f2, f3, h1, h2 and h3 are temporary marks, not part of the program.
32. An HMM defines a probability distribution over strings in the given set of alphabets, and works as a stochastic string generator (Rabiner & Juang, 1993) such that an output string is a sample from the defined distribution.

and also an alphabet from {a,b} to emit. Note that to specify a fact set $F_h$ and the associated distribution compactly, we introduce here a new notation `values(`$i$`,[`$v_1$`,...,`$v_m$`])`. It declares that $F_h$ contains `msw` atoms of the form `msw(`$i$`,`$n$`,`$v$`)` ($v \in \{v_1, \ldots, v_m\}$) whose distribution is $P_{(i,n)}$ given by (7) in Subsection 4.2. For example, (f3), `values(tr(_),[s0,s1])` introduces `msw(tr(`$t$`),`$n$`,`$v$`)` atoms into the program such that $t$ can be any ground term, $v \in \{\mathtt{s0}, \mathtt{s1}\}$ and for a ground term $n$, they have a distribution

$$P_{(\mathtt{tr}(t),n)}(\mathtt{msw(tr(}t\mathtt{)},n\mathtt{,s0)} = x, \mathtt{msw(tr(}t\mathtt{)},n\mathtt{,s1)} = y \mid \theta_{i,s0}, \theta_{i,s1}) = \theta_{i,s0}^{x} \theta_{i,s1}^{y}$$

where $i = \mathtt{tr}(t)$, $x, y \in \{0,1\}$ and $x + y = 1$.

This program runs like a Prolog program. For a non-ground top-goal $\leftarrow$ `hmm(S)`, it functions as a stochastic string generator returning a list of alphabets such as `[a,b,a]` in the variable `S` as follows. The top-goal calls clause (h1) and (h1) selects the initial state by executing subgoal `msw(init,once,Si)`[33] which returns in `Si` an initial state probabilistically chosen from {s0, s1}. The second clause (h2) is called from (h1) with ground `S` and ground `T`. It makes a probabilistic choice of an output alphabet `C` by asking `msw(out(S),T,C)` and then determines `NextS`, the next state, by asking `msw(tr(S),T,NextS)`. (h3) is there to stop the transition. For simplicity, the length of output strings is fixed to three. This way of execution is termed as *sampling execution* because it corresponds to a random sampling from $P_{DB_h}$. If the top-goal is ground like $\leftarrow$ `hmm([a,b,a])`, it works as an acceptor, i.e. returning success (yes) or failure (no).

If all explanations for `hmm([a,b,a])` are sought for, we keep all `msw` atoms resolved upon during the refutation as a conjunction (explanation), and repeat this process by backtracking until no more refutation is found. If we need t-explanations however, backtracking must be abandoned because sharing of partial explanations through t-explanations, the purpose of t-explanations itself, becomes impossible. We therefore instead use OLDT search for all

```
t1:  top_hmm(Cs,Ans):- tab_hmm(Cs,Ans,[]).
t2:  tab_hmm(Cs,[hmm(Cs)|X],X):- hmm(Cs,_,[]).
t3:  tab_hmm(T,S,Cs,[hmm(T,S,Cs)|X],X):- hmm(T,S,Cs,_,[]).
t4:  e_msw(init,T,s0,[msw(init,T,s0)|X],X).
t4': e_msw(init,T,s1,[msw(init,T,s1)|X],X).
  :
t7:  hmm(Cs,X0,X1):- e_msw(init,once,Si,X0,X2), tab_hmm(1,Si,Cs,X2,X1).
t8:  hmm(T,S,[C|Cs],X0,X1):-
         T=<3, e_msw(out(S),T,C,X0,X2), e_msw(tr(S),T,NextS,X2,X3),
         T1 is T+1, tab_hmm(T1,NextS,Cs,X3,X1).
t9:  hmm(T,S,[],X,X):- T>3.
```

Figure 5: Translated program of $DB_h$

33. If `msw(`$i$`,`$n$`,V)` is called with ground $i$ and ground $n$, `V`, a logical variable, behaves like a random variable. It is instantiated to some term $v$ with probability $\theta_{i,v}$ selected from the value set $V_i$ declared by a `values` atom. If, on the other hand, `V` is a ground term $v$ when called, the procedural semantics of `msw(`$i$`,`$n$`,`$v$`)` is equal to that of `msw(`$i$`,`$n$`,V)` $\wedge$ `V` $= v$.

t-explanation search. In the case of the HMM program for example, to build a hierarchical system of t-explanations for `hmm([a,b,a])` by OLDT search, we first declare `hmm/1` and `hmm/3` as table predicate.[34] So a t-explanation will be a conjunction of `hmm/1` atoms, `hmm/3` atoms and `msw` atoms. We then translate the program into another logic program, analogously to the translation of *definite clause grammars* (DCGs) in Prolog (Sterling & Shapiro, 1986). We add two arguments (which forms a *D-list*) to each predicate for the purpose of accumulating `msw` atoms and table atoms as conjuncts in a t-explanation. The translation applied to $DB_h$ yields the program in Figure 5.

In the translated program, clause (`t1`) corresponds to the top-goal $\leftarrow$ `hmm(`$l$`)` with an input string $l$, and a t-explanation for the table atom `hmm(`$l$`)` will be returned in `Ans`. (`t2`) and (`t3`) are auxiliary clauses to add to the callee's D-list a table atom of the form `hmm(`$l$`)` and `hmm(`$t$`,`$s$`,`$l$`)` respectively ($t$: time step, $s$: state). In general, if `p`$/n$ is a table predicate in the original program, `p`$/(n+2)$ becomes a table predicate in the translated program and an auxiliary predicate `tab_p`$/(n+2)$ is inserted to signal the OLDT interpreter to check the solution table for `p`$/n$, i.e. to check if there already exist t-explanations for `p`$/n$. Likewise clauses (`t4`) and (`t4'`) are a pair corresponding to (`f1`) which insert `msw(init,T,`$\cdot$`)` to the callee's D-list with `T` = `once`. Clauses (`t7`), (`t8`) and (`t9`) respectively correspond to (`h1`), (`h2`) and (`h3`).

```
hmm([a,b,a]):[hmm([a,b,a])]
                  '---------> [ [msw(init,once,s0), hmm(1,s0,[a,b,a])],
                                [msw(init,once,s1), hmm(1,s1,[a,b,a])] ]

hmm(1,s0,[a,b,a]):[hmm(1,s0,[a,b,a])]
                      '---------> [ [msw(out(s0),1,a), msw(tr(s0),1,q0), hmm(2,s0,[b,a])],
                                    [msw(out(s0),1,a), msw(tr(s0),1,s1), hmm(2,s1,[b,a])] ]
hmm(1,s1,[a,b,a]):[hmm(1,s1,[a,b,a])]
                      '---------> [ [msw(out(s1),1,a), msw(tr(s1),1,s0), hmm(2,s0,[b,a])],
                                    [msw(out(s1),1,a), msw(tr(s1),1,s1), hmm(2,s1,[b,a])] ]
hmm(2,s0,[b,a]):[hmm(2,s0,[b,a])]
                      '--------> [ [msw(out(s0),2,b), msw(tr(s0),2,s0), hmm(3,s0,[a])],
                                   [msw(out(s0),2,b), msw(tr(s0),2,s1), hmm(3,s1,[a])] ]
hmm(2,s1,[b,a]):[hmm(2,s1,[b,a])]
                      '--------> [ [msw(out(s1),2,b), msw(tr(s1),2,s0), hmm(3,s0,[a])],
                                   [msw(out(s1),2,b), msw(tr(s1),2,s1), hmm(3,s1,[a])] ]
hmm(3,s0,[a]):[hmm(3,s0,[a])]
                     '--------> [ [msw(out(s0),3,a), msw(tr(s0),3,s0), hmm(4,s0,[])],
                                  [msw(out(s0),3,a), msw(tr(s0),3,s1), hmm(4,s1,[])] ]
hmm(3,s1,[a]):[hmm(3,s1,[a])]
                     '--------> [ [msw(out(s1),3,a), msw(tr(s1),3,s0), hmm(4,s0,[])],
                                  [msw(out(s1),3,a), msw(tr(s1),3,s1), hmm(4,s1,[])] ]
hmm(4,s0,[]):[hmm(4,s0,[])]
                     '--------> [[]]

hmm(4,s1,[]):[hmm(4,s1,[])]
                     '--------> [[]]
```

Figure 6: Solution table

34. In general, `p`$/n$ means a predicate `p` with arity $n$. So although `hmm/1` and `hmm/3` share the predicate name `hmm`, they are different predicates.

Then after translation, we apply OLDT search to $\leftarrow$top_hmm([a,b,a],Ans) while noting (i) the added D-list does not influence the OLDT procedure, and (ii) we associate with each solution of a table atom in the solution table a list of t-explanations. The resulting solution table is shown in Figure 6. The first row reads that a call to hmm([a,b,a]) occurred and entered the solution table and its solution, hmm([a,b,a]) (no variable binding generated), has two t-explanations, msw(init,once,s0) $\wedge$ hmm(1,s0,[a,b,a]) and msw(init,once,s1) $\wedge$ hmm(1,s1,[a,b,a]). The remaining task is the topological sorting of the table atoms stored in the solution table respecting the acyclic support condition. This can be done by using depth-first search (trace) of t-explanations from the top-goal for example. Thus we obtain a hierarchical system of t-explanations for hmm([a,b,a]).

### 4.8 Support Graphs

Looking back, all we need to compute inside and outside probability is a hierarchical system of t-explanations, which essentially is a boolean combination of primitive events (msw atoms) and compound events (table atoms) and as such can be more intuitively representable as a graph. For this reason, and to help visualizing our learning algorithm, we introduce a new data-structure termed *support graphs*, though the new EM algorithm in the next subsection itself is described solely by the hierarchical system of t-explanations.

As illustrated in Figure 7 (a), the support graph for $G_t$ is a graphical representation of the hierarchical system of t-explanations $\tau^t_{DB} = \langle \tau^t_0, \tau^t_1, \ldots, \tau^t_{K_t} \rangle$ ($\tau^t_0 = G_t$) for $G_t$ in (20). It consists of totally ordered disconnected subgraphs, each of which is labeled with the corresponding table atom $\tau^t_k$ in $\tau^t_{DB}$ ($0 \leq k \leq K_t$). A subgraph labeled $\tau^t_k$ comprises two special nodes (the *start node* and the *end node*) and *explanation graph*s, each corresponding to a t-explanation $\widetilde{S}^t_{k,j}$ in $\widetilde{\psi}_{DB}(\tau^t_k)$ ($1 \leq j \leq m_k$).

An *explanation graph* of $\widetilde{S}^t_{k,j}$ is a linear graph in which a node is labeled either with a table atom $\tau$ or with a switch msw($\cdot,\cdot,\cdot$) in $\widetilde{S}^t_{k,j}$. They are called a *table node* and a *switch node* respectively. Figure 7 (b) is the support graph for hmm([a,b,a]) obtained from the solution table in Figure 6. Each table node labeled $\tau$ refers to the subgraph labeled $\tau$, so data-sharing is achieved through the distinct table nodes referring to the same subgraph.

### 4.9 Graphical EM Algorithm

We describe here an efficient EM learning algorithm termed the *graphical EM algorithm* (Figure 8) introduced by Kameya and Sato (2000), that runs on support graphs. Suppose we have a random sample $\mathcal{G} = G_1, \ldots, G_T$ of observable atoms. Also suppose support graphs for $G_t$ ($1 \leq t \leq T$), i.e. hierarchical systems of t-explanations satisfying the acyclic support condition, the t-exclusiveness condition and the independent condition, have been successfully constructed from a parameterized logic program *DB* satisfying the uniqueness condition and the finite support condition.

The graphical EM algorithm refines *learn-naive(DB,$\mathcal{G}$)* by introducing two subroutines, *get-inside-probs(DB, $\mathcal{G}$)* to compute inside probabilities and *get-expectations(DB, $\mathcal{G}$)* to compute outside probabilities. They are called from the main routine *learn-gEM(DB,$\mathcal{G}$)*. When learning, we prepare four arrays for each support graph for $G_t$ in $\mathcal{G}$:

- $\mathcal{P}[t,\tau]$ for the *inside probability* of $\tau$, i.e. $\beta(\tau) = P_{DB}(\tau \mid \boldsymbol{\theta})$ (see (12))

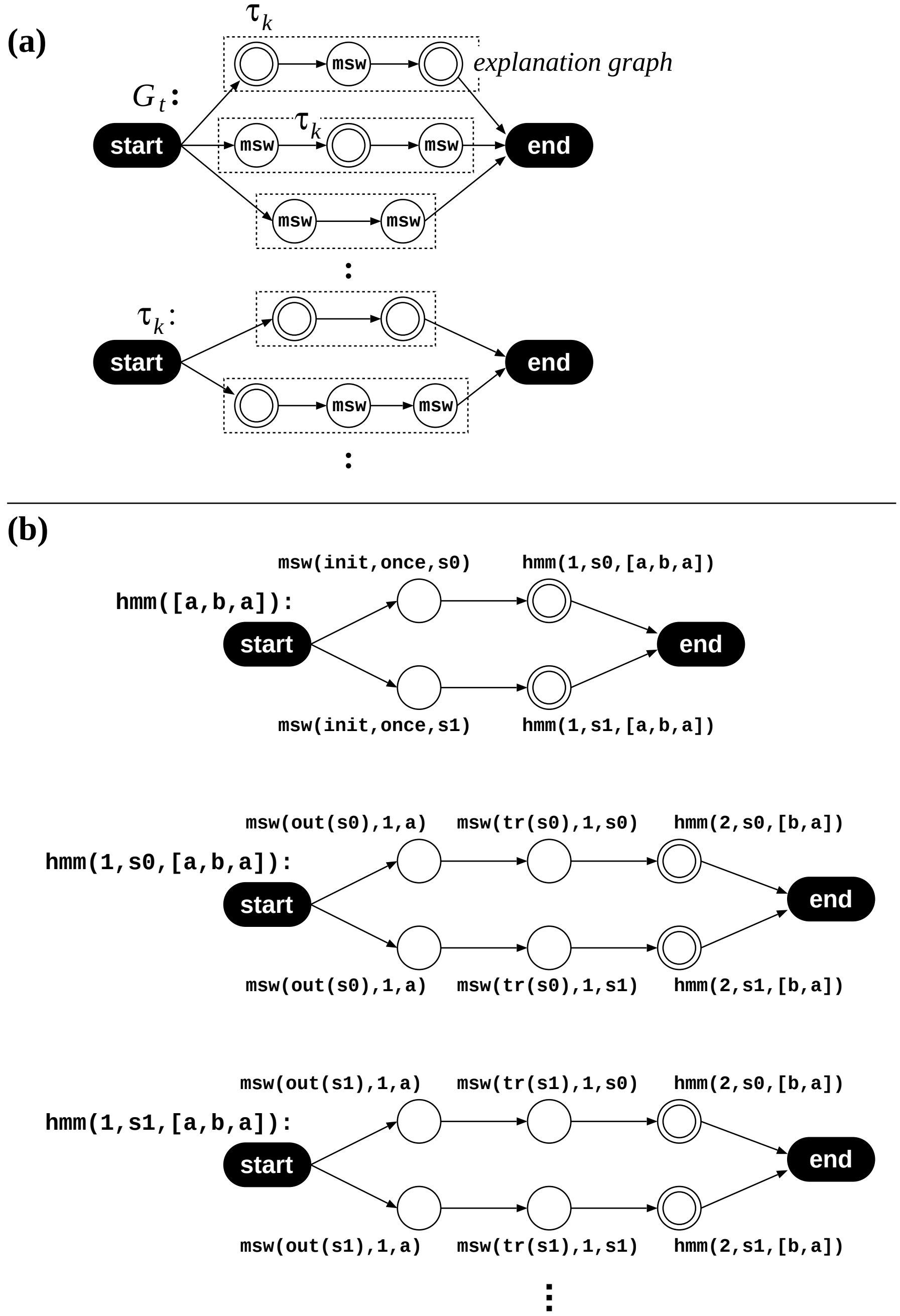

Figure 7: A support graph (a) in general form, (b) for $G_t$ = `hmm([a,b,a])` in the HMM program $DB_h$. A double-circled node refers to a table node.

- $\mathcal{Q}[t,\tau]$ for the *outside probability* of $\tau$ w.r.t. $G_t$, i.e. $\alpha(G_t,\tau)$ (see (17) and (18))
- $\mathcal{R}[t,\tau,\widetilde{S}]$ for the *explanation probability* of $\widetilde{S}\,(\in \widetilde{\psi}_{DB}(\tau_k^t))$, i.e. $P_{DB}(\widetilde{S} \mid \boldsymbol{\theta})$

1: **procedure** *learn-gEM*$(DB, \mathcal{G})$
2: **begin**
3: Select some $\boldsymbol{\theta}$ as initial parameters;
4: *get-inside-probs*$(DB, \mathcal{G})$;
5: $\lambda^{(0)} := \sum_{t=1}^{T} \ln \mathcal{P}[t, G_t]$;
6: **repeat**
7: *get-expectations*$(DB, \mathcal{G})$;
8: **foreach** $i \in I, v \in V_i$ **do**
9: $\eta[i, v] := \sum_{t=1}^{T} \eta[t, i, v] / \mathcal{P}[t, G_t]$;
10: **foreach** $i \in I, v \in V_i$ **do**
11: $\theta_{i,v} := \eta[i, v] / \sum_{v' \in V_i} \eta[i, v']$;
12: *get-inside-probs*$(DB, \mathcal{G})$;
13: $m := m + 1$;
14: $\lambda^{(m)} := \sum_{t=1}^{T} \ln \mathcal{P}[t, G_t]$
15: **until** $\lambda^{(m)} - \lambda^{(m-1)} < \varepsilon$
16: **end.**

1: **procedure** *get-inside-probs*$(DB, \mathcal{G})$
2: **begin**
3: **for** $t := 1$ **to** $T$ **do begin**
4: Let $\tau_0^t = G_t$;
5: **for** $k := K_t$ **downto** $0$ **do begin**
6: $\mathcal{P}[t, \tau_k^t] := 0$;
7: **foreach** $\widetilde{S} \in \widetilde{\psi}_{DB}(\tau_k^t)$ **do begin**
8: Let $\widetilde{S} = \{A_1, A_2, \ldots, A_{|\widetilde{S}|}\}$;
9: $\mathcal{R}[t, \tau_k^t, \widetilde{S}] := 1$;
10: **for** $l := 1$ **to** $|\widetilde{S}|$ **do**
11: **if** $A_l = \mathtt{msw}(i, \cdot, v)$ **then**
12: $\mathcal{R}[t, \tau_k^t, \widetilde{S}] *= \theta_{i,v}$
13: **else** $\mathcal{R}[t, \tau_k^t, \widetilde{S}] *= \mathcal{P}[t, A_l]$;
14: $\mathcal{P}[t, \tau_k^t] += \mathcal{R}[t, \tau_k^t, \widetilde{S}]$
15: **end** /* **foreach** $\widetilde{S}$ */
16: **end** /* **for** $k$ */
17: **end** /* **for** $t$ */
18: **end.**

1: **procedure** *get-expectations*$(DB, \mathcal{G})$ **begin**
2: **for** $t := 1$ **to** $T$ **do begin**
3: **foreach** $i \in I, v \in V_i$ **do** $\eta[t, i, v] := 0$;
4: Let $\tau_0^t = G_t$; $\mathcal{Q}[t, \tau_0^t] := 1.0$;
5: **for** $k := 1$ **to** $K_t$ **do** $\mathcal{Q}[t, \tau_k^t] := 0$;
6: **for** $k := 0$ **to** $K_t$ **do**
7: **foreach** $\widetilde{S} \in \widetilde{\psi}_{DB}(\tau_k^t)$ **do begin**
8: Let $\widetilde{S} = \{A_1, A_2, \ldots, A_{|\widetilde{S}|}\}$;
9: **for** $l := 1$ **to** $|\widetilde{S}|$ **do**
10: **if** $A_l = \mathtt{msw}(i, \cdot, v)$ **then** $\eta[t, i, v] += \mathcal{Q}[t, \tau_k^t] \cdot \mathcal{R}[t, \tau_k^t, \widetilde{S}]$
11: **else** $\mathcal{Q}[t, A_l] += \mathcal{Q}[t, \tau_k^t] \cdot \mathcal{R}[t, \tau_k^t, \widetilde{S}] / \mathcal{P}[t, A_l]$
12: **end** /* **foreach** $\widetilde{S}$ */
13: **end** /* **for** $t$ */
14: **end.**

Figure 8: graphical EM algorithm.

- $\eta[t, i, v]$ for the *expected count* of $\mathtt{msw}(i, \cdot, v)$, i.e. $\sum_{S \in \psi_{DB}(G_t)} P_{\mathtt{msw}}(S \mid \boldsymbol{\theta}) \sigma_{i,v}(S)$

and call the procedure *learn-gEM(DB,$\mathcal{G}$)* in Figure 8. The main routine *learn-gEM(DB,$\mathcal{G}$)* initially computes all inside probabilities (Line 4) and enters a loop in which *get-expectations(DB,$\mathcal{G}$)* is called first to compute the expected count $\eta[t, i, v]$ of $\mathtt{msw}(i, \cdot, v)$ and parameters are updated (Line 11). Inside probabilities are renewed by using the updated parameters before entering the next loop (Line 12).

The subroutine *get-inside-probs(DB,$\mathcal{G}$)* computes the inside probability $\beta(\tau) = P_{DB}(\tau \mid \boldsymbol{\theta})$ (and stores it in $\mathcal{P}[t,\tau]$) of a table atom $\tau$ from the bottom layer up to the topmost layer $\tau_0 = G_t$ (Line 4) of the hierarchical system of t-explanations for $G_t$ (see (20) in Subsection 4.6). It takes a t-explanation $\widetilde{S}$ in $\widetilde{\psi}_{DB}(\tau_k^t)$ one by one (Line 7), decomposes $\widetilde{S}$ into conjuncts and multiplies their inside probabilities which are either known (Line 12) or already computed (Line 13).

The other subroutine *get-expectations(DB,$\mathcal{G}$)* computes outside probabilities following the recursive definitions (17) and (18) in Subsection 4.6 and stores the outside probability $\alpha(G_t,\tau)$ of a table atom $\tau$ in $\mathcal{Q}[t,\tau]$. It first sets the outside probability of the top-goal $\tau_0 = G_t$ to 1.0 (Line 4) and computes the rest of outside probabilities (Line 6) going down the layers of the t-explanation for $G_t$ described by (20) in Subsection 4.6. (Line 10) adds $\mathcal{Q}[t,\tau_k^t]\cdot\mathcal{R}[t,\tau_k^t,\widetilde{S}] = \alpha(G_t,\tau_k^t)\cdot\beta(\widetilde{S})$ to $\eta[t,i,v]$, the expected count of $\mathtt{msw}(i,\cdot,v)$, as a contribution of $\mathtt{msw}(i,\cdot,v)$ in $\widetilde{S}$ through $\tau_k^t$ to $\eta[t,i,v]$. (Line 11) increments the outside probability $\mathcal{Q}[t,A_l] = \alpha(G_t,A_l)$ of $A_l$ according to the equation (18). Notice that $\mathcal{Q}[t,\tau_k^t]$ has already been computed and $\mathcal{R}[t,\tau_k^t,\widetilde{S}]/\mathcal{P}[t,A_l] = \beta(W)$ for $\widetilde{S} = A_l \wedge W$. As shown in Subsection 4.5, *learn-naive(DB,$\mathcal{G}$)* is the MLE procedure, hence the following theorem holds.

**Theorem 4.1** *Let DB be a parameterized logic program, and $\mathcal{G} = G_1,\ldots,G_T$ a random sample of observable atoms. Suppose the five conditions (uniqueness, finite support (Subsection 4.2), acyclic support, t-exclusiveness and independence (Subsection 4.7)) are met. Then* learn-gEM $(DB,\mathcal{G})$ *finds the MLE $\boldsymbol{\theta}^*$ which (locally) maximizes the likelihood $L(\mathcal{G}\mid\boldsymbol{\theta}) = \prod_{t=1}^{T} P_{DB}(G_t\mid\boldsymbol{\theta})$.*

(Proof) Sketch.[35] Since the main routine *learn-gEM(DB,$\mathcal{G}$)* is the same as *learn-naive(DB,$\mathcal{G}$)* except the computation of $\eta[i,v] = \sum_{t=1}^{T}\eta[t,i,v]$, we show that $\eta[t,i,v] = \sum_{S\in\psi_{DB}(G_t)} P_{\mathtt{msw}}(S\mid\boldsymbol{\theta})\sigma_{i,v}(S)$ $(= \sum_n \sum_{\mathtt{msw}(i,n,v)\in S\in\psi_{DB}(G_t)} P_{\mathtt{msw}}(S\mid\boldsymbol{\theta}))$. However,

$$\begin{aligned}
\eta[t,i,v] &= \sum_{0\le k\le K_t}\sum_{n}\sum_{\mathtt{msw}(i,n,v)\in\widetilde{S}\in\widetilde{\psi}_{DB}(\tau_k^t)} \alpha(G_t,\tau_k^t)\beta(\widetilde{S}) \\
&\quad \text{(see (Line 10) in } \textit{get-expectations}(DB,\mathcal{G})) \\
&= \sum_n \alpha(G_t,\mathtt{msw}(i,n,v))\beta(\mathtt{msw}(i,n,v)) \\
&= \sum_n \mu(G_t,\mathtt{msw}(i,n,v)) \quad \text{(see the equation (16))} \\
&= \sum_n \sum_{\mathtt{msw}(i,n,v)\in S\in\psi_{DB}(G_t)} P_{\mathtt{msw}}(S\mid\boldsymbol{\theta}). \qquad \text{Q.E.D.}
\end{aligned}$$

Here we used the fact that if $\widetilde{S}$ contains $\mathtt{msw}(i,n,v)$ like $\widetilde{S} = \widetilde{S}'\wedge\mathtt{msw}(i,n,v)$, $\beta(\widetilde{S}) = \beta(\widetilde{S}')\beta(\mathtt{msw}(i,n,v))$ holds, and hence

$$\begin{aligned}
&\alpha(G_t,\tau_k^t)\beta(\widetilde{S}) = \alpha(G_t,\tau_k^t)\beta(\widetilde{S}')\beta(\mathtt{msw}(i,n,v)) \\
&\quad = \text{(contribution of } \mathtt{msw}(i,n,v) \text{ in } \widetilde{S} \text{ through } \tau_k^t \text{ to } \alpha(G_t,\mathtt{msw}(i,n,v)))\beta(\mathtt{msw}(i,n,v)).
\end{aligned}$$

35. A formal proof is given by Kameya (2000). It is proved there that under the common parameters $\theta$, $\eta[i,v]$ in *learn-naive(DB,$\mathcal{G}$)* coincides with $\eta[i,v]$ in *learn-gEM(DB,$\mathcal{G}$)*. So, the parameters are updated to the same values. Hence, starting with the same initial values, the parameters converge to the same values.

The five conditions on the applicability of the graphical EM algorithm may look hard to satisfy at once. Fortunately, the modeling principle in Section 4.3 still stands, and with due care in modeling, it is likely to lead us to a program that meets all of them. Actually, we will see in the next section, programs for standard symbolic-statistical frameworks such as Bayesian networks, HMMs and PCFGs all satisfy the five conditions.

## 5. Complexity

In this section, we analyze the time complexity of the graphical EM algorithm applied to various symbolic-statistical frameworks including HMMs, PCFGs, pseudo PCSGs and Bayesian networks. The results show that the graphical EM algorithm is competitive with these specialized EM algorithms developed independently in each research field.

### 5.1 Basic Property

Since the EM algorithm is an iterative algorithm and since we are unable to predict when it converges, we measure time complexity by the time taken for one iteration. We therefore estimate time per iteration on the **repeat** loop of *learn-gEM*$(DB, \mathcal{G})$ ($\mathcal{G} = G_1, \ldots, G_T$). We observe that in one iteration, each support graph for $G_t$ ($1 \leq t \leq T$) is scanned twice, once by *get-inside-probs*$(DB, \mathcal{G})$ and once by *get-expectations*$(DB, \mathcal{G})$. In the scan, addition is performed on the t-explanations, and multiplication (possibly with division) is performed on the `msw` atoms and table atoms once for each. So time spent for $G_t$ per iteration by the graphical EM algorithm is linear in the *size of the support graph*, i.e. the number of nodes in the support graph for $G_t$. Put

$$\begin{aligned}
\widetilde{\Delta}^t_{DB} &\stackrel{\text{def}}{=} \bigcup_{\tau \in \tau^t_{DB}} \widetilde{\psi}_{DB}(\tau) \\
\xi_{\text{num}} &\stackrel{\text{def}}{=} \max_{1 \leq t \leq T} |\widetilde{\Delta}^t_{DB}| \\
\xi_{\text{maxsize}} &\stackrel{\text{def}}{=} \max_{1 \leq t \leq T, \widetilde{S} \in \widetilde{\Delta}^t_{DB}} |\widetilde{S}|.
\end{aligned}$$

Recall that $\tau^t_{DB}$ is the set of table atoms for $G_t$, and hence $\widetilde{\Delta}^t_{DB}$ is the set of all t-explanations appearing in the right hand side of (20) in Subsection 4.7. So $\xi_{\text{num}}$ is the maximum number of t-explanations in a support graph for the $G_t$'s and $\xi_{\text{maxsize}}$ the maximum size of a t-explanation for the $G_t$'s respectively. The following is obvious.

**Proposition 5.1** *The time complexity of the graphical EM algorithm per iteration is linear in the total size of support graphs,* $O(\xi_{\text{num}}\xi_{\text{maxsize}}T)$ *in notation, which coincides with the space complexity because the graphical EM algorithm runs on support graphs.*

This is a rather general result, but when we compare the graphical EM algorithm with other EM algorithms, we must remember that the input to the graphical EM algorithm is support graphs (one for each observed atom) and our actual total learning time is

$$\text{OLDT time} + (\text{the number of iterations}) \times O(\xi_{\text{num}}\xi_{\text{maxsize}}T)$$

where "OLDT time" denotes time to construct all support graphs for $\mathcal{G}$. It is the sum of time for OLDT search and time for the topological sorting of the table atoms, but because the latter is part of the former order-wise,[36] we represent "OLDT time" by time for OLDT search. Also observe that the total size of support graphs does not exceed time for OLDT search for $\mathcal{G}$ order-wise.

To evaluate OLDT time for a specific class of models such as HMMs, we need to know time for table operations. Observe that our OLDT search in this paper is special in the sense that table atoms are always ground when called and there is no resolution with solved goals. Accordingly a solution table is used only

- to check if a goal $G$ already has an entry in the solution table, i.e. if it was called before, and
- to add a new searched t-explanation for $G$ to the list of discovered t-explanations under $G$'s entry.

The time complexity of these operations is equal to that of table access which depends both on the program and on the implementation of the solution table.[37] We first suppose programs are carefully written in such a way that the arguments of table atoms used as indecies for table access are integers. Actually all programs used in the subsequent complexity analysis ($DB_h$ in Subsection 4.7, $DB_g$ and $DB_{g'}$ in Subsection 5.3, $DB_{G^{\tau}}$ in Subsection 5.5) satisfy or can satisfy this condition by replacing non-integer terms with appropriate integers. We also suppose that the solution table is implemented using an array so that table access can be done in $O(1)$ time.[38]

In what follows, we present a detailed analysis of the time complexity of the graphical EM algorithm applied to HMMs, PCFGs, pseudo PCSGs and Bayesian networks, assuming $O(1)$ time access to the solution table. We remark by the way that their space complexity is just the total size of solution tables (support graphs).

### 5.2 HMMs

The standard EM algorithm for HMMs is the Baum-Welch algorithm (Rabiner, 1989; Rabiner & Juang, 1993). An example of HMM is shown in Figure 3 in Subsection 4.7.[39] Given $T$ observations $w^1, \ldots, w^T$ of output string of length $L$, it computes in $O(N^2LT)$ time in each iteration the forward probability $\alpha^t_m(q) = P(o^t_1 o^t_2 \cdots o^t_{m-1}, q \mid \boldsymbol{\theta})$ and the backward probability $\beta^t_m(q) = P(o^t_m o^t_{m+1} \cdots o^t_L \mid q, \boldsymbol{\theta})$ for each state $q \in Q$, time step $m$ $(1 \leq m \leq L)$ and a string $w^t = o^t_1 o^t_2 \cdots o^t_L$ $(1 \leq t \leq T)$, where $Q$ is the set of states and $N$ the number of states. The factor $N^2$ comes from the fact that every state has $N$ possible destinations and

36. Think of OLDT search for a top-goal $G_t$. It searches for `msw` atoms and table atoms to create a solution table, while doing some auxiliary computations. Therefore its time complexity is never less than $O(|\text{the number of } \texttt{msw} \text{ atoms and table atoms in the support graph for } G_t|)$, which coincides with the time we need to topologically sort table atoms in the solution table by depth-first search from $\tau_0 = G_t$.
37. Sagonas et al. (1994) and Ramakrishnan et al. (1995) discuss about the implementation of OLDT.
38. If arrays are not available, we may be able to use balanced trees, giving $O(\log n)$ access time where $n$ is the number data in the solution table, or we may be able to use hashing, giving average $O(1)$ time access under a certain condition (Cormen, Leiserson, & Rivest, 1990).
39. We treat here only "state-emission HMMs" which emit a symbol depending on the state. Another type, "arc-emission HMMs" in which the emitted symbol depends on the transition arc, is treated similarly.

we have to compute the forward and backward probability for every destination and every state. After computing all $\alpha^t_m(q)$'s and $\beta^t_m(q)$'s, parameters are updated. So, the total computation time in each iteration of the Baum-Welch algorithm is estimated as $O(N^2LT)$ (Rabiner & Juang, 1993; Manning & Schütze, 1999).

To compare this result with the graphical EM algorithm, we use the HMM program $DB_h$ in Figure 4 with appropriate modifications to $L$, the length of a string, $Q$, the state set, and declarations in $F_h$ for the output alphabets. For a string $w = o_1 o_2 \cdots o_L$, $\mathtt{hmm}(n,q,[o_m,o_{m+1},\ldots,o_L])$ in $DB_h$ reads that the HMM is in state $q \in Q$ at time $n$ and has to output $[o_m, o_{m+1}, \ldots, o_L]$ until it reaches the final state. After declaring hmm/1 and hmm/3 as table predicate and translation (see Figure 5), we apply OLDT search to the goal $\leftarrow \mathtt{top_hmm}([o_1,\ldots,o_L],\mathtt{Ans})$ w.r.t. the translated program to obtain all t-explanations for $\mathtt{hmm}([o_1,\ldots,o_L])$. For a complexity argument however, the translated program and $DB_h$ are the same, so we talk in terms of $DB_h$ for the sake of simplicity. In the search, we fix the search strategy to *multi-stage depth-first strategy* (Tamaki & Sato, 1986). We assume that the solution table is accessible in $O(1)$ time.[40] Since the length of the list in the third argument of hmm/3 decreases by one on each recursion, and there are only finitely many choices of the state transition and the output alphabet, the search terminates, leaving finitely many t-explanations in the solution table like Figure 6 that satisfy the acyclic support condition respectively. Also the sampling execution of $\leftarrow$hmm(L) w.r.t. $DB_h$ is nothing but a sequential decision process such that decisions made by msw atoms are exclusive, independent and generate a unique string, which means $DB_h$ satisfies the t-exclusiveness condition, the independence condition and the uniqueness condition respectively. So, the graphical EM algorithm is applicable to the set of hierarchical systems of t-explanations for $\mathtt{hmm}(w^t)$ $(1 \le t \le T)$ produced by OLDT search for $T$ observations $w^1,\ldots,w^T$ of output string. Put $w^t = o^t_1 o^t_2 \cdots o^t_L$. It follows from

$$\tau^t_{DB_h} = \{\mathtt{hmm}(m,q,[o^t_m,\ldots,o^t_L]) \mid 1 \le m \le L+1, q \in Q\} \cup \{\mathtt{hmm}([o^t_1,\ldots,o^t_L])\}$$

$$\underset{(1 \le m \le L)}{\widetilde{\psi}_{DB_h}(\mathtt{hmm}(m,q,[o^t_m,\ldots,o^t_L]))} = \left\{ \begin{array}{c} \mathtt{msw}(\mathtt{out}(q),m,o_m),\mathtt{msw}(\mathtt{tr}(q),m,q'), \\ \mathtt{hmm}(m+1,q',[o^t_{m+1},\ldots,o^t_L]) \end{array} \middle| \, q' \in Q \right\}$$

that for a top-goal $\mathtt{hmm}([o^t_1,\ldots,o^t_L])$, there are at most $O(NL)$ calling patterns of hmm/3 and each call causes at most $N$ calls to hmm/3, implying there occur $O(NL \cdot N) = O(N^2L)$ calls to hmm/3. Since each call is computed once due to the tabling mechanism, we have $\xi_{\mathrm{num}} = O(N^2L)$. Also $\xi_{\mathrm{maxsize}} = 3$. Applying Proposition 5.1, we reach

**Proposition 5.2** *Suppose we have $T$ strings of length $L$. Also suppose each table operation in OLDT search is done in $O(1)$ time. OLDT time by $DB_h$ is $O(N^2LT)$ and the graphical EM algorithm takes $O(N^2LT)$ time per iteration where $N$ is the number of states.*

$O(N^2LT)$ is the time complexity of the Baum-Welch algorithm. So the graphical EM algorithm runs as efficiently as the Baum-Welch algorithm.[41]

40. $O(1)$ is possible because in the translated program $DB_h$ in Section 4.7, we can identify a goal pattern of $\mathtt{hmm}(\cdot,\cdot,\cdot,\cdot,\cdot)$ by the first two arguments which are constants (integers).
41. Besides, the Baum-Welch algorithm and the graphical EM algorithm whose input are support graphs generated by $DB_h$ update parameters to the same value if initial values are the same.

By the way, the Viterbi algorithm (Rabiner, 1989; Rabiner & Juang, 1993) provides for HMMs an efficient way of finding the most likely transition path for a given input/output string. A similar algorithm for parameterized logic programs that determines the most likely explanation for a given goal can be derived. It runs in time linear in the size of the support graph, thereby $O(N^2L)$ in the case of HMMs, the same complexity as the Viterbi algorithm (Sato & Kameya, 2000).

### 5.3 PCFGs

We now compare the graphical EM algorithm with the Inside-Outside algorithm (Baker, 1979; Lari & Young, 1990). The Inside-Outside algorithm is a well-known EM algorithm for PCFGs (Wetherell, 1980; Manning & Schütze, 1999).[42] It takes a grammar in *Chomsky normal form*. Given $N$ nonterminals, a production rule in the grammar takes the form $i \rightarrow j, k$ ($1 \leq i, j, k \leq N$) (nonterminals are named by numbers from 1 to $N$ and 1 is a starting symbol) or the form $i \rightarrow w$ where $1 \leq i \leq N$ and $w$ is a terminal. In each iteration, it computes the inside probability and the outside probability of every partial parse tree of the given sentence to update parameters for these production rules. Time complexity is measured by time per iteration, and is described by $N$, the number of nonterminals, and $L$, the number of terminals in a sentence. It is $O(N^3L^3T)$ for $T$ observed sentences (Lari & Young, 1990).

To compare the graphical EM algorithm with the Inside-Outside algorithm, we start from a propositional program $DB_g = F_g \cup R_g$ below representing the largest grammar containing all possible rules $i \rightarrow j, k$ in $N$ nonterminals where nonterminal 1 is a starting symbol, i.e. sentence.

$$F_g = \{\texttt{msw}(i,\texttt{[}d,d'\texttt{]},\texttt{[}j,k\texttt{]}) \mid 1 \leq i,j,k \leq N, d, d' \text{ are numbers}\} \cup \{\texttt{msw}(i,d,w) \mid 1 \leq i \leq N, d \text{ is a number}, w \text{ is a terminal}\}$$

$$R_g = \left\{ \begin{array}{lll} \texttt{q}(i,d_0,d_2) & \texttt{:-} & \texttt{msw}(i,\texttt{[}d_0,d_2\texttt{]},\texttt{[}j,k\texttt{]}), \\ & & \texttt{q}(j,d_0,d_1), \\ & & \texttt{q}(k,d_1,d_2). \end{array} \middle| \begin{array}{l} 1 \leq i,j,k \leq N, \\ 0 \leq d_0 < d_1 < d_2 \leq L \end{array} \right\}$$

$$\cup \left\{ \texttt{q}(i,d,d+1) \;\texttt{:-}\; \texttt{msw}(i,d,w_{d+1}). \;\middle|\; 1 \leq i \leq N, 0 \leq d \leq L-1 \right\}$$

Figure 9: PCFG program $DB_g$

$DB_g$ is an artificial parsing program whose sole purpose is to measure the size of an OLDT tree[43] created by the OLDT interpreter when it parses a sentence $w_1 w_2 \cdots w_L$. So

42. A PCFG (probabilistic context free grammar) is a backbone CFG with probabilities (parameters) assigned to each production rule. For a nonterminal $A$ having $n$ production rules $\{A \rightarrow \alpha_i \mid 1 \leq i \leq n\}$, a probability $p_i$ is assigned to $A \rightarrow \alpha_i$ ($1 \leq i \leq n$) where $\sum_{i=1}^{n} p_i = 1$. The probability of a sentence $s$ is the sum of probabilities of each (leftmost) derivation of $s$. The latter is the product of probabilities of rules used in the derivation.

43. To be more precise, an OLDT structure, but in this case, it is a tree because $DB_g$ contains only constants (Datalog program) and there never occurs the need of creating a new root node.

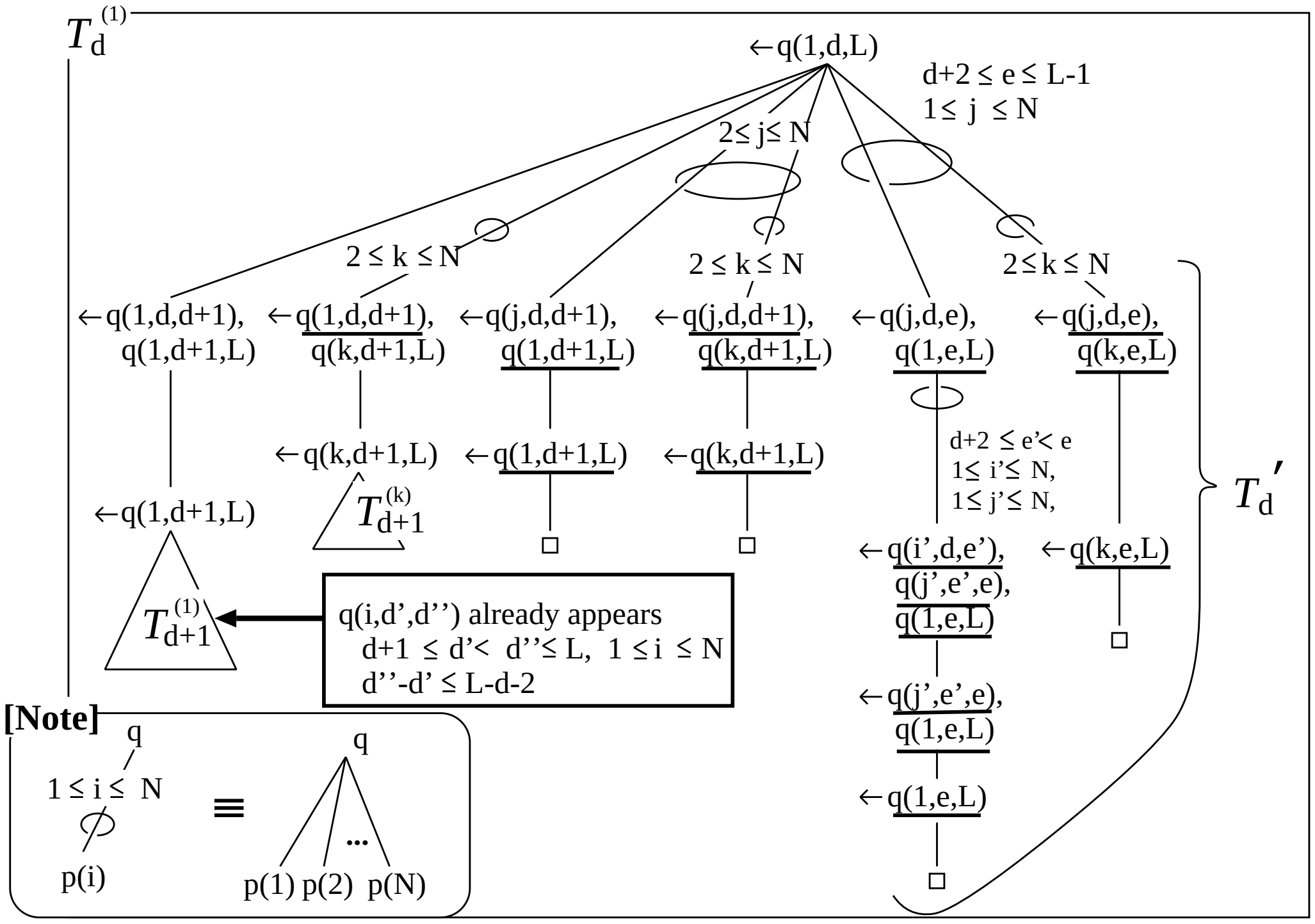

Figure 10: OLDT tree for the query $\leftarrow \mathtt{q(1,}d\mathtt{,}L\mathtt{)}$

the input sentence $w_1 w_2 \cdots w_L$ is embedded in the program separately as $\mathtt{msw(}i\mathtt{,}d\mathtt{,}w_{d+1}\mathtt{)}$ $(0 \leq d \leq L-1)$ in the second clauses of $R_g$ (this treatment does not affect the complexity argument). $\mathtt{q(}i\mathtt{,}d_0\mathtt{,}d_1\mathtt{)}$ reads that the $i$-th nonterminal spans from position $d_0$ to position $d_1$, i.e. the substring $w_{d_0+1} \cdots w_{d_1}$. The first clauses $\mathtt{q(}i\mathtt{,}d_0\mathtt{,}d_2\mathtt{)}$ `:- msw(·,·,·), ` $\mathtt{q(}j\mathtt{,}d_0\mathtt{,}d_1\mathtt{),}$ $\mathtt{q(}k\mathtt{,}d_1\mathtt{,}d_2\mathtt{)}$ are supposed to be textually ordered according to the lexicographic order for tuples $\langle i, j, k, d_0, d_2, d_1 \rangle$. As a parser, the top-goal is set to $\leftarrow \mathtt{q(1,0,}L\mathtt{)}$.[44] It asks the parser to parse the whole sentence $w_1 w_2 \cdots w_L$ as the syntactic category "1" (sentence).

We make an exhaustive search for this query by OLDT search.[45] As before, the multi-stage depth-first search strategy and $O(1)$ time access to the solution table are assumed. Then the time complexity of OLDT search is measured by the number of nodes in the OLDT tree. Let $T_d^{(k)}$ be the OLDT tree for $\leftarrow \mathtt{q(}k\mathtt{,}d\mathtt{,}L\mathtt{)}$. Figure 10 illustrates $T_d^{(1)}$ for $d$ $(0 \leq d \leq L-3)$ where `msw` atoms are omitted. As can be seen, the tree has many similar subtrees, so we put them together (see **Note** in Figure 10). Due to the depth-first strategy, $T_d^{(1)}$ has a recursive structure and contains $T_{d+1}^{(1)}$ as a subtree. Nodes whose leftmost atom is not underlined are solution nodes, i.e. they solve their leftmost atoms for the first time in the entire refutation process. The underlined atoms are already computed in the subtrees to their left.[46] They only check the solution table if there are their entries (= already

44. $L$ here is not a Prolog variable but a constant denoting the sentence length.

45. `q` is a table predicate.

46. It can be inductively proved that $T_{d+1}^{(1)}$ contains every computed $\mathtt{q(}i\mathtt{,}d'\mathtt{,}d''\mathtt{)}$ $(0 \leq d \leq L-3,\ d+1 \leq d' < d'' \leq L,\ 1 \leq i \leq N,\ d''-d' \leq L-d-2)$.

computed) in $O(1)$ time. Since all clauses are ground, such execution only generates a single child node.

We enumerate $h_d^{(1)}$, the number of nodes in $T_d^{(1)}$ but not in $T_{d+1}^{(k)}$ $(1 \leq k \leq N)$. From Figure 10, we see $h_d^{(1)} = O(N^3(L-d)^2)$.[47] Let $h_d^{(k)}$ $(2 \leq k \leq N)$ be the number of nodes in $T_{d+1}^{(k)}$ not contained in $T_{d+1}^{(1)}$. It is estimated as $O(N^2(L-d-2))$. Consequently, the number of nodes that are newly created in $T_d^{(1)}$ is $h_d^{(1)} + \sum_{k=2}^{N} h_d^{(k)} = O(N^3(L-d)^2)$. As a result, total time for OLDT search is computed as $\sum_{d=0}^{L-3} h_d = O(N^3L^3)$[48] which is also the size of the support graph.

We now consider a non-propositional parsing program $DB_{g'} = F_{g'} \cup R_{g'}$ in Figure 11 whose ground instances constitute the propositional program $DB_g$. $DB_{g'}$ is a probabilistic variant of DCG program (Pereira & Warren, 1980) in which `q'/1`, `q'/6` and `between/3` are declared as table predicate. Semantically $DB_{g'}$ specifies a probability distribution over the atoms of the form $\{$`q'(`$l$`)` $\mid l$ is a list of terminals$\}$.

$$F_{g'} = \{\texttt{msw(}s_i\texttt{,}t\texttt{,[}s_j\texttt{,}s_k\texttt{])} \mid 1 \leq i,j,k \leq N, t \text{ is a number}\} \cup \{\texttt{msw(}s_i\texttt{,}t\texttt{,}w\texttt{)} \mid 1 \leq i \leq N, t \text{ is a number}, w \text{ is a terminal}\}$$

$R_{g'}$ =

```
q'(S)  :-  length(S,D), q'(s1,0,D,0,_,S-[]).

q'(I,D0,D2,C0,C2,L0-L2)  :-  between(D0,D1,D2),
                             msw(I,C0,[J,K]),
                             q'(J,D0,D1,s(C0),C1,L0-L1),
                             q'(K,D1,D2,s(C1),C2,L1-L2).

q'(I,D,s(D),C0,s(C0),[W|X]-X)  :-  msw(I,C0,W).
```

Figure 11: Probabilistic DCG like program $DB_{g'}$

The top-goal to parse a sentence `S` = `[`$w_1,\ldots,w_L$`]` is $\leftarrow$`q'([`$w_1,\ldots,w_L$`])`. It invokes `q'(`$s_1$`,0,D,0,_,[`$w_1,\ldots,w_L$`]-[])` after measuring the length `D` of an input sentence `S` by calling `length/2`. [49] [50] In general, `q'(`$i$`,`$d_0$`,`$d_2$`,`$c_0$`,`$c_2$`,`$l_0$`-`$l_2$`)` works identically to `q(`$i$`,`$d_0$`,`$d_2$`)` but three arguments, $c_0$, $c_2$ and $l_0$-$l_2$, are added. $c_0$ supplies a unique trial-id for `msw`s to be used in the body, $c_2$ the latest trial-id in the current computation, and $l_0$-$l_2$ a D-list holding a substring between $d_0$ and $d_2$. Since the added arguments do not affect the shape of the

47. We here focus on the subtree $T'_d$. $j$, $i'$ and $j'$ range from 1 to $N$, and $\left|\{(e,e') \mid d+2 \leq e' < e \leq L-1\}\right| = O((L-d)^2)$. Hence, the number of nodes in $T'_d$ is $O(N^3(L-d)^2)$. The number of nodes in $T_d^{(1)}$ but neither in $T_{d+1}^{(1)}$ nor in $T'_d$ is negligible, therefore $h_d^{(1)} = O(N^3(L-d)^2)$.
48. The number of nodes in $T_{L-1}^{(1)}$ and $T_{L-2}^{(1)}$ is negligible.
49. To make the program as simple as possible, we assume that an integer $n$ is represented by a ground term $s_n \stackrel{\text{def}}{=} \overbrace{\texttt{s(}\cdots\texttt{s}}^{(n)}\texttt{(0)}\cdots\texttt{)}$. We also assume that when `D0` and `D2` are ground, the goal $\leftarrow$`between(D0, D1, D2)` returns an integer `D1` between them in time proportional to $|\texttt{D1} - \texttt{D0}|$.
50. We omit an obvious program for `length(`$l$`,`$s_n$`)` which computes the length $s_n$ of a list $l$ in $O(|l|)$ time.

search tree in Figure 10 and the extra computation caused by `length/2` is $O(L)$ and the one by the insertion of `between(D0,D1,D2)` is $O(NL^3)$ respectively,[51] OLDT time remains $O(N^3L^3)$, and hence so is the size of the support graph.

To apply the graphical EM algorithm correctly, we need to confirm the five conditions on its applicability. It is rather apparent however that the OLDT refutation of any top-goal of the form $\leftarrow$ `q'([`$w_1$`,...,`$w_L$`])` w.r.t. $DB_{g'}$ terminates, and leaves a support graph satisfying the finite support condition and the acyclic support condition. The t-exclusiveness condition and the independent condition also hold because the refutation process faithfully simulates the leftmost stochastic derivation of $w_1 \cdots w_L$ in which the choice of a production rule made by `msw(`$s_i$`,`$s_c$`,[`$s_j$`,`$s_k$`])` is exclusive and independent (trial-ids are different on different choices).

What remains is the uniqueness condition. To confirm it, let us consider another program $DB_{g''}$, a modification of $DB_{g'}$ such that the first goal `length(S,D)` in the body of the first clause and the first goal `between(D0,D1,D2)` in the second clause of $R_{g'}$ are moved to the last position in their bodies respectively. $DB_{g''}$ and $DB_{g'}$ are logically equivalent, and semantically equivalent as well from the viewpoint of distribution semantics. Then think of the sampling execution by the OLDT interpreter of a top-goal $\leftarrow$ `q'(S)` w.r.t. $DB_{g''}$ where `S` is a variable, using the multi-stage depth-first search strategy. It is easy to see first that the execution never fails, and second that when the OLDT refutation terminates, a sentence `[`$w_1, \ldots, w_L$`]` is returned in `S`, and third that conversely, the set of `msw` atoms resolved upon in the refutation uniquely determines the output sentence `[`$w_1, \ldots, w_L$`]`.[52] Hence, if the sampling execution is guaranteed to always terminate, every sampling from $P_{F_{g''}}$ $(= P_{F_{g'}})$ uniquely generates a sentence, an observable atom, so the uniqueness condition is satisfied by $DB_{g''}$, and hence by $DB_{g'}$.

Then when is the sampling execution guaranteed to always terminate? In other words, when does the grammar only generate finite sentences? Giving a general answer seems difficult, but it is known that if the parameter values in a PCFG are obtained by learning from finite sentences, the stochastic derivation by the PCFG terminates with probability one (Chi & Geman, 1998). In summary, assuming appropriate parameter values, we can say that the parameterized logic program $DB_{g'}$ for the largest PCFG with $N$ nonterminal symbols satisfies all applicability conditions, and the OLDT time for a sentence of length $L$ is $O(N^3L^3)$[53] and this is also the size of the support graph. From Proposition 5.1, we conclude

**Proposition 5.3** *Let DB be a parameterized logic program representing a PCFG with $N$ nonterminals in the form of $DB'_g$ in Figure 11, and $\mathcal{G} = G_1, G_2, \ldots, G_T$ be the sampled atoms representing sentences of length $L$. We suppose each table operation in OLDT search is done in $O(1)$ time. Then OLDT search for $\mathcal{G}$ and one iteration in* learn-gEM *are respectively done in $O(N^3L^3T)$ time.*

51. `between(D0,D1,D2)` is called $O(N(L-d)^2)$ times in $T_d^{(1)}$. So it is called $\sum_{d=0}^{L-3} O(N(L-d)^2) = O(NL^3)$ times in $T_0^{(1)}$.
52. Because the trial-ids used in the refutation record which rule is used at what step in the derivation of $w_1 \cdots w_L$.
53. In $DB_{g'}$, we represent integers by ground terms made out of `0` and `s(·)` to keep the program short. If we use integers instead of ground terms however, the first three arguments of `q'(·,·,·,·,·)` are enough to check whether the goal is previously called or not, and this check can be done in $O(1)$ time.

$O(N^3L^3T)$ is also the time complexity of the Inside-Outside algorithm per iteration (Lari & Young, 1990), hence our algorithm is as efficient as the Inside-Outside algorithm.

### 5.4 Pseudo PCSGs

PCFGs can be improved by making choices context-sensitive, and one of such attempts is pseudo PCSGs (pseudo probabilistic context sensitive grammars) in which a rule is chosen probabilistically depending on both the nonterminal to be expanded and its parent nonterminal (Charniak & Carroll, 1994).

A pseudo PCSG is easily programmed. We add one extra-argument, `N`, representing the parent node, to the predicate `q'(I,D0,D2,C0,C2,L0-L2)` in Figure 11 and replace `msw(I,C0,[J,K])` with `msw([N,I],C0,[J,K])`. Since the (leftmost) derivation of a sentence from a pseudo PCSG is still a sequential decision process described by the modified program, the graphical EM algorithm applied to the support graphs generated from the modified program and observed sentences correctly performs the ML estimation of parameters in the pseudo PCSG.

A pseudo PCSG is thought to be a PCFG with rules of the form $[n,i] \rightarrow [i,j][i,k]$ $(1 \leq n,i,j,k \leq N)$ where $n$ is the parent nonterminal of $i$, so the arguments in the previous subsection are carried over with minor changes. We therefore have (details omitted)

**Proposition 5.4** *Let DB be a parameterized logic program for a pseudo PCSG with $N$ nonterminals as shown above, and $\mathcal{G} = G_1, G_2, \ldots, G_T$ the observed atoms representing sampled sentences of length $L$. Suppose each table operation in OLDT search can be done in $O(1)$ time. Then OLDT search for $\mathcal{G}$ and each iteration in* learn-gEM *is completed in $O(N^4L^3T)$ time.*

### 5.5 Bayesian Networks

A relationship between cause $C$ and its effect $E$ is often probabilistic such as the one between diseases and symptoms, and as such it is mathematically captured as the conditional probability $P(E = e \mid C = c)$ of effect $e$ given the cause $c$. What we wish to know however is the inverse, i.e. the probability of a candidate cause $c$ given evidence $e$, i.e. $P(C = c \mid E = e)$ which is calculated by Bayes' theorem as $P(E = e \mid C = c)P(C = c)/\sum_{c'} P(E = e \mid C = c')P(C = c')$. Bayesian networks are a representational/computational framework that fits best this type of probabilistic inference (Pearl, 1988; Castillo et al., 1997).

A Bayesian network is a graphical representation of a joint distribution $P(X_1 = x_1, \ldots, X_N = x_N)$ of finitely many random variables $X_1, \ldots, X_N$. The graph is a dag (directed acyclic graph) such as ones in Figure 12, and each node is a random variable.[54]

In the graph, a conditional probability table (CPT) representing $P(X_i = x_i \mid \mathbf{\Pi}_i = \boldsymbol{u}_i)$ $(1 \leq i \leq N)$ is associated with each node $X_i$ where $\mathbf{\Pi}_i$ represents $X_i$'s parent nodes and $\boldsymbol{u}_i$ their values. When $X_i$ has no parent, i.e. a topmost node in the graph, the table is just a marginal distribution $P(X_i = x_i)$. The whole joint distribution is defined as the product of

54. We only deal with discrete cases.

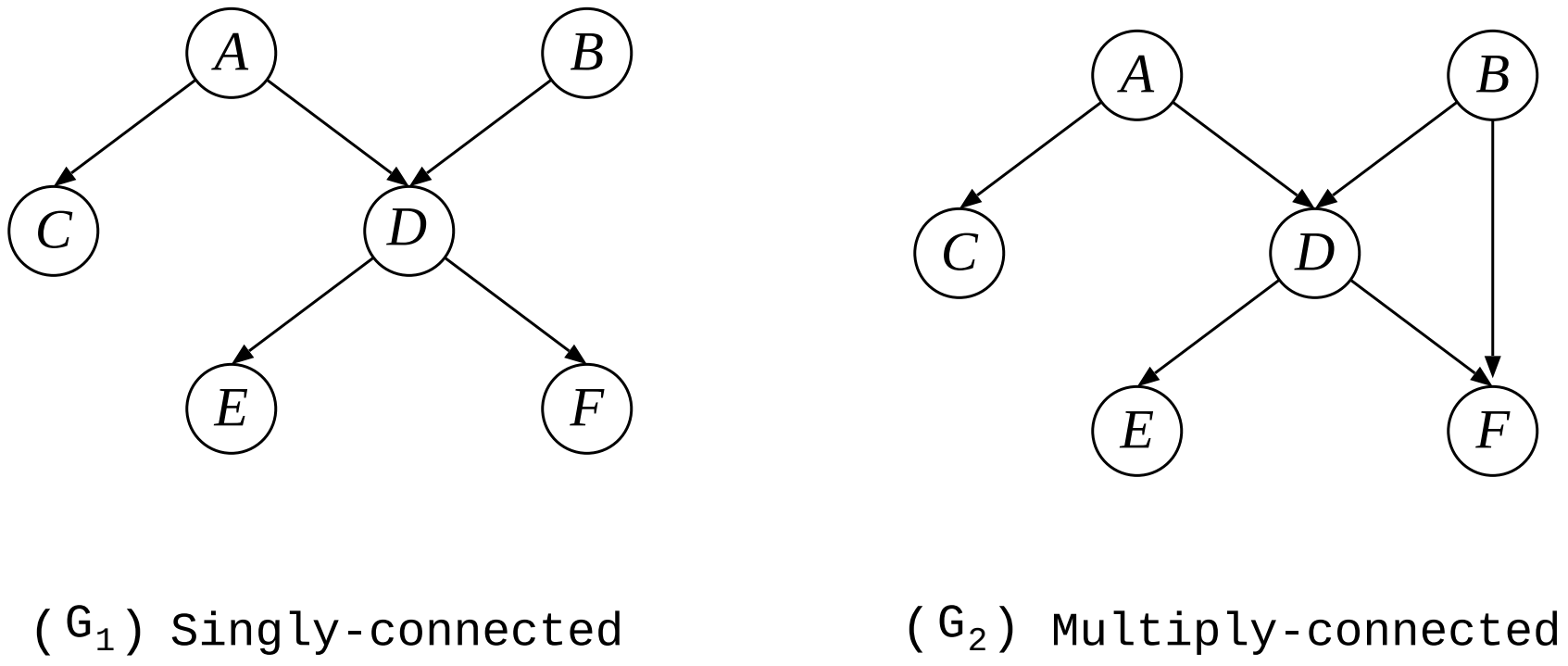

Figure 12: Bayesian networks

these conditional distributions:

$$P(X_1 = x_1, \ldots, X_N = x_N)^{55} = \prod_{i=1}^{N} P(X_i = x_i \mid \mathbf{\Pi}_i = \boldsymbol{u}_i). \tag{21}$$

Thus the graph $G_1$ in Figure 12 defines

$$P_{G_1}(a,b,c,d,e,f) = P_{G_1}(a)P_{G_1}(b)P_{G_1}(c \mid a)P_{G_1}(d \mid a,b)P_{G_1}(e \mid d)P_{G_1}(f \mid d)$$

where $a$, $b$, $c$, $d$, $e$ and $f$ are values of corresponding random variables $A$, $B$, $C$, $D$, $E$ and $F$, respectively.[56] As mentioned before, one of the basic tasks of Bayesian networks is to compute marginal probabilities. For example, the marginal distribution $P_{G_1}(c,d)$ is computed either by (22) or (23) below.

$$P_{G_1}(c,d) = \sum_{a,b,e,f} P_{G_1}(a)P_{G_1}(b)P_{G_1}(c \mid a)P_{G_1}(d \mid a,b)P_{G_1}(e \mid d)P(f \mid d) \tag{22}$$

$$= \left(\sum_{a,b} P_{G_1}(a)P_{G_1}(b)P_{G_1}(c \mid a)P_{G_1}(d \mid a,b)\right)\left(\sum_{e,f} P_{G_1}(e \mid d)P_{G_1}(f \mid d)\right) \tag{23}$$

(23) is clearly more efficient than (22). Observe that if the graph were like $G_2$ in Figure 12, there would be no way to factorize computations like (23) but to use (22) requiring exponentially many operations. The problem is that computing marginal probabilities is NP-hard in general, and factorization such as (23) is assured only when the graph is *singly connected* like $G_1$, i.e. has no loop when viewed as an undirected graph. In such case, the computation is possible in $O(|V|)$ time where $V$ is the set of vertices in the graph (Pearl, 1988). Otherwise, the graph is called *multiply-connected*, and might need exponential time to compute marginal probabilities. In the sequel, we show the following.

- For any discrete Bayesian network $G$ defining a distribution $P_G(x_1, \ldots, x_N)$, there is a parameterized logic program $DB_G$ for a predicate $\mathtt{bn}(\cdot)$ such that $P_{DB_G}(\mathtt{bn(}x_1\mathtt{,}\ldots\mathtt{,}x_N\mathtt{)})$ $= P_G(x_1, \ldots, x_N)$.

55. Thanks to the acyclicity of the graph, without losing generality, we may assume that if $X_i$ is an ancestor node of $X_j$, then $i < j$ holds.

56. For notational simplicity, we shall omit random variables when no confusion arises.

- For arbitrary factorizations and their order to compute a marginal distribution, there exists a tabled program that accomplishes the same computation in the specified way.
- When the graph is singly connected and evidence $e$ is given, there exists a tabled program $DB_{G^{\mathcal{T}}}$ such that OLDT time for $\leftarrow$ bn($e$) is $O(|V|)$, and hence the time complexity per iteration of the graphical EM algorithm is $O(|V|)$ as well.

Let $G$ be a Bayesian network defining a joint distribution $P_G(x_1, \ldots, x_N)$ and $\{P_G(X_i = x_i \mid \mathbf{\Pi}_i = \boldsymbol{u}_i) \mid 1 \leq i \leq N,\ x_i \in val(X_i),\ \boldsymbol{u}_i \in val(\mathbf{\Pi}_i)\}$ the conditional probabilities associated with $G$ where $val(X_i)$ is the set of $X_i$'s possible values and $val(\mathbf{\Pi}_i)$ denotes the set of possible values of the parent nodes $\mathbf{\Pi}_i$ as a random vector, respectively. We construct a parameterized logic program that defines the same distribution $P_G(x_1, \ldots, x_N)$. Our program $DB_G = F_G \cup R_G$ is shown in Figure 13.

$$F_G = \{\ \texttt{msw(par(}i\texttt{,}\boldsymbol{u}_i\texttt{),once,}x_i\texttt{)} \ \mid 1 \leq i \leq N, \boldsymbol{u}_i \in val(\mathbf{\Pi}_i), x_i \in val(X_i)\ \}$$

$$R_G = \{\ \texttt{bn(X}_1\texttt{,...,X}_N\texttt{):-}\textstyle\bigwedge_{i=1}^{N}\texttt{msw(par(}i\texttt{,}\mathbf{\Pi}_i\texttt{),once,X}_i\texttt{).}\ \}$$

Figure 13: Bayesian network program $DB_G$

$F_G$ is comprised of msw atoms of the form msw(par($i$,$\boldsymbol{u}_i$),once,$x_i$) whose probability is exactly the conditional probability $P_G(X_i = x_i \mid \mathbf{\Pi}_i = \boldsymbol{u}_i)$. When $X_i$ has no parents, $\boldsymbol{u}_i$ is the empty list []. $R_G$ is a singleton, containing only one clause whose body is a conjunction of msw atoms which corresponds to the product of conditional probabilities. Note that we *intentionally* identify random variables $X_1, \ldots, X_N$ with logical variables $\texttt{X}_1, \ldots, \texttt{X}_N$ for convenience.

**Proposition 5.5** *$DB_G$ denotes the same distributions as $G$.*

(Proof) Let $\langle x_1, \ldots, x_N \rangle$ be a realization of the random vector $\langle X_1, \ldots, X_N \rangle$. It holds by construction that

$$\begin{aligned} P_{DB_G}(\texttt{bn(}x_1\texttt{,...,}x_N\texttt{))} &= \prod_{h=1}^{N} P_{\texttt{msw}}(\texttt{msw(par(}i\texttt{,}\boldsymbol{u}_i\texttt{),once,}x_i\texttt{))} \\ &= \prod_{h=1}^{N} P_G(X_i = x_i \mid \mathbf{\Pi}_i = \boldsymbol{u}_i) \\ &= P_G(x_1, \ldots, x_N). \qquad Q.E.D. \end{aligned}$$

In the case of $G_1$ in Figure 12, the program becomes[57]

```
bn(A,B,C,D,E,F)  :-  msw(par('A',[]),once,A),   msw(par('B',[]),once,B),
                     msw(par('C',[A]),once,C),  msw(par('D',[A,B]),once,D),
                     msw(par('E',[D]),once,E),  msw(par('F',[D]),once,F).
```

57. 'A', 'B', ... are Prolog constants used in place of integers.

and the left-to-right sampling execution gives a sample realization of the random vector $\langle$ A,B,C,D,E,F $\rangle$. A marginal distribution is computed from bn($x_1$,...,$x_N$) by adding a new clause to $DB_G$. For example, to compute $P_{G_1}(c,d)$, we add bn(C,D):- bn(A,B,C,D,E,F) to $DB_{G_1}$ (let the result be $DB'_{G_1}$) and then compute $P_{DB'_{G_1}}(\texttt{bn}(c,d))$ which is equal to $P_{G_1}(c,d)$ because

$$\begin{aligned} P_{DB'_{G_1}}(\texttt{bn}(c,d)) &= P_{DB_{G_1}}(\exists\, a,b,e,f\ \texttt{bn}(a,b,c,d,e,f)) \\ &= \sum_{a,b,e,f} P_{DB_{G_1}}(\texttt{bn}(a,b,c,d,e,f)) \\ &= P_{G_1}(c,d). \end{aligned}$$

Regrettably this computation corresponds to (22), not to the factorization (23). Efficient probability computation using factorization is made possible by carrying out summations in a proper order.

We next sketch by an example how to carry out specified summations in a specified order by introducing new clauses. Suppose we have a joint distribution $P(x,y,z,w) = \phi_1(x,y)\phi_2(y,z,w)\phi_3(x,z,w)$ such that $\phi_1(x,y)$, $\phi_2(y,z,w)$ and $\phi_3(x,z,w)$ are respectively computed by atoms $\texttt{p}_1$(X,Y), $\texttt{p}_2$(Y,Z,W) and $\texttt{p}_3$(X,Z,W). Suppose also that we hope to compute the sum

$$P(x) = \sum_y \phi_1(x,y)\left(\sum_{z,w} \phi_2(y,z,w)\phi_3(x,z,w)\right)$$

in which we first eliminate $z,w$ and then $y$. Corresponding to each elimination, we introduce two new predicates, q(X,Y) to compute $\phi_4(x,y) = \sum_{z,w}\phi_2(y,z,w)\phi_3(x,z,w)$ and p(X) to compute $P(x) = \sum_y \phi_1(x,y)\phi_4(x,y)$ as follows.

```
  p(X)   :-  p1(X,Y), q(X,Y).
q(X,Y)   :-  p2(Y,Z,W), p3(X,Z,W).
```

Note that the clause body of q/2 contains Z and W as (existentially quantified) local variables and the clause head q(X,Y) contains variables shared with *other atoms*. In view of the correspondence between $\sum$ and $\exists$, it is easy to confirm that this program realizes the required computation. It is also easy to see by generalizing this example, though we do not prove here, that there exists a parameterized logic program that carries out the given summations in the given order for an arbitrary Bayesian network, in particular we are able to simulate VE (variable elimination, Zhang & Poole, 1996; D'Ambrosio, 1999) in our approach.

Efficient computation of marginal distributions is not always possible but there is a well-known class of Bayesian networks, singly connected Bayesian networks, for which there exists an efficient algorithm to compute marginal distributions by message passing (Pearl, 1988; Castillo et al., 1997). We here show that when the graph is singly connected, we can construct an efficient tabled Bayesian network program $DB_{G^\tau}$ assigning a table predicate to each node. To avoid complications, we explain the construction procedure informally and concentrate on the case where we have only one interested variable. Let $G$ be a singly connected graph. First we pick up a node $U$ whose probability $P_G(u)$ is what we seek. We construct a tree $G^\tau$ with the root node $U$ from $G$, by letting other nodes dangling from $U$. Figure 14 shows how $G_1$ is transformed to a tree when we select node B as the root node.

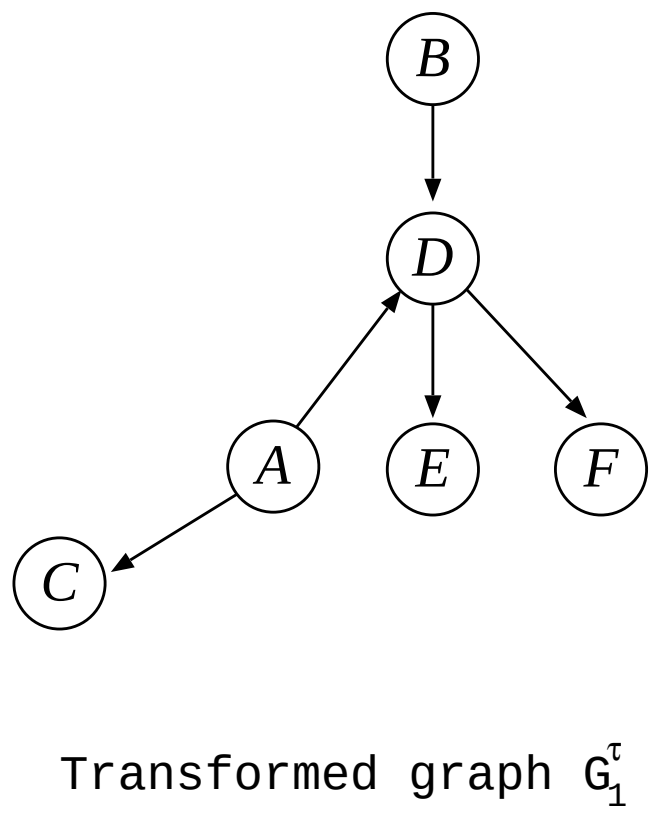

Figure 14: Transforming $G_1$ to a tree

Then we examine each node in $G^\tau$ one by one. We add for each node $X$ in the graph a corresponding clause to $DB_{G^\tau}$ whose purpose is to visit all nodes connected to $X$ except the one that calls $X$. Suppose we started from the root node $U_1$ in Figure 15 where evidence $u$ is given, and have generated clause (24). Now we proceed to an inner node $X$ ($U_1$ calls $X$). In the original graph $G$, $X$ has parent nodes $\{U_1, U_2, U_3\}$ and child nodes $\{V_1, V_2\}$. $U_3$ is a topmost node in $G$.

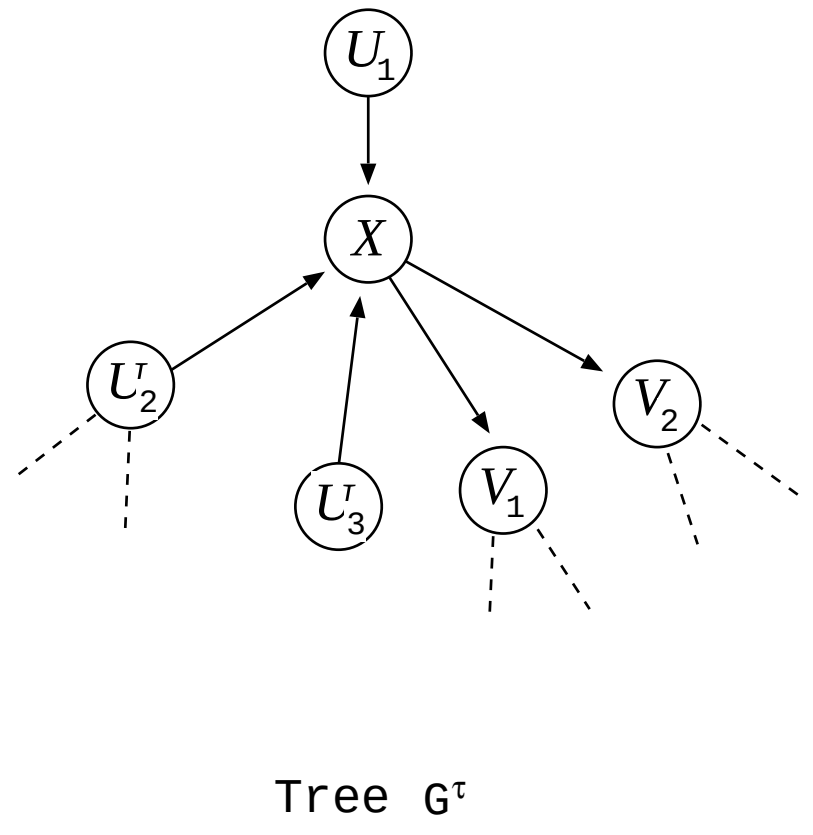

Figure 15: General situation

For node $X$ in Figure 15, we add clause (25). When it is called from the parent node $U_1$ with $\texttt{U}_1$ being ground, we first generate possible values of $U_2$ by calling $\texttt{val_}U_2\texttt{(U}_2\texttt{)}$, and then call $\texttt{call_}X\texttt{_}U_2\texttt{(U}_2\texttt{)}$ to visit all nodes connected to $X$ through $U_2$. $U_3$ is similary treated. After visiting all nodes in $G$ connecting to $X$ through the parent nodes $U_2$ and $U_3$ (nodes connected to $U_1$ have already been visited), the value of random variable $X$ is determined by sampling the `msw` atom jointly indexed by 'X' and the values of $U_1$, $U_2$ and $U_3$. Then we visit $X$'s children, $V_1$ and $V_2$. For a topmost node $U_3$ in the original graph, we add clause (26).

$$\texttt{tbn(U}_1\texttt{)} \quad \texttt{:-} \quad \texttt{msw(par('U}_1\texttt{',[]),once,U}_1\texttt{), call_}U_1\texttt{_}X\texttt{(U}_1\texttt{).} \tag{24}$$

$$\begin{aligned}\texttt{call_}U_1\texttt{_}X\texttt{(U}_1\texttt{)} \quad \texttt{:-} \quad & \texttt{val_}U_2\texttt{(U}_2\texttt{), call_}X\texttt{_}U_2\texttt{(U}_2\texttt{),}\\ & \texttt{val_}U_3\texttt{(U}_3\texttt{), call_}X\texttt{_}U_3\texttt{(U}_3\texttt{),}\\ & \texttt{msw(par('X',[U}_1\texttt{,U}_2\texttt{,U}_3\texttt{]),once,X),}\\ & \texttt{call_}X\texttt{_}V_1\texttt{(X), call_}X\texttt{_}V_2\texttt{(X).}\end{aligned} \tag{25}$$

$$\texttt{call_}X\texttt{_}U_3\texttt{(U}_3\texttt{)} \quad \texttt{:-} \quad \texttt{msw(par('U}_3\texttt{',[]),once,U}_3\texttt{).} \tag{26}$$

Let $DB_{G^\tau}$ be the final program containing clauses like (24), (25) and (26). Apparently $DB_{G^\tau}$ can be constructed in time linear in the number of nodes in the network. Also note that successive unfolding (Tamaki & Sato, 1984) of atoms of the form `call_...(·)` in the clause bodies that starts from (24) yields a program $DB'_G$ similar to the one in Figure 13 which contains `msw` atoms but no `call_...(·)`'s. As $DB_{G^\tau}$ and $DB'_G$ define the same distribution,[58] it can be proved from Proposition 5.5 that $P_G(u) = P_{DB'_G}(\texttt{bn}(u)) = P_{DB_{G^\tau}}(\texttt{tbn}(u))$ holds (details omitted). By the way, in Figure 15 we assume the construction starts from the topmost node $\texttt{U}_1$ where the evidence $u$ is given, but this is not necessary. Suppose we change to start from the inner node `X`. In that case, we replace clause (24) with $\texttt{call_}X\texttt{_}U_1\texttt{(U}_1\texttt{) :- msw(par('U}_1\texttt{',[]),once,U}_1\texttt{)}$ just like (26). At the same time we replace the head of clause (25) with `tbn()` and add a goal $\texttt{call_}X\texttt{_}U_1(u)$ to the body and so on. For the changed program $DB''_{G^\tau}$, it is rather straightforward to prove that $P_{DB''_{G^\tau}}(\texttt{tbn()}) = P_G(u)$ holds. It is true that the construction of the tabled program $DB_{G^\tau}$ shown here is very crude and there is a lot of room for optimization, but it suffices to show that a parameterized logic program for a singly connected Bayesian network runs in $O(|V|)$ time where $V$ is the set of nodes.

To estimate time complexity of OLDT search w.r.t. $DB_{G^\tau}$, we declare `tbn` and every predicate of the form `call_...(·)` as table predicate and verify the five conditions on the applicability of the graphical EM algorithm (details omitted). We now estimate the time complexity of OLDT search for the goal $\leftarrow\texttt{tbn}(u)$ w.r.t. $DB_{G^\tau}$.[59] We notice that calls occur according to the pre-order scan (parents – the node – children) of the tree $G^\tau$, and calls to $\texttt{call_}Y\texttt{_}X(\cdot)$ occur $val(Y)$ times. Each call to $\texttt{call_}Y\texttt{_}X(\cdot)$ invokes calls to the rest of nodes, $X$'s parents and $X$'s children in the graph $G^\tau$ except the caller node, with diffrent set of variable instantiations, but from the second call on, every call only refers to solutions stored in the solution table in $O(1)$ time. Thus, the number of added computation steps in

58. Since distribution semantics is based on the least model semantics, and because unfold/fold transformation (Tamaki & Sato, 1984) preserves the least Herbrand model of the transformed program, unfold/fold transformation applied to parameterized logic programs preserves the denotation of the transformed program.

59. $DB_{G^\tau}$ is further transformed for the OLDT interpreter to collect `msw` atoms like the case of the HMM program.

OLDT search by $X$ is bounded from above, by constant $O(val(U1)val(U2)val(U3)val(X))$ in the case of Figure 15. As a result OLDT time is proportional to the number of nodes in the original graph $G$ (and so is the size of the support graph) provided that the number of edges connecting to a node, and that of values of a random variable are bounded from above. So we have

**Proposition 5.6** *Let $G$ be a singly connected Bayesian network defining distribution $P_G$, $V$ the set of nodes, and $DB_{G^T}$ the tabled program derived as above. Suppose the number of edges connecting to a node, and that of values of a random variable are bounded from above by some constant. Also suppose table access can be done in $O(1)$ time. Then, OLDT time for computing $P_G(u)$ for an observed value $u$ of a random variable $U$ by means of $DB_{G^T}$ is $O(|V|)$ and so is time per iteration required by the graphical EM algorithm. If there are $T$ observations, time complexity is $O(|V|T)$.*

$O(|V|)$ is the time complexity required to compute a marignal distribution for a singly connected Bayesian network by a standard algorithm (Pearl, 1988; Castillo et al., 1997), and also is that of the EM algorithm using it. We therefore conclude that the graphical EM algorithm is as efficient as a specialzed EM algorithm for singly connected Bayesian networks.[60] We must also quickly add that the graphical EM algorithm is applicable to arbitrary Bayesian networks,[61] and what Proposition 5.6 says is that an explosion of the support graph can be avoided by appropriate programming in the case of singly connected Bayesian networks.

To summarize, the graphical EM algorithm, a single generic EM algorithm, is proved to have the same time complexity as specialized EM algorithms, i.e. the Baum-Welch algorithm for HMMs, the Inside-Outside algorithm for PCFGs, and the one for singly connected Bayesian networks that have been developed independently in each research field.

Table 1 summarizes the time complexity of EM learning using OLDT search and the graphical EM algorithm in the case of one observation. In the first column, "sc-BNs" represents singly connected Bayesian networks. The second column shows a program to use. $DB_h$ is an HMM proram in Subsection 4.7, $DB_{g'}$ a PCFG program in Subsection 5.3 and $DB_{G^T}$ a transformed Bayesian network program in Subsection 5.5, respectively. OLDT time in the third column is time for OLDT search to complete the search of all t-explanations. gEM in the fourth column is time in one iteration taken by the graphical EM algorithm to update parameters. We use $N$, $M$, $L$ and $V$ respectively for the number of states in an HMM, the number of nonterminals in a PCFG, the length of an input string and the number of nodes in a Bayesian network. The last column is a standard (specialized) EM algorithm for each model.

60. When a marginal distribution of $P_G$ for more than one variable is required, we can construct a similar tabled program that computes marginal probabilities still in $O(|V|)$ time by adding extra-arguments that convey other evidence or by embedding other evidnece in the program.
61. We check the five conditions with $DB_G$ in Figure 13. The uniqueness condition is obvious as sampling always uniquely generates a sampled value for each random variable. The finite support condition is satisfied because there are only a finite number of random variables and their values. The acyclic support condition is immediate because of the acyclicity of Bayesian networks. The t-exclusiveness condition and the independent condition are easy to verify.

| Model | Program | OLDT time | gEM | Specialized EM |
|---|---|---|---|---|
| HMMs | $DB_h$ | $O(N^2L)$ | $O(N^2L)$ | Baum-Welch |
| PCFGs | $DB_{g'}$ | $O(M^3L^3)$ | $O(M^3L^3)$ | Inside-Outside |
| sc-BNs | $DB_{G^\tau}$ | $O(\|V\|)$ | $O(\|V\|)$ | (Castillo et al., 1997) |
| user model | | $O(\|\text{OLDT tree}\|)$ | $O(\|\text{support graph}\|)$ | |

Table 1: Time complexity of EM learning by OLDT search and the graphical EM algorithm

### 5.6 Modeling Language PRISM

We have been developing a symbolic-statistical modeling laguage PRISM since 1995 (URL = `http://mi.cs.titech.ac.jp/prism/`) as an implementation of distribution semantics (Sato, 1995; Sato & Kameya, 1997; Sato, 1998). The language is intented for modeling complex symbolic-statistical phenomena such as discourse interpretation in natural language processing and gene inheritance interacting with social rules. As a programming language, it looks like an extension of Prolog with new built-in predicates including the `msw` predicate and other special predicates for manipulating `msw` atoms and their parameters.

A PRISM program is comprised of three parts, one for directives, one for modeling and one for utilities. The directive part contains declarations such as `values`, telling the system what `msw` atoms will be used in the execution. The modeling part is a set of non-unit definite clauses that define the distribution (denotation) of the program by using `msw` atoms. The last part, the utility part, is an arbitary Prolog program which refers to predicates defined in the modeling part. We can use in the utility part `learn` built-in predicate to carry out EM learning from observed atoms.

PRISM provides three modes of execution. The sampling execution correponds to a random sampling drawn from the distribution defined by the modeling part. The second one computes the probability of a given atom. The third one returns the support set for a given goal. These execution modes are available through built-in predicates.

We must report however that while the implementation of the graphical EM algorithm with a simpified OLDT search mechanism has been under way, it is not completed yet. So currently, only Prolog search and *learn-naive*$(DB, \mathcal{G})$ in Section 4 are available for EM learning though we realized, partially, structrure sharing of explanations in the implemention of *learn-naive*$(DB, \mathcal{G})$. Putting computational efficiecy aside however, there is no problem in expressing and learning HMMs, PCFGs, pseudo PCSGs, Bayesian networks and other probailistic models by the current version. The learning experiments in the next section used a parser as a substitute for the OLDT interpreter, and the independently implemented graphical EM algorithm.

## 6. Learning Experiments

After complexity analysis of the graphical EM algorithm for popular symbolic-probabilistic models in the previous section, we look at an actual behavior of the graphical EM algorithm with real data in this section. We conducted learning experiments with PCFGs using two corpora which have contrasting characters, and compared the performance of the graphical EM algorithm against that of the Inside-Outside algorithm in terms of time per iteration (= time for updating parameters). The results indicate that the graphical EM algorithm can outperform the Inside-Outside algorithm by orders of magnigude. Detalis are reported by Sato, Kameya, Abe, and Shirai (2001). Before proceeding, we review the Inside-Outside algorithm for completeness.

### 6.1 The Inside-Outside Algorithm

The Inside-Outside algorithm was proposed by Baker (1979) as a generalization of the Baum-Welch algorithm to PCFGs. The algorithm is designed to estimate parameters for a CFG grammar in Chomsky normal form containing rules expressed by numbers like $i \rightarrow j, k$ ($1 \leq i, j, k \leq N$ for $N$ nonterminals, where 1 is a starting symbol). Suppose an input sentence $w_1, \ldots, w_L$ is given. In each iteration, it first computes in a bottom up manner inside probabilities $e(s,t,i) = P(i \stackrel{*}{\Rightarrow} w_s, \ldots, w_t)$ and then computes outside probabilities $f(s,t,i) = P(S \stackrel{*}{\Rightarrow} w_1, \ldots, w_{s-1}\ i\ w_{t+1}, \ldots, w_L)$ in a top-down manner for every $s$, $t$ and $i$ ($1 \leq s \leq t \leq L, 1 \leq i \leq N$). After computing both probabilities, parameters are updated by using them, and this process iterates until some predetermined criterion such as a convergence of the likelihood of the input sentence is achieved. Although Baker did not give any analysis of the Inside-Outside algorithm, Lari and Young (1990) showed that it takes $O(N^3 L^3)$ time in one iteration and Lafferty (1993) proved that it is the EM algorithm.

While it is true that the Inside-Outside algorithm has been recognized as a standard EM algortihm for training PCFGs, it is notoriously slow. Although there is not much literature explicitly stating time required by the Inside-Outside algorithm (Carroll & Rooth, 1998; Beil, Carroll, Prescher, Riezler, & Rooth, 1999), Beil et al. (1999) reported for example that when they trained a PCFG with 5,508 rules for a corpus of 450,526 German subordinate clauses whose average ambiguity is 9,202 trees/clause using four machines (167MHz Sun UltraSPARC×2 and 296MHz Sun UltraSPARC-II×2), it took 2.5 hours to complete *one iteration*. We discuss later why the Inside-Outside algorithm is slow.

### 6.2 Learning Experiments Using Two Corpora

We report here parameter learning of existing PCFGs using two corpora of moderate size and compare the graphical EM algorithm against the Inside-Outside algorithm in terms of time per iteration. As mentioned before, support graphs, input to the garphical EM algorithm, were generated by a parser, i.e. MSLR parser.[62] All measurements were made on a 296MHz Sun UltraSPARC-II with 2GB memory under Solaris 2.6 and the threshold for an increase of the log likelihood of input sentences was set to $10^{-6}$ as a stopping criterion for the EM algorithms.

In the experiments, we used ATR corpus and EDR corpus (each converted to a POS (part of speech)-tagged corpus). They are similar in size (about 10,000) but contrasting in their characters, sentence length and ambiguity of their grammars. The first experiment employed ATR corpus which is a Japanese-English corpus (we used only the Japanese part) developed by ATR (Uratani, Takezawa, Matsuo, & Morita, 1994). It contains 10,995 short

62. MSLR parser is a Tomita (Generalized LR) parser developed by Tanaka-Tokunaga Laboratory in Tokyo Institute of Technology (Tanaka, Takezawa, & Etoh, 1997).

conversational sentences, whose minimum length, average length and maximum length are respectively 2, 9.97 and 49. As a skeleton of PCFG, we employed a context free grammar $G_{\mathrm{atr}}$ comprising 860 rules (172 nonterminals and 441 terminals) manually developed for ATR corpus (Tanaka et al., 1997) which yields 958 parses/sentence.

Because the Inside-Outside algorithm only accepts a CFG in Chomsky normal form, we converted $G_{\mathrm{atr}}$ into Chomsky normal form $G^*_{\mathrm{atr}}$. $G^*_{\mathrm{atr}}$ contains 2,105 rules (196 nonterminals and 441 terminals). We then divided the corpus into subgroups of similar length like $(L = 1, 2), (L = 3, 4), \ldots, (L = 25, 26)$, each containing randomly chosen 100 sentences. After these preparations, we compare at each length the graphical EM algorithm applied to $G_{\mathrm{atr}}$ and $G^*_{\mathrm{atr}}$ against the Inside-Outside algorithm applied to $G^*_{\mathrm{atr}}$ in terms of time per iteration by running them until convergence.

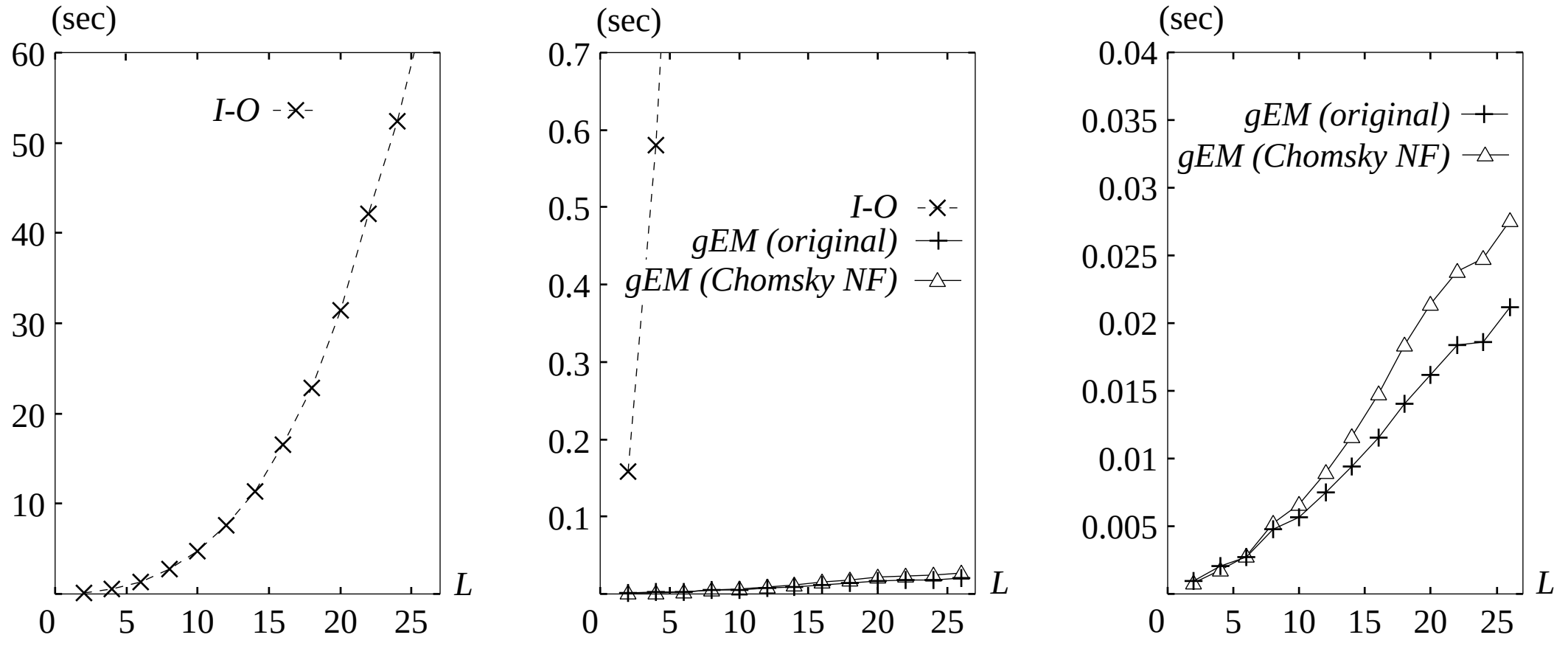

Figure 16: Time per iteration : I-O vs. gEM (ATR)

Curves in Figure 16 show the learning results where an x-axis is the length $L$ of an input sentence and a y-axis is average time taken by the EM algorithm in one iteration to update all parameters contained in the support graphs generated from the chosen 100 sentences (other parameters in the grammar do not change). In the left graph, the Inside-Outside algorithm plots a cubic curve labeled "I-O". We omitted a curve drawn by the graphical EM algorithm as it drew the x-axis. The middle graph magnifies the left graph. The curve labeled "gEM (original)" is plotted by the graphical EM algorithm applied to the original grammar $G_{\mathrm{atr}}$ whereas the one labeled "gEM (Chomsky NF)" used $G^*_{\mathrm{atr}}$. At length 10, the average sentence length, it is measured that whichever grammar is employed, the graphical EM algorithm runs several hundreds times faster (845 times faster in the case of $G_{\mathrm{atr}}$ and 720 times faster in the case of $G^*_{\mathrm{atr}}$) than the Inside-Outside algorithm per iteration. The right graph shows (almost) linear dependency of updating time by the graphical EM algorithm within the measuared sentence length.

Although some difference is anticipated in their learning speed, the speed gap between the Inside-Outside algorithm and the graphical EM algorithm is unexpectedly large. The most conceivable reason is that ATR corpus only contains short sentences and $G_{\mathrm{atr}}$ is not much ambiguous so that parse trees are sparse and generated support graphs are small, which affects favorably the perforamnce of the graphical EM algorithm.

We therefore conducted the same experiment with another corpus which contains much longer sentences using a more ambiguous grammar that generates dense parse trees. We used EDR Japanese corpus (Japan EDR, 1995) containing 220,000 Japanese news article sentences. It is however under the process of re-annotation, and only part of it (randomly sampled 9,900 sentences) has recently been made available as a labeled corpus. Compared with ATR corpus, sentences are much longer (the average length of 9,900 sentences is 20, the minimum length 5, the maximum length 63) and a CFG grammar $G_{\mathrm{edr}}$ (2,687 rules, converted to Chomsky normal form grammar $G^*_{\mathrm{edr}}$ containing 12,798 rules) developed for it is very ambiguous (to keep a coverage rate), having $3.0 \times 10^8$ parses/sentence at length 20 and $6.7 \times 10^{19}$ parses/sentence at length 38.

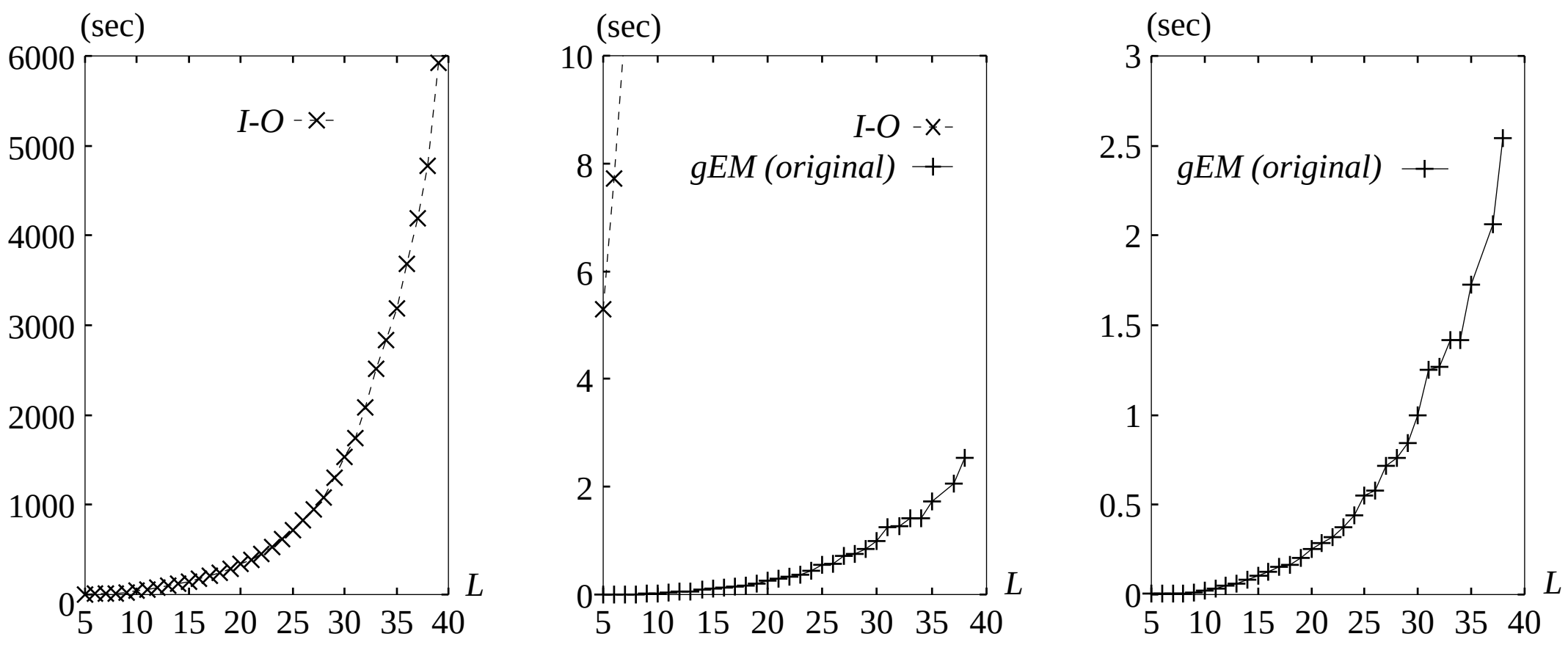

Figure 17: Time per iteration : I-O vs. gEM (EDR)

Figure 17 shows the obtained curves from the experiments with EDR corpus (the graphical EM algorithm applied to $G_{\mathrm{edr}}$ vs. the Inside-Outside algorithm applied to $G^*_{\mathrm{edr}}$) under the same condition as ATR corpus, i.e. plotting average time per iteration to process 100 sentences of the designated length, except that the plotted time for the Inside-Outside algorithm is the average of 20 iterations whereas that for the graphical EM algorithm is the average of 100 iterations. As is clear from the middle graph, this time again, the graphical EM algorithm runs orders of magnitude faster than the Inside-Outside algorithm. At average sentence length 20, the former takes 0.255 second whereas the latter takes 339 seconds, giving a speed ratio of 1,300 to 1. At sentence length 38, the former takes 2.541 seconds but the latter takes 4,774 seconds, giving a speed ratio of 1,878 to 1. Thus the speed ratio even widens compared to ATR corpus. This can be explained by the mixed effects of $O(L^3)$, time complexity of the Inside-Outside algorithm, and a moderate increase in the total size of support graphs w.r.t. $L$. Notice that the right graph shows how the total size of support graphs grows with sentence length $L$ as time per iteration by the graphical EM algorithm is linear in the total size of support graphs.

Since we implemented the Inside-Outside algorithm faithfully to Baker (1979), Lari and Young (1990), there is much room for improvement. Actually Kita gave a refined Inside-Outside algorithm (Kita, 1999). There is also an implementation by Mark Johnson of the Inside-Outside algorithm down-loadable from `http://www.cog.brown.edu/%7Emj/`. The use of such implementations may lead to different conclusions. We therefore conducted learning experiments with the *entire* ATR corpus using these two implementations and measured updating time per iteration (Sato et al., 2001). It turned out that both implementations run twice as fast as our naive implementation and take about 630 seconds per iteration while the graphical EM algorithm takes 0.661 second per iteration, which is still orders of magnitude faster than the former two. Regrettably a similar comparison using the *entire* EDR corpus available at the moment was abandoned due to memory overflow during parsing for the construction of support graphs.

Learning experiments so far only compared time per iteration which ignore extra time for search (parsing) required by the graphical EM algorithm. So a question naturally arises w.r.t. comparison in terms of total learning time. Assuming 100 iterations for learning ATR corpus however, it is estimated that even considering parsing time, the graphical EM algorithm combined with MSLR parser runs orders of magnitude faster than the three implementations (ours, Kita's and Johnson's) of the Inside-Outside algorithm (Sato et al., 2001). Of course this estimation does not directly apply to the graphical EM algorithm combined with OLDT search, as the OLDT interpreter will take more time than a parser and how much more time is needed depends on the implementaiton of OLDT search.[63] Conversely, however, we may be able to take it as a rough indication of how far our approach, the graphical EM algorithm combined with OLDT search via support graphs, can go in the domain of EM learning of PCFGs.

### 6.3 Examing the Performance Gap

In the previous subsection, we compared the performance of the graphical EM algorithm against the Inside-Outside algorithm when PCFGs are given, using two corpora and three implementations of the Inside-Outside algorithm. In all experiments, the graphical EM algorithm considerably outperformed the Inside-Outside algorithm despite the fact that both have the same time complexity. Now we look into what causes such a performance gap.

Simply put, the Inside-Outside algorithm is slow (primarily) because it lacks parsing. Even when a backbone CFG grammar is explicitly given, it does not take any advantage of the constraints imposed by the grammar. To see it, it might help to review how the inside probability $e(s,t,A)$, i.e. P(nonterminal $A$ spans from $s$-th word to $t$-th word) ($s \leq t$), is calculated by the Inside-Outside algorithm for the given grammar.

$$e(s,t,A) = \sum_{B,C \text{ s.t. } A\rightarrow BC \text{ in the grammar}} \sum_{r=s}^{r=t-1} \mathrm{P}(A \rightarrow BC)e(s,r,B)e(r+1,t,C)$$

Here $\mathrm{P}(A \rightarrow BC)$ is a probability associated with a production rule $A \rightarrow BC$. Note that for a fixed triplet $(s,t,A)$, it is usual that the term $\mathrm{P}(A \rightarrow BC)e(s,r,B)e(r+1,t,C)$ is non-zero

63. We cannot answer this question right now as the implementation of OLDT search is not completed.

only for a relatively small number of $(B, C, r)$'s determined from successful parses and the rest of combinations always give 0 to the term. Nonetheless the Inside-Outside algorithm attempts to compute the term in every iteration for all possible combinations of $B$, $C$ and $r$ and this is repeated for every possible $(s, t, A)$, resulting in a lot of redundancy. The same kind of redundancy occurs in the computation of outside probability by the Inside-Outside algorithm.

The graphical EM algorithm is free of such redundancy because it runs on parse trees (a parse forest) represented by the support graph.[64] It must be added, on the other hand, that superiority in learning speed of the graphical EM algorithm is realized at the cost of space complexity because while the Inside-Outside algorithm merely requires $O(NL^2)$ space for its array to store probabilities, the graphical EM algorithm needs $O(N^3L^3)$ space to store the support graph where $N$ is the number of nonterminals and $L$ is the sentence length. This trade-off is understandable if one notices that the graphical EM algorithm applied to a PCFG can be considered as partial evaluation of the Inside-Outside algorithm by the grammar (and the introduction of appropriate data structure for the output).

Finally we remark that the use of parsing as a preprocess for EM learning of PCFGs is not unique to the graphical EM algorithm (Fujisaki, Jelinek, Cocke, Black, & Nishino, 1989; Stolcke, 1995). These approaches however still seem to contain redundancies compared with the graphical EM algorithm. For instance Stolcke (1995) uses an Earley chart to compute inside and outside probability, but parses are implicitly reconstructed in each iteration dynamically by combining completed items.

## 7. Related Work and Discussion

### 7.1 Related Work

The work presented in this paper is at the crossroads of logic programming and probability theory, and considering an enormous body of work done in these fields, incompleteness is unavoidable when reviewing related work. Having said that, we look at various attempts made to integrate probability with computational logic or logic programming.[65] In reviewing, one can immediately notice there are two types of usage of probability. One type, *constraint approach*, emphasizes the role of probability as constraints and does not necessarily seek for a unique probability distribution over logical formulas. The other type, *distribution approach*, explicitly defines a unique distribution by model theoretical means or proof theoretical means, to compute various probabilities of propositions.

A typical constraint approach is seen in the early work of probabilistic logic by Nilsson (1986). His central problem, "probabilistic entailment problem", is to compute the upper and lower bound of probability $P(\phi)$ of a target sentence $\phi$ in such a way that the bounds are compatible with a given knowledge base containing logical sentences (not necessarily logic programs) annotated with a probability. These probabilities work as constraints on

64. We emphasize that the difference between the Inside-Outside algorithm and the graphical EM algorithm is solely computational efficiency, and they converge to the same parameter values when starting from the same initial values. Linguistic evaluations of the estimated parameters by the graphical EM algorithm are also reported by Sato et al. (2001).
65. We omit literature leaning strongly toward logic. For logic(s) concerning uncertainty, see an overview by Kyburg (1994).

the possible range of P($\phi$). He used the linear programming technique to solve this problem that inevitably delimits the applicability of his approach to finite domains.

Later Lukasiewicz (1999) investigated the computational complexity of the probabilistic entailment problem in a slightly different setting. His knowledge base comprises statements of the form $(H \mid G)[u_1, u_2]$ representing $u_1 \leq \mathrm{P}(H \mid G) \leq u_2$. He showed that inferring "tight" $u_1, u_2$ is NP-hard in general, and proposed a tractable class of knowledge base called conditional constraint trees.

After the influential work of Nilsson, Frish and Haddawy (1994) introduced a deductive system for probabilistic logic that remedies "drawbacks" of Nilsson's approach, that of computational intractability and the lack of a proof system. Their system deduces a probability range of a proposition by rules of probabilistic inferences about unconditional and conditional probabilities. For instance, one of the rules infers $P(\alpha \mid \delta) \in [0\ y]$ from $P(\alpha \vee \beta \mid \delta) \in [x\ y]$ where $\alpha$,$\beta$ and $\delta$ are propositional variables and $[x\ y]$ $(x \leq y)$ designates a probability range.

Turning to logic programming, probabilistic logic programming formalized by Ng and Subrahmanian (1992) and Dekhtyar and Subrahmanian (1997) was also a constraint approach. Their program is a set of annotated clauses of the form $A : \mu \leftarrow F_1 : \mu_1, \ldots, F_n : \mu_n$ where $A$ is an atom, $F_i$ $(1 \leq i \leq n)$ a basic formula, i.e. a conjunction or a disjunction of atoms, and $\mu_j$ $(0 \leq j \leq n)$ a sub-interval of $[0, 1]$ indicating a probability range. A query $\leftarrow \exists\, (F_1 : \mu_1, \ldots, F_n : \mu_n)$ is answered by an extension of SLD refutation. On formalization, it is assumed that their language contains only a finite number of constant and predicate symbols, and no function symbol is allowed.

A similar framework was proposed by Lakshmanan and Sadri (1994) under the same syntactic restrictions (finitely many constant and predicate symbols but no function symbols) in a different uncertainty setting. They used annotated clauses of the form $A \stackrel{c}{\leftarrow} B_1, \ldots, B_n$ where $A$ and $B_i$ $(1 \leq i \leq n)$ are atoms and $c = \langle[\alpha, \beta], [\gamma, \delta]\rangle$, a confidence level, represents a belief interval $[\alpha, \beta]$ $(0 \leq \alpha \leq \beta \leq 1)$ and a doubt interval $[\gamma, \delta]$ $(0 \leq \gamma \leq \delta \leq 1)$, which an expert has in the clause.

As seen above, defining a unique probability distribution is of secondary or no concern to the constraint approach. This is in sharp contrast with Bayesian networks as the whole discipline rests on the ability of the networks to define a unique probability distribution (Pearl, 1988; Castillo et al., 1997). Researchers in Bayesian networks have been seeking for a way of mixing Bayesian networks with a logical representation to increase their inherently propositional expressive power.

Breese (1992) used logic programs to automatically build a Bayesian network from a query. In Breese's approach, a program is the union of a definite clause program and a set of conditional dependencies of the form $\mathrm{P}(P \mid Q_1 \wedge \cdots \wedge Q_n)$ where $P$ and $Q_i$s are atoms. Given a query, a Bayesian network is constructed dynamically that connects the query and relevant atoms in the program, which in turn defines a *local* distribution for the connected atoms. Logical variables can appear in atoms but no function symbol is allowed.

Ngo and Haddawy (1997) extended Breese's approach by incorporating a mechanism reflecting context. They used a clause of the form $\mathrm{P}(A_0 \mid A_1, \ldots, A_n) = \alpha \leftarrow L_1, \ldots, L_k$, where $A_i$'s are called p-atoms (probabilistic atoms) whereas $L_j$'s are context atoms disjoint from p-atoms, and computed by another general logic program (satisfying certain restric-tions). Given a query, a set of evidence and context atoms, relevant ground p-atoms are identified by resolving context atoms away by SLDNF resolution, and a local Bayesian network is built to calculate the probability of the query. They proved the soundness and completeness of their query evaluation procedure under the condition that programs are acyclic[66] and domains are finite.

Instead of defining a local distribution for each query, Poole (1993) defined a *global* distribution in his "probabilistic Horn abduction". His program consists of definite clauses and *disjoint declarations* of the form `disjoint(`$[h_1{:}p_1,\ldots,h_n{:}p_n]$`)` which specifies a probability distribution over the hypotheses (abducibles) $\{h_1,\ldots,h_n\}$. He assigned probabilities to all ground atoms with the help of the theory of logic programming, and furthermore proved that Bayesian networks are representable in his framework. Unlike previous approaches, his language contains function symbols, but the acyclicity condition imposed on the programs for his semantics to be definable seems to be a severe restriction. Also, probabilities are not defined for quantified formulas.

Bacchus et al. (1996) used a much more powerful first-order probabilistic language than clauses annotated with probabilities. Their language allows a statistically quantified term such as $\| \phi(x) | \theta(x) \|_x$ to denote the ratio of individuals in a finite domain satisfying $\phi(x) \wedge \theta(x)$ to those satisfying $\theta(x)$. Assuming that every world (interpretation for their language) is equally likely, they define the probability of a sentence $\varphi$ under the given knowledge base $KB$ as the limit $\lim_{N\rightarrow\infty}\left(\frac{\#\mathrm{worlds}^{\tau}_{N}(\varphi\wedge KB)}{\#\mathrm{worlds}^{\tau}_{N}(KB)}\right)$ where $\#\mathrm{worlds}^{\tau}_{N}(\chi)$ is the number of possible worlds containing $N$ individuals satisfying $\chi$, and $\tau$ parameters used in judging approximations. Although the limit does not necessarily exist and the domain must be finite, they showed that their method can cope with difficulties arising from "direct inference" and default reasoning.

In a more linguistic vein, Muggleton (1996, and others) formulated SLPs (stochastic logic programs) procedurally, as an extension of PCFGs to probabilistic logic programs. So, a clause $C$, which must be range-restricted,[67] is annotated with a probability $p$ like $p : C$. The probability of a goal $G$ is the product of such $p$s appearing in its refutation but with a modification such that if a subgoal $g$ can invoke $n$ clauses, $p_i : C_i$ $(1 \leq i \leq n)$ at some refutation step, the probability of choosing $k$-th clause is normalized to $p_k / \sum_{i=1}^{n} p_i$.

More recently, Cussens (1999, 2001) enriched SLPs by introducing a special class of log-linear models for SLD refutations w.r.t. a given goal. He for example considers all possible SLD refutations for the most general goal $\leftarrow s(X)$ and defines probability $\mathrm{P}(R)$ of a refutation $R$ as $\mathrm{P}(R) = Z^{-1} \exp\left(\sum_i \lambda_i \nu(R,i)\right)$. Here $\lambda_i$ is a number associated with a clause $C_i$ and $\nu(R,i)$ is a feature, i.e. the number of occurrences of $C_i$ in $R$. $Z$ is the normalizing constant. Then, the probability assigned to $s(a)$ is the sum of probabilities of refutation for $\leftarrow s(a)$.

66. The condition says that every ground atom $A$ must be assigned a unique integer $n(A)$ such that $n(A) > n(B_1),\ldots,n(B_n)$ holds for any ground instance of a clause of the form $A \leftarrow B_1,\ldots,B_n$. Under this condition, when a program includes $p(X) \leftarrow q(X,Y)$, we cannot write recursive clauses about $q$ such as $q(X,[H|Y]) \leftarrow q(X,Y)$.

67. A syntactic property that variables appearing in the head also appear in the body of a clause. A unit clause must be ground.

### 7.2 Limitations and Potential Problems

Approaches described so far have more or less similar limitations and potential problems. Descriptive power confined to finite domains is the most common limitation, which is due to the use of the linear programming technique (Nilsson, 1986), or due to the syntactic restrictions not allowing for infinitely many constant, function or predicate symbols (Ng & Subrahmanian, 1992; Lakshmanan & Sadri, 1994). Bayesian networks have the same limitation as well (only a finite number of random variables are representable).[68] Also there are various semantic/syntactic restrictions on logic programs. For instance the acyclicity condition imposed by Poole (1993) and Ngo and Haddawy (1997) prevents the unconditional use of clauses with local variables, and the range-restrictedness imposed by Muggleton (1996) and Cussens (1999) excludes programs such as the usual membership Prolog program.

There is another type of problem, the possibility of assigning conflicting probabilities to logically equivalent formulas. In SLPs, $\mathrm{P}(A)$ and $\mathrm{P}(A \wedge A)$ do not necessarily coincide because $A$ and $A \wedge A$ may have different refutations (Muggleton, 1996; Cussens, 1999, 2001). Consequently in SLPs, we would be in trouble if we naively interpret $\mathrm{P}(A)$ as the probability of $A$'s being true. Also assigning probabilities to arbitrary quantified formulas seems out of scope of both approaches to SLPs.

Last but not least, there is a big problem common to any approach using probabilities: *where do the numbers come from*? Generally speaking, if we use $n$ binary random variables in a model, we have to determine $2^n$ probabilities to completely specify their joint distribution, and fulfilling this requirement with reliable numbers quickly becomes impossible as $n$ grows. The situation is even worse when there are unobservable variables in the model such as possible causes of a disease. Apparently parameter learning from observed data is a natural solution to this problem, but parameter learning of logic programs has not been well studied.

Distribution semantics proposed by Sato (1995) was an attempt to solve these problems along the line of the global distribution approach. It defines a distribution (probability measure) over the possible interpretations of ground atoms for an arbitrary logic program in any first order language and assigns consistent probabilities to all closed formulas. Also distribution semantics enabled us to derive an EM algorithm for the parameter learning of logic programs for the first time. As it was a naive algorithm however, dealing with large problems was difficult when there are exponentially many explanations for an observation like HMMs. We believe that the efficiency problem is solved to a large extent by the graphical EM algorithm presented in this paper.

### 7.3 EM Learning

Since EM learning is one of the central issues in this paper, we separately mention work related to EM learning for symbolic frameworks. Koller and Pfeffer (1997) used in their approach to KBMC (knowledge-based model construction) EM learning to estimate parameters labeling clauses. They express probabilistic dependencies among events by definite clauses annotated with probabilities, similarly to Ngo and Haddawy's (1997) approach, and locally build a Bayesian network relevant to the context and evidence as well as the

68. However, RPMs (recursive probability models) proposed by Pfeffer and Koller (2000) as an extension of Bayesian networks allow for infinitely many random variables. They are organized as attributes of classes and a probability measure over attribute values is introduced.

query. Parameters are learned by applying to the constructed network the specialized EM algorithm for Bayesian networks (Castillo et al., 1997).

Dealing with a PCFG by a statically constructed Bayesian network was proposed Pynadath and Wellman (1998), and it is possible to combine the EM algorithm with their method to estimate parameters in the PCFG. Unfortunately, the constructed network is not singly connected, and time complexity of probability computation is potentially exponential in the length of an input sentence.

Closely related to our EM learning is parameter learning of log-linear models. Riezler (1998) proposed the IM algorithm in his approach to probabilistic constraint programming. The IM algorithm is a general parameter estimation algorithm from incomplete data for log-linear models whose probability function $\mathrm{P}(x)$ takes the form $\mathrm{P}(x) = Z^{-1} \exp\left(\sum_{i=1}^{n} \lambda_i \nu_i(x)\right) p_0(x)$ where $(\lambda_1, \ldots, \lambda_n)$ are parameters to be estimated, $\nu_i(x)$ the $i$-th feature of an observed object $x$ and $Z$ the normalizing constant. Since a feature can be any function of $x$, the log-linear model is highly flexible and includes our distribution $P_{\mathtt{msw}}$ as a special case of $Z = 1$. There is a price to pay however; the computational cost of $Z$. It requires a summation over exponentially many terms. To avoid the cost of exact computation, approximate computation by a Monte Carlo method is possible. Whichever one may choose however, learning time increases compared to the EM algorithm for $Z = 1$.

The FAM (failure-adjusted maximization) algorithm proposed by Cussens (2001) is an EM algorithm applicable to pure normalized SLPs that may fail. It deals with a special class of log-linear models but is more efficient than the IM algorithm. Because the statistical framework of the FAM is rather different from distribution semantics, comparison with the graphical EM algorithm seems difficult.

Being slightly tangential to EM learning, Koller et al. (1997) developed a functional modeling language defining a probability distribution over symbolic structures in which they showed "cashing" of computed results leads to efficient probability computation of singly connected Bayesian networks and PCFGs. Their cashing corresponds to the computation of inside probability in the Inside-Outside algorithm and the computation of outside probability is untouched.

### 7.4 Future Directions

Parameterized logic programs are expected to be a useful modeling tool for complex symbolic-statistical phenomena. We have tried various types of modeling, besides stochastic grammars and Bayesian networks, such as the modeling of gene inheritance in the Kariera tribe (White, 1963) where the rules of bi-lateral cross-cousin marriage for four clans interact with the rules of genetic inheritance (Sato, 1998). The model was quite interdisciplinary, but the flexibility of combining `msw` atoms by means of definite clauses greatly facilitated the modeling process.

Although satisfying the five conditions in Section 4

- the uniqueness condition (roughly, one cause yields one effect)
- the finite support condition (there are a finite number of explanations for one observation)
- the acyclic support condition (explanations must not be cyclic)

- the t-exclusiveness condition (explanations must be mutually exclusive)
- the independence condition (events in an explanation must be independent)

for the applicability of the graphical EM algorithm seems daunting, our modeling experiences so far tell us that the modeling principle in Section 4 effectively guides us to successful modeling. In return, we can obtain a declarative model described compactly by a high level language whose parameters are efficiently learnable by the graphical EM algorithm as shown in the preceding section.

One of the future directions is however to relax some of the applicability conditions, especially the uniqueness condition that prohibits a generative model from failure or from generating multiple observable events. Although we pointed out in Section 4.4 that the MAR condition in Appendix B adapted to our semantics can replace the uniqueness condition and validates the use of the graphical EM algorithm even when a complete data does not uniquely determine the observed data just like the case of "partially bracketed corpora" (Pereira & Schabes, 1992), we feel the need to do more research on this topic. Also investigating the role of the acyclicity condition seems theoretically interesting as the acyclicity is often related to the learning of logic programs (Arimura, 1997; Reddy & Tadepalli, 1998).

In this paper we only scratched the surface of individual research fields such as HMMs, PCFGs and Bayesian networks. Therefore, there remains much to be done about clarifying how experiences in each research field are reflected in the framework of parameterized logic programs. For example, we need to clarify the relationship between symbolic approaches to Bayesian networks such as SPI (Li, Z. & D'Ambrosio, B., 1994) and our approach. Also it is unclear how a compiled approach using the junction tree algorithm for Bayesian networks can be incorporated into our approach. Aside from exact methods, approximate methods of probability computation specialized for parameterized logic programs must also be developed.

There is also a direction of improving learning ability by introducing priors instead of ML estimation to cope with data sparseness. The introduction of basic distributions that make probabilistic switches correlated seems worth trying in the near future. It is also important to take advantage of the logical nature of our approach to handle uncertainty. For example, it is already shown by Sato (2001) that we can learn parameters from negative examples such as "the grass is *not* wet" but the treatment of negative examples in parameterized logic programs is still in its infancy.

Concerning developing complex statistical models based on the "programs as distributions" scheme, stochastic natural language processing which exploits semantic information seems promising. For instance, unification-based grammars such as HPSGs (Abney, 1997) may be a good target beyond PCFGs because they use feature structures logically describable, and the ambiguity of feature values seems to be expressible by a probability distribution.

Also building a mathematical basis for logic programs with continuous random variables is a challenging research topic.

## 8. Conclusion

We have proposed a logical/mathematical framework for statistical parameter learning of parameterized logic programs, i.e. definite clause programs containing probabilistic facts with a parameterized probability distribution. It extends the traditional least Herbrand model semantics in logic programming to *distribution semantics*, possible world semantics with a probability distribution over possible worlds (Herbrand interpretations) which is unconditionally applicable to arbitrary logic programs including ones for HMMs, PCFGs and Bayesian networks.

We also have presented a new EM algorithm, the *graphical EM algorithm* in Section 4, which learns statistical parameters from observations for a class of parameterized logic programs representing a sequential decision process in which each decision is exclusive and independent. It works on *support graph*s, a new data structure specifying a logical relationship between an observed goal and its explanations, and estimates parameters by computing inside and outside probability generalized for logic programs.

The complexity analysis in Section 5 showed that when OLDT search, a complete tabled refutation method for logic programs, is employed for the support graph construction and table access is done in $O(1)$ time, the graphical EM algorithm, despite its generality, has the same time complexity as existing EM algorithms, i.e. the Baum-Welch algorithm for HMMs, the Inside-Outside algorithm for PCFGs and the one for singly connected Bayesian networks that have been developed independently in each research field. In addition, for pseudo probabilistic context sensitive grammars with $N$ nonterminals, we showed that the graphical EM algorithm runs in time $O(N^4L^3)$ for a sentence of length $L$.

To compare actual performance of the graphical EM algorithm against the Inside-Outside algorithm, we conducted learning experiments with PCFGs in Section 6 using two real corpora with contrasting characters. One is ATR corpus containing short sentences for which the grammar is not much ambiguous (958 parses/sentence), and the other is EDR corpus containing long sentences for which the grammar is rather ambiguous ($3.0 \times 10^8$ at average sentence length 20). In both cases, the graphical EM algorithm outperformed the Inside-Outside algorithm by orders of magnitude in terms of time per iteration, which suggests the effectiveness of our approach to EM learning by the graphical EM algorithm.

Since our semantics is not limited to finite domains or finitely many random variables but applicable to any logic programs of arbitrary complexity, the graphical EM algorithm is expected to give a general yet efficient method of parameter learning for models of complex symbolic-statistical phenomena governed by rules and probabilities.

## Acknowledgments

The authors wish to thank three anonymous referees for their comments and suggestions. Special thanks go to Takashi Mori and Shigeru Abe for stimulating discussions and learning experiments, and also to Tanaka-Tokunaga Laboratory for kindly allowing them to use MSLR parser and the linguistic data.

## Appendix A. Properties of $P_{DB}$

In this appendix, we list some properties of $P_{DB}$ defined by a parameterized logic program $DB = F \cup R$ in a countable first-order language $\mathcal{L}$.[69] First of all, $P_{DB}$ assigns consistent probabilities[70] to every closed formula $\phi$ in $\mathcal{L}$ by

$$P_{DB}(\phi) \stackrel{\text{def}}{=} P_{DB}(\{\omega \in \Omega_{DB} \mid \omega \models \phi\})$$

while guaranteeing continuity in the sense that

$$\begin{array}{rcl} \lim_{n\to\infty} P_{DB}(\phi(t_1) \wedge \cdots \wedge \phi(t_n)) & = & P_{DB}(\forall x \phi(x)) \\ \lim_{n\to\infty} P_{DB}(\phi(t_1) \vee \cdots \vee \phi(t_n)) & = & P_{DB}(\exists x \phi(x)) \end{array}$$

where $t_1, t_2, \ldots$ is an enumeration of ground terms in $\mathcal{L}$.

The next proposition, Proposition A.1, relates $P_{DB}$ to the Herbrand model. To prove it, we need some terminology. A *factor* is a closed formula in prenex disjunctive normal form $Q_1 \cdots Q_n M$ where $Q_i$ $(1 \leq i \leq n)$ is either an existential quantification or a universal quantification and $M$ a matrix. The length of quantifications $n$ is called the *rank* of the factor. Define $\Phi$ as a set of formulas made out of factors, conjunctions and disjunctions. Associate with each formula $\phi$ in $\Phi$ a multi-set $r(\phi)$ of ranks by

$$r(\phi) = \begin{cases} \emptyset & \text{if } \phi \text{ is a factor with no quantification} \\ \{n\} & \text{if } \phi \text{ is a factor with rank } n \\ r(\phi_1) \uplus r(\phi_2) & \text{if } \phi = \phi_1 \vee \phi_2 \text{ or } \phi = \phi_1 \wedge \phi_2. \end{cases}$$

Here $\uplus$ stands for the union of two multi-sets. For instance $\{1,2,3\} \uplus \{2,3,4\} = \{1,2,2,3,3,4\}$. We use the multi-set ordering in the proof of Proposition A.1 because the usual induction on the complexity of formulas does not work.

**Lemma A.1** *Let $\phi$ be a boolean formula made out of ground atoms in $\mathcal{L}$. $P_{DB}(\phi) = P_F(\{\nu \in \Omega_F \mid M_{DB}(\nu) \models \phi\})$.*

(Proof) We have only to prove the lemma about a conjunction of atoms of the form $D_1^{x_1} \wedge \cdots \wedge D_n^{x_n}$ $(x_i \in \{0,1\}, 1 \leq i \leq n)$.

$$\begin{array}{rcl} P_{DB}(D_1^{x_1} \wedge \cdots \wedge D_n^{x_n}) & = & P_{DB}(\{\omega \in \Omega_{DB} \mid \omega \models D_1^{x_1} \wedge \cdots \wedge D_n^{x_n}\}) \\ & = & P_{DB}(D_1 = x_1, \ldots, D_n = x_n) \\ & = & P_F(\{\nu \in \Omega_F \mid M_{DB}(\nu) \models D_1^{x_1} \wedge \cdots \wedge D_n^{x_n}\}) \end{array}$$

Q.E.D.

**Proposition A.1** *Let $\phi$ be a closed formula in $\mathcal{L}$. $P_{DB}(\phi) = P_F(\{\nu \in \Omega_F \mid M_{DB}(\nu) \models \phi\})$.*

69. For definitions of $\Omega_F$, $P_F$, $M_{DB}(\nu)$, $\Omega_{DB}$, $P_{DB}$ and others used below, see Section 3.
70. By consistent, we mean probabilities assigned to logical formulas respect the laws of probability such as $0 \leq P(A) \leq 1$, $P(\neg A) = 1 - P(A)$ and $P(A \vee B) = P(A) + P(B) - P(A \wedge B)$.

(Proof) Recall that a closed formula has an equivalent prenex disjunctive normal form that belongs to $\Phi$. We prove the proposition for formulas in $\Phi$ by using induction on the multi-set ordering over $\{r(\phi) \mid \phi \in \Phi\}$. If $r(\phi) = \emptyset$, $\phi$ has no quantification. So the proposition is correct by Lemma A.1. Suppose otherwise. Write $\phi = G[Q_1 Q_2 \cdots Q_n F]$ where $Q_1 Q_2 \cdots Q_n F$ indicates a *single* occurrence of a factor in $G$.[71] We assume $Q_1 = \exists x$ ($Q_1 = \forall x$ is similarly treated). We also assume that bound variables are renamed to avoid name clash. Then $G[\exists x Q_2 \cdots Q_n F]$ is equivalent to $\exists x G[Q_2 \cdots Q_n F]$ in light of the validity of $(\exists x A) \wedge B = \exists x(A \wedge B)$ and $(\exists x A) \vee B = \exists x(A \vee B)$ when $B$ contains no free $x$.

$$
\begin{aligned}
P_{DB}(\phi) &= P_{DB}(G[Q_1 Q_2 \cdots Q_n F]) \\
&= P_{DB}(\exists x G[\,Q_2 \cdots Q_n F[x]\,]) \\
&= \lim_{k \to \infty} P_{DB}(G[\,Q_2 \cdots Q_n F[t_1]] \vee \cdots \vee G[Q_2 \cdots Q_n F[t_k]\,]) \\
&= \lim_{k \to \infty} P_{DB}(G[\,Q_2 \cdots Q_n F[t_1] \vee \cdots \vee Q_2 \cdots Q_n F[t_k]\,]) \\
&= \lim_{k \to \infty} P_F(\{\nu \in \Omega_F \mid M_{DB}(\nu) \models G[\,Q_2 \cdots Q_n F[t_1] \vee \cdots \vee Q_2 \cdots Q_n F[t_k]\,]\}) \\
&\phantom{=} \text{(by induction hypothesis)} \\
&= P_F(\{\nu \in \Omega_F \mid M_{DB}(\nu) \models \exists x G[Q_2 \cdots Q_n F[x]]\}) \\
&= P_F(\{\nu \in \Omega_F \mid M_{DB}(\nu) \models \phi\}) \qquad \text{Q.E.D.}
\end{aligned}
$$

We next prove a theorem on the iff definition introduced in Section 4. Distribution semantics considers the program $DB = F \cup R$ as a set of infinitely many ground definite clauses such that $F$ is a set of facts (with a probability measure $P_F$) and $R$ a set of rules, and no clause head in $R$ appears in $F$. Put

$$head(R) \stackrel{\text{def}}{=} \{B \mid B \text{ appears in } R \text{ as a clause head}\}.$$

For $B \in head(R)$, let $B \leftarrow W_i$ $(i = 1, 2, \ldots)$ be an enumeration of clauses about $B$ in $R$. Define iff$(B)$, *the iff (if-and-only-if) form of rules about* $B$ in $DB$[72] by

$$\text{iff}(B) \stackrel{\text{def}}{=} B \leftrightarrow W_1 \vee W_2 \vee \cdots$$

Since $M_{DB}(\nu)$ is a least Herbrand model, the following is obvious.

**Lemma A.2** *For $B$ in $head(R)$ and $\nu \in \Omega_F$, $M_{DB}(\nu) \models iff(B)$.*

Theorem A.1 below is about iff$(B)$. It states that at general level, both sides of the iff definition $p(x) \leftrightarrow \exists y_1(x = t_1 \wedge W_1) \vee \cdots \vee \exists y_n(x = t_n \wedge W_n)$ of $p(\cdot)$ coincide as random variables whenever $x$ is instantiated to a ground term.

**Theorem A.1** *Let $iff(B) = B \leftrightarrow W_1 \vee W_2 \vee \cdots$ be the iff form of rules about $B \in head(R)$. $P_{DB}(iff(B)) = 1$ and $P_{DB}(B) = P_{DB}(W_1 \vee W_2 \vee \cdots)$.*

71. For an expression $E$, $E[\gamma]$ means that $\gamma$ may occur in the specified positions of $E$. If $\gamma_1 \vee \gamma_2$ in $E[\gamma_1 \vee \gamma_2]$ indicates a *single* occurrence of $\gamma_1 \vee \gamma_2$ in a positive boolean formula $E$, $E[\gamma_1 \vee \gamma_2] = E[\gamma_1] \vee E[\gamma_2]$ holds.
72. This definition is different from the usual one (Lloyd, 1984; Doets, 1994) as we are here talking at ground level. $W_1 \vee W_2 \vee \cdots$ is true if and only if one of the disjuncts is true.

(Proof)

$$
\begin{aligned}
P_{DB}(\mathrm{iff}(B)) &= P_{DB}(\{\omega \in \Omega_{DB} \mid \omega \models B \wedge (W_1 \vee W_2 \vee \cdots)\}) \\
&\quad + P_{DB}(\{\omega \in \Omega_{DB} \mid \omega \models \neg B \wedge \neg(W_1 \vee W_2 \vee \cdots)\}) \\
&= \lim_{k\to\infty} P_{DB}(\{\omega \in \Omega_{DB} \mid \omega \models B \wedge \bigvee_{i=1}^{k} W_i\}) \\
&\quad + \lim_{k\to\infty} P_{DB}(\{\omega \in \Omega_{DB} \mid \omega \models \neg B \wedge \neg \bigvee_{i=1}^{k} W_i\}) \\
&= \lim_{k\to\infty} P_F(\{\nu \in \Omega_F \mid M_{DB}(\nu) \models B \wedge \bigvee_{i=1}^{k} W_i\}) \\
&\quad + \lim_{k\to\infty} P_F(\{\nu \in \Omega_F \mid M_{DB}(\nu) \models \neg B \wedge \neg \bigvee_{i=1}^{k} W_i\}) \\
&\quad \text{(Lemma A.1)} \\
&= P_F(\{\nu \in \Omega_F \mid M_{DB}(\nu) \models \mathrm{iff}(B)\}) \\
&= P_F(\Omega_F) \ \text{(Lemma A.2)} \\
&= 1
\end{aligned}
$$

It follows from $P_{DB}(\mathrm{iff}(B)) = 1$ that

$$P_{DB}(B) = P_{DB}(B \wedge \mathrm{iff}(B)) = P_{DB}(W_1 \vee W_2 \vee \cdots). \qquad \text{Q.E.D.}$$

We then prove a proposition useful in probability computation. Let $\psi_{DB}(B)$ be the support set for an atom $B$ introduced in Section 4 (it is the set of all explanations for $B$). In the sequel, $B$ is a ground atom. Write $\psi_{DB}(B) = \{S_1, S_2, \ldots\}$ and $\bigvee \psi_{DB}(B) = S_1 \vee S_2 \vee \cdots$[73] Define a set $\Lambda_B$ by

$$\Lambda_B \stackrel{\text{def}}{=} \{\omega \in \Omega_{DB} \mid \omega \models B \leftrightarrow \bigvee \psi_{DB}(B)\}.$$

**Proposition A.2** *For every $B \in head(R)$, $P_{DB}(\Lambda_B) = 1$ and $P_{DB}(B) = P_{DB}(\bigvee \psi_{DB}(B))$.*

(Proof) We first prove $P_{DB}(\Lambda_B) = 1$ but the proof exactly parallels that of Theorem A.1 except that $W_1 \vee W_2 \vee \cdots$ is replaced by $S_1 \vee S_2 \vee \cdots$ using the fact that $B \leftrightarrow S_1 \vee S_2 \vee \cdots$ is true in every least Herbrand model of the form $M_{DB}(\nu)$. Then from $P_{DB}(\Lambda_B) = 1$, we have

$$
\begin{aligned}
P_{DB}(B) &= P_{DB}(B \wedge (B \leftrightarrow \bigvee \psi_{DB}(B))) \\
&= P_{DB}(\bigvee \psi_{DB}(B)). \qquad \text{Q.E.D.}
\end{aligned}
$$

Finally, we show that distribution semantics is a probabilistic extension of the traditional least Herbrand model semantics in logic programming by proving Theorem A.2. It says that the probability mass is distributed exclusively over possible least Herbrand models.

Define $\Lambda$ as the set of least Herbrand models generated by fixing $R$ and varying a subset of $F$ in the program $DB = F \cup R$. In symbols,

73. For a set $K = \{E_1, E_2, \ldots\}$ of formulas, $\bigvee K$ denotes a (-n infinite) disjunction $E_1 \vee E_2 \vee \cdots$

$$\Lambda \stackrel{\text{def}}{=} \{\omega \in \Omega_{DB} \mid \omega = M_{DB}(\nu) \text{ for some } \nu \in \Omega_F\}.$$

Note that as $\Lambda$ is merely a subset of $\Omega_{DB}$, we cannot conclude $P_{DB}(\Lambda) = 1$ a priori, but the next theorem, Theorem A.2, states $P_{DB}(\Lambda) = 1$, i.e. distribution semantics distributes the probability mass exclusively over $\Lambda$, i.e. possible least Herbrand models.

To prove the theorem, we need some preparations. Recalling that atoms outside $\mathit{head}(R) \cup F$ have no chance of being proved from $DB$, we introduce

$$\Lambda' \stackrel{\text{def}}{=} \{\omega \in \Omega_{DB} \mid \omega \models \neg D \text{ for every ground atom } D \notin \mathit{head}(R) \cup F\}.$$

For a Herbrand interpretation $\omega \in \Omega_{DB}$, $\omega|_F$ ($\in \Omega_F$) is the restriction of $\omega$ to those atoms in $F$.

**Lemma A.3** *Let $\omega \in \Omega_{DB}$ be a Herbrand interpretation.*
*$\omega = M_{DB}(\nu)$ for some $\nu \in \Omega_F$ iff $\omega \in \Lambda'$ and $\omega \models B \leftrightarrow \bigvee \psi_{DB}(B)$ for every $B \in \mathit{head}(R)$.*

(Proof) Only-if part is immediate from the property of the least Herbrand model. For if-part, suppose $\omega$ satisfies the right hand side. We show that $\omega = M_{DB}(\omega|_F)$. As $\omega$ and $M_{DB}(\omega|_F)$ coincide w.r.t. atoms not in $\mathit{head}(R)$, it is enough to prove that they also give the same truth values to atoms in $\mathit{head}(R)$. Take $B \in \mathit{head}(R)$ and write $\bigvee \psi_{DB}(B) = S_1 \vee S_2 \vee \cdots$ Suppose $\omega \models B \leftrightarrow S_1 \vee S_2 \vee \cdots$ Then if $\omega \models B$, we have $\omega \models S_j$ for some $j$, thereby $\omega|_F \models S_j$, and hence $M_{DB}(\omega|_F) \models S_j$, which implies $M_{DB}(\omega|_F) \models B$. Otherwise $\omega \models \neg B$. So $\omega \models \neg S_j$ for every $j$. It follows that $M_{DB}(\omega|_F) \models \neg B$. Since $B$ is arbitrary, we conclude that $\omega$ and $M_{DB}(\omega|_F)$ agree on the truth values assigned to atoms in $\mathit{head}(R)$ as well. Q.E.D.

**Theorem A.2** $P_{DB}(\Lambda) = 1$.

(Proof) From Lemma A.3, we have

$$\begin{aligned} \Lambda &= \{\omega \in \Omega_{DB} \mid \omega = M_{DB}(\nu) \text{ for some } \nu \in \Omega_F\} \\ &= \Lambda' \cap \bigcap_{B \in \mathit{head}(R)} \Lambda_B. \end{aligned}$$

$P_{DB}(\Lambda_B) = 1$ by Proposition A.2. To prove $P_{DB}(\Lambda') = 1$, let $D_1, D_2, \ldots$ be an enumeration of atoms not belonging to $\mathit{head}(R) \cup F$. They are not provable from $DB = F \cup R$, and hence false in every least Herbrand model $M_{DB}(\nu)$ ($\nu \in \Omega_F$). So

$$\begin{aligned} P_{DB}(\Lambda') &= \lim_{m \to \infty} P_{DB}(\{\omega \in \Omega_{DB} \mid \omega \models \neg D_1 \wedge \cdots \wedge \neg D_m\}) \\ &= \lim_{m \to \infty} P_F(\{\nu \in \Omega_F \mid M_{DB}(\nu) \models \neg D_1 \wedge \cdots \wedge \neg D_m\}) \\ &= P_F(\Omega_F) = 1. \end{aligned}$$

Since a countable conjunction of measurable sets of probability measure one has also probability measure one, it follows from $P_{DB}(\Lambda_B) = 1$ for every $B \in \mathit{head}(R)$ and $P_{DB}(\Lambda') = 1$ that $P_{DB}(\Lambda) = 1$. Q.E.D.

## Appendix B. The MAR (missing at random) Condition

In the original formulation of the EM algorithm by Dempster et al. (1977), it is assumed that there exists a many-to-one mapping $y = \chi(x)$ from a complete data $x$ to an incomplete (observed) data $y$. In the case of parsing, $x$ is a parse tree and $y$ is the input sentence and $x$ uniquely determines $y$. In this paper, the uniqueness condition ensures the existence of such a many-to-one mapping from explanations to observations. We however sometimes face a situation where there is no such many-to-one mapping from complete data to incomplete data but nonetheless we wish to apply the EM algorithm.

This dilemma can be solved by the introduction of a missing-data mechanism which makes a complete data incomplete. The missing-data mechanism, $m$, has a distribution $g_\phi(m \mid x)$ parameterized by $\phi$ and $y$, the observed data, is described as $y = \chi_m(x)$. It says $x$ becomes incomplete $y$ by $m$. The correspondence between $x$ and $y$, i.e. $\{\langle x, y\rangle \mid \exists m(y = \chi_m(x))\}$ naturally becomes many-to-many.

Rubin (1976) derived two conditions on $g_\phi$ (data are missing at random and data are observed at random) collectively called the *MAR (missing at random) condition*, and showed that if we assume a missing-data mechanism behind our observations that satisfies the MAR condition, we may estimate parameters of the distribution over $x$ by simply applying the EM algorithm to $y$, the observed data.

We adapt the MAR condition to parameterized logic programs as follows. We keep a generative model satisfying the uniqueness condition that outputs goals $G$ such as parse trees. We further extend the model by additionally inserting a missing-data mechanism $m$ between $G$ and our observation $O$ like $O = \chi_m(G)$ and assume $m$ satisfies the MAR condition. Then the extended model has a many-to-many correspondence between explanations and observations, and generates non-exclusive observations such that $P(O \wedge O') > 0$ $(O \neq O')$, which causes $\sum_O P(O) \geq 1$ where $P(O) = \sum_{G:\exists m\ O=\chi_m(G)} P_{DB}(G)$. Thanks to the MAR condition however, we are still allowed to apply the EM algorithm to such non-exclusive observations. Put it differently, even if the uniqueness condition is seemingly destroyed, the EM algorithm is applicable just by (imaginarily) assuming a missing-data mechanism satisfying the MAR condition.